\documentclass{article}

     \PassOptionsToPackage{numbers, compress}{natbib}
    \usepackage[preprint]{neurips_2024}

\usepackage[utf8]{inputenc} % allow utf-8 input
\usepackage[T1]{fontenc}    % use 8-bit T1 fonts
\usepackage{url}            % simple URL typesetting
\usepackage{booktabs}       % professional-quality tables
\usepackage{amsfonts}       % blackboard math symbols
\usepackage{amsmath}
\usepackage{amssymb}
\usepackage{nicefrac}       % compact symbols for 1/2, etc.
\usepackage{microtype}      % microtypography
\usepackage{xcolor}         % colors
\usepackage{stix}

\usepackage{mathtools,xparse}

\usepackage{multirow}
\usepackage{makecell} %this is for mark cell
\usepackage{tabulary}
\usepackage{subfigure}

\usepackage{bm}
\usepackage{color}
\usepackage{colortbl} % setting table color
\usepackage{ulem}
\usepackage{float}

\usepackage{pifont}% http://ctan.org/pkg/pifont
\newcommand{\cmark}{\ding{51}}%
\newcommand{\xmark}{\ding{55}}%

\newcommand{\dt}{\Delta}

\usepackage{xspace}

\title{Mamba-360: Survey of State Space Models as Transformer Alternative for Long Sequence Modelling: Methods, Applications, and Challenges}

\author{Badri Narayana Patro\\
Microsoft\\
{\tt\small badripatro@microsoft.com}
\and
Vijay Srinivas Agneeswaran\\
Microsoft\\
{\tt\small vagneeswaran@microsoft.com}
}

\begin{document}

\maketitle

\begin{abstract}

Sequence modeling is a crucial area across various domains, including Natural Language Processing (NLP), speech recognition, time series forecasting, music generation, and bioinformatics. Recurrent Neural Networks (RNNs) and Long Short Term Memory Networks (LSTMs)  have historically dominated sequence modeling tasks like Machine Translation, Named Entity Recognition (NER), etc. However, the advancement of transformers has led to a shift in this paradigm, given their superior performance. Yet, transformers suffer from $O(N^2)$ attention complexity and challenges in handling inductive bias. Several variations have been proposed to address these issues which use spectral networks or convolutions and have performed well on a range of tasks. However, they still have difficulty in dealing with long sequences. State Space Models(SSMs) have emerged as promising alternatives for sequence modeling paradigms in this context, especially with the advent of S4 and its variants, such as S4nd, Hippo, Hyena, Diagnol State Spaces (DSS), Gated State Spaces (GSS), Linear Recurrent Unit (LRU), Liquid-S4, Mamba, etc. In this survey, we categorize the foundational SSMs based on three paradigms namely, Gating architectures, Structural architectures, and Recurrent architectures. This survey also highlights diverse applications of SSMs across domains such as vision, video, audio, speech, language (especially long sequence modeling), medical (including genomics), chemical (like drug design), recommendation systems, and time series analysis, including tabular data. Moreover, we consolidate the performance of SSMs on benchmark datasets like Long Range Arena (LRA), WikiText, Glue, Pile, ImageNet, Kinetics-400, sstv2, as well as video datasets such as Breakfast, COIN, LVU, and various time series datasets. The project page for Mamba-360 work is available on this webpage.\url{https://github.com/badripatro/mamba360}.

\end{abstract}
% \vspace{-0.45cm}
%%%%%%%%% BODY TEXT

\section{Introduction}
Recurrent Neural Networks (RNNs) have long been the cornerstone of sequence modeling, excelling in tasks like machine translation and next-word prediction. RNNs predict the next state given the current input token and the previous state. However, RNNs allow us to look at only the last state and current input for predicting the next state, implying that output is restrictive. Recurrent Neural Networks (RNNs) can efficiently handle sequences of length L without requiring memory resources greater than O(1). However, due to gradient calculations being limited to the hidden state and current input, RNNs may not have a sufficient look-back window or memory capacity. Moreover, RNNs struggle with the exploding or vanishing gradient problem and lack sufficient memory for long sequences without an exponential increase in computational complexity. To address these limitations, Long Short-Term Memory (LSTM) networks were introduced. While LSTMs mitigate some issues of traditional RNNs, they introduce complexity with their gating mechanisms and exhibit challenges in transfer learning.

Transformers have emerged as a revolutionary alternative, offering solutions to the shortcomings of both RNNs and LSTMs, and have become the dominant paradigm in the fields of NLP and vision. By employing attention mechanisms, transformers enable each token to interact with every other token in the input sequence, thereby capturing long-range dependencies efficiently. However, the $O(N^2)$ attention complexity of transformers poses scalability challenges, especially for processing long sequences in domains like genomics and high-resolution image analysis. This implies though Transformers perform quite well on many sequence processing tasks, they have certain inefficiencies, especially with respect to the amount of memory and computing required, which grow quadratically with sequence length. In this context, our study explores the landscape of sequence modeling and introduces State Space Models (SSMs) as a promising paradigm. By leveraging SSMs, we aim to address the inefficiencies of traditional RNNs and transformers while offering scalable solutions for sequence processing tasks across diverse domains.

\begin{figure*}
\centering
  \includegraphics[width=0.95\linewidth]{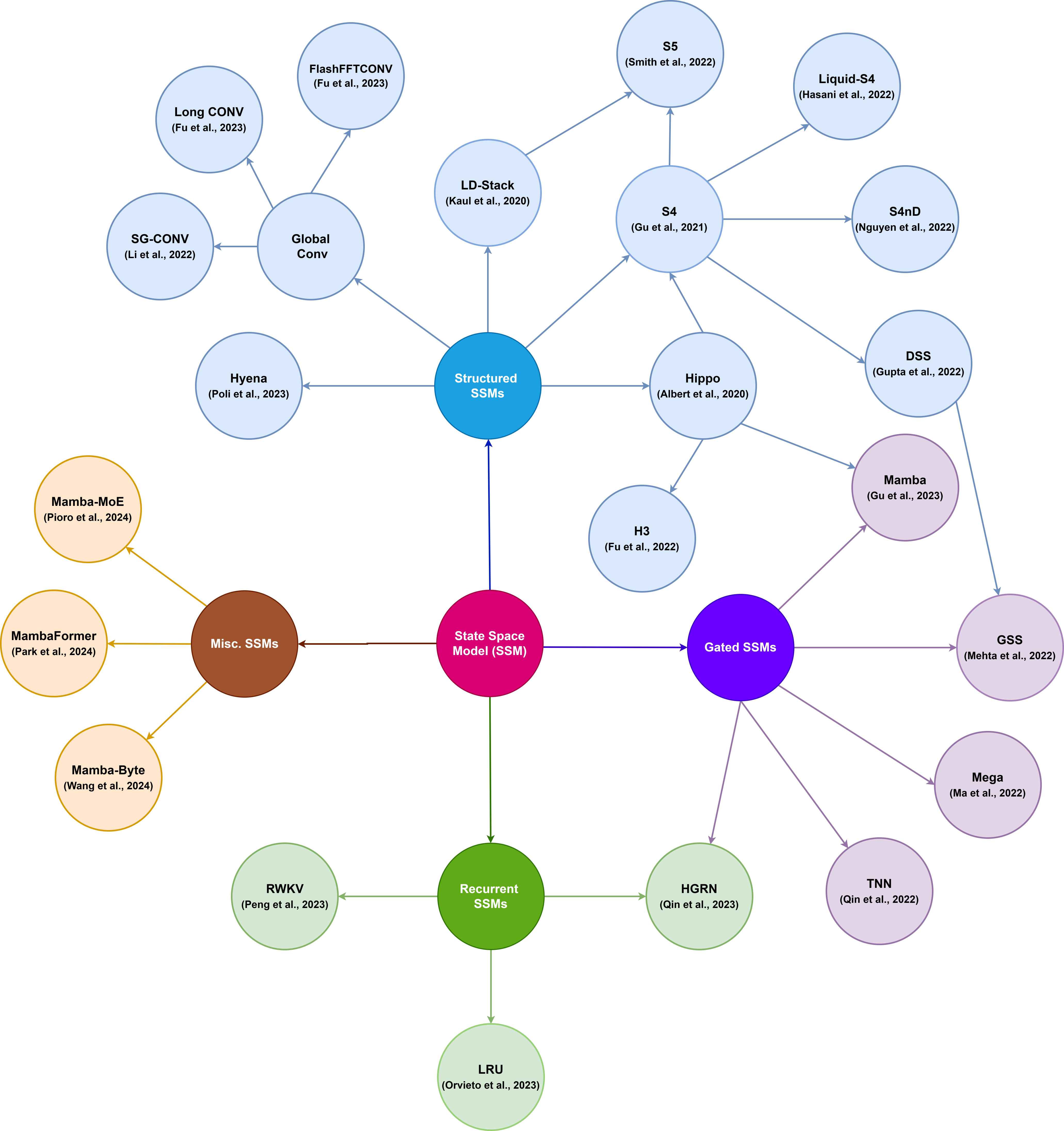}
  \caption{Categorization of State Space Models (SSMs) based on their structural, recurrent, and gated nature. We discuss key SSMs from the literature for each category.}
  \label{fig:mamba_arch}
  \vspace{-0.3in}
\end{figure*}

State Space Models (SSMs) have emerged as compelling alternatives to transformers, particularly for processing long sequences. SSMs can be conceptualized as RNNs with fixed lengths, which do not grow with input length This brings significant efficiency benefits in terms of inference speed and computation/memory complexity compared to transformers. However, despite their efficiency advantages, SSMs often fall short of the performance gap with state-of-the-art transformers in certain data modalities, notably in vision tasks.

A notable drawback of SSMs is their compromise on core capabilities essential for certain sequence processing tasks, such as copying long input sequences\cite{jelassi2024repeat}, in-context learning, and induction heads\cite{olsson2022context}. This work presents a comprehensive survey of the field of state space models, highlighting their strengths and weaknesses compared to state-of-the-art transformers. We discuss how task-oriented SSMs are better suited than transformers for specific tasks, such as those in the long-range arena tasks \cite{tay2020long} while acknowledging the superior performance of transformers in computer vision tasks like image recognition and instance segmentation.

In our survey, we observe that SSMs, especially exemplified by models like Mamba, demonstrate competitive performance with transformers across various domains and tasks. For instance, in the realm of Language Domain Tasks, SSMs exhibit commendable performance, particularly in standard regression in-context learning (ICL) tasks. Notably, SSMs outshine transformers in tasks such as sparse parity learning¹. However, in the domain of Video and Audio Tasks, while transformers show promise in natural language processing, their effectiveness in multimodal tasks, such as video and audio understanding, remains relatively unexplored. Addressing this gap, ongoing research endeavors aim to leverage the long-range modeling capacity of Large Language Models (LLMs) to enhance video understanding. 

Furthermore, our survey highlights the need for further investigation into the relative strengths of SSMs and transformers across a range of tasks, including time series prediction, recommendation systems, reinforcement learning, and various medical domain tasks. These domains offer fertile ground for exploring how SSMs and transformers can complement each other, providing insights into their respective capabilities and limitations.

This work incorporates four important aspects including:
\begin{itemize}
\item 
\textbf{Understanding of State Space Models (SSMs)}: This work discusses the fundamentals of SSMs, and explains their inner workings and mathematical underpinnings.
\item  \textbf{Categorization and Recent Advances of SSMs}: It provides a systematic categorization of SSMs, shedding light on recent developments in this field. By organizing SSMs, researchers gain insights into their unique characteristics and potential applications.
% Provides a categorization of the state space models and brings to light recent SSMs.
\item \textbf{Application of SSMs Across Domains}: The study explores how SSMs find utility in diverse domains, from natural language processing to medical diagnostics. Understanding these applications helps practitioners tailor SSMs to specific tasks effectively.
% Applications - details out the applications of SSMs in various domains.
\item \textbf{Performance Comparison of SSMs with Transformers}: By consolidating state-of-the-art results, the work evaluates SSMs alongside Transformers. This comparative analysis informs us about the strengths and limitations of each approach for specific domains and tasks. 
% Consolidated the state-of-art results in various domains and compared performance of various SSMs and transformers.
\end{itemize}

\begin{figure}
\centering
  \includegraphics[width=0.95\linewidth]{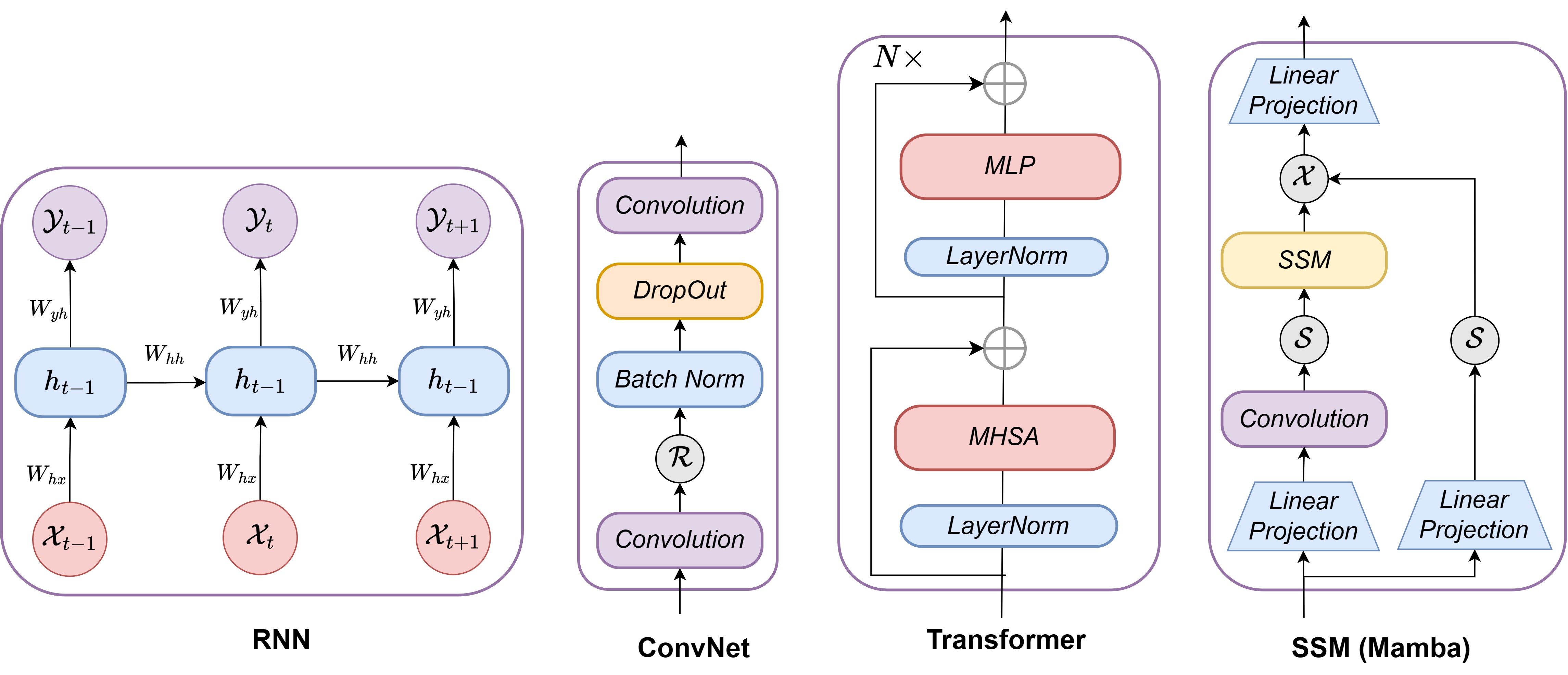}
  \caption{This figure illustrates the evolutionary progression of sequential data modeling paradigms, from Recurrent Neural Networks (RNNs) and Convolutional Neural Networks (CNNs) to Transformer models and State-Space Models (SSMs), highlighting advancements in capturing temporal dynamics, spatial hierarchies, and complex system interactions.}
  \label{fig:model_evolution}
  % \vspace{-0.3in}
\end{figure}

In summary, we introduce the basics of State Space Models (SSMs) and discuss their fundamental principles, including their mathematical formulation and conceptual framework, in Section-\ref{basic_ssm}. In Section \ref{ssm}, we outline recent advances in the SSM literature, showcasing recent breakthroughs and shedding light on cutting-edge research and innovations. In Section-\ref{application_ssm}, we explore the applications of SSMs in modeling long sequences, examining their efficacy and performance in comparison to other models. In Section-\ref{sota}, we discuss the performance of SSMs in modeling long sequences and compare it with that of state-of-the-art Transformers. We show that while SSMs achieve efficiencies compared to Transformers, they still fall short of the performance of state-of-the-art Transformers in certain domains. The evolution of sequential data modeling approaches from Recurrent Neural Networks (RNNs), Convolutional Neural Networks (CNNs), Transformers, to State-Space Models (SSMs) reflects a trajectory of innovations addressing diverse challenges in capturing temporal dependencies, spatial hierarchies, global interactions, and dynamic systems behavior as shown in figure-\ref{fig:model_evolution}.

\section{Basics of State Space Models}\label{basic_ssm}

State-space models offer a powerful framework for modeling dynamic systems, particularly by enabling the representation of high-order derivatives through the utilization of only first-order derivatives and vector quantities. For instance, consider the second-order differential equation describing the dynamics of a damped mass-spring system:

\[m \frac{d^2y(t)}{dt^2} + c \frac{dy(t)}{dt} + ky(t) = u(t),\]

where \(u(t)\) represents the external force acting on the mass and \(y(t)\) denotes the vertical position. Here, \(\frac{dy(t)}{dt}\) and \(\frac{d^2y(t)}{dt^2}\) stand for the first and second derivatives of \(y\) respectively.

To express this equation solely in terms of first-order derivatives and vector quantities, we introduce the state vector:

\[ x(t) := \begin{pmatrix} y(t) \\ \dot{y}(t) \end{pmatrix}. \]

Though this transition results in dealing with a vector equation instead of a scalar one:

\[ \dot{x}(t) = \begin{pmatrix} 0 & 1 \\ -\frac{c}{m} & -\frac{k}{m} \end{pmatrix} x(t) + \begin{pmatrix} 0 \\ 1 \end{pmatrix} u(t). \]

The position \(y(t)\) is then expressed as a linear function of the state:

\[ y(t) = C x(t), \]

with \(C = (1,0)\).

\subsection{Spring-Mass-Damper system}
The spring-mass-damper system is a classic example used to illustrate principles of dynamics and control theory. Here's the basic mathematical formulation of the state-space model for the spring-mass-damper system. Consider a spring-mass-damper system consisting of a mass \(m\) connected to a wall via a spring with spring constant \(k\) and a damper with a damping coefficient \(c\).  The goal is to describe the system's behavior using state variables. The displacement of the mass is represented as \(x\), its velocity as \(\dot{x}\), and the external force applied to the mass as \(F\). 

1. State Variables: We'll define two state variables: \(x_1\): Displacement of the mass from its equilibrium position (position of the mass relative to the spring's rest position). The \(\dot{x}_1\): Velocity of the mass.

2. System Dynamics: The dynamics of the spring-mass-damper system can be expressed using Newton's second law of motion as follows:
     \[m\ddot{x}_1 = -kx_1 - c\dot{x}_1\]
     where \(\ddot{x}_1\) represents the acceleration of the mass. The first term \(-kx_1\) represents the spring force (proportional to displacement). The second term \(-c\dot{x}_1\) represents the damping force (proportional to velocity).

3. State-Space Formulation:

A state-space model represents the dynamics of a system using a set of first-order differential equations. It's a powerful framework for describing linear time-invariant (LTI) systems. Its fundamental components are:
\begin{itemize}
    \item  State Vector (x): The state vector contains the state variables that describe the system's internal state. Let's denote it as \(\mathbf{x}\), where \(\mathbf{x} \in \mathbb{R}^n\).

    \item   Input Vector (u): The input vector represents the control or external input to the system. Denote it as \(\mathbf{u}\), where \(\mathbf{u} \in \mathbb{R}^m\).
    
    \item   Output Vector (y): The output vector contains the measurable quantities of interest. Denote it as \(\mathbf{y}\), where \(\mathbf{y} \in \mathbb{R}^p\).
    
    \item  System Dynamics: The state dynamics are described by the first-order differential equation. In the Time-Invariant (LTI) case, the matrices \(\mathbf{A}\), \(\mathbf{B}\), \(\mathbf{C}\), and \(\mathbf{D}\)  remain constant over time. The LTI state dynamics become
\begin{itemize}
    \item 
    \[\dot{\mathbf{x}} = \mathbf{Ax} + \mathbf{Bu}\]
     where:\(\mathbf{x} = [x_1, \dot{x}_1]^T\) is the state vector. \(\mathbf{u}\) represents the input (external force or control input). \(\mathbf{A} \in \mathbb{R}^{n \times n}\) is the dynamics matrix. \(\mathbf{B} \in \mathbb{R}^{n \times m}\) is the input matrix. The state matrix A is defined as \(\mathbf{A} = \begin{bmatrix} 0 & 1 \\ -\frac{k}{m} & -\frac{c}{m} \end{bmatrix}\), the input matrix \(\mathbf{B} = \begin{bmatrix} 0 \\ \frac{1}{m} \end{bmatrix}\)
     \(\dot{\mathbf{x}}\) represents the time derivative of \(\mathbf{x}\).

\item The output equation relates the output to the state and input:
     \[\mathbf{y} = \mathbf{Cx} + \mathbf{Du}\]
     where \(\mathbf{C} \in \mathbb{R}^{p \times n}\) is the output or sensor matrix. \(\mathbf{D} \in \mathbb{R}^{p \times m}\) is the feedthrough matrix.  \(\mathbf{C} = \begin{bmatrix} 1 & 0 \end{bmatrix}\) and \(\mathbf{D} = 0\)
     
\end{itemize}
\end{itemize}

4. Interpretation:  The state vector \(\mathbf{x}\) contains information about the mass's position and velocity.  The input vector \(\mathbf{u}\) could represent an external force applied to the mass.  The output vector \(\mathbf{y}\) is typically the displacement \(x_1\). In this system, the state vector \(\mathbf{x}\) represents the position and velocity of the mass. The state dynamics are governed by the equations of motion derived from Newton's second law. The input \(u\) represents any external force applied to the mass, which in this case is assumed to be zero.

5. Stability and Control: Stability analysis involves examining the eigenvalues of the matrix \(\mathbf{A}\).  Control design can be achieved by adjusting the control input \(\mathbf{u}\) to achieve desired behavior (e.g., damping out oscillations).

\subsection{State Space Models}\label{sec:ss-continuous}

\begin{figure}
\centering

  \includegraphics[width=0.85\linewidth]{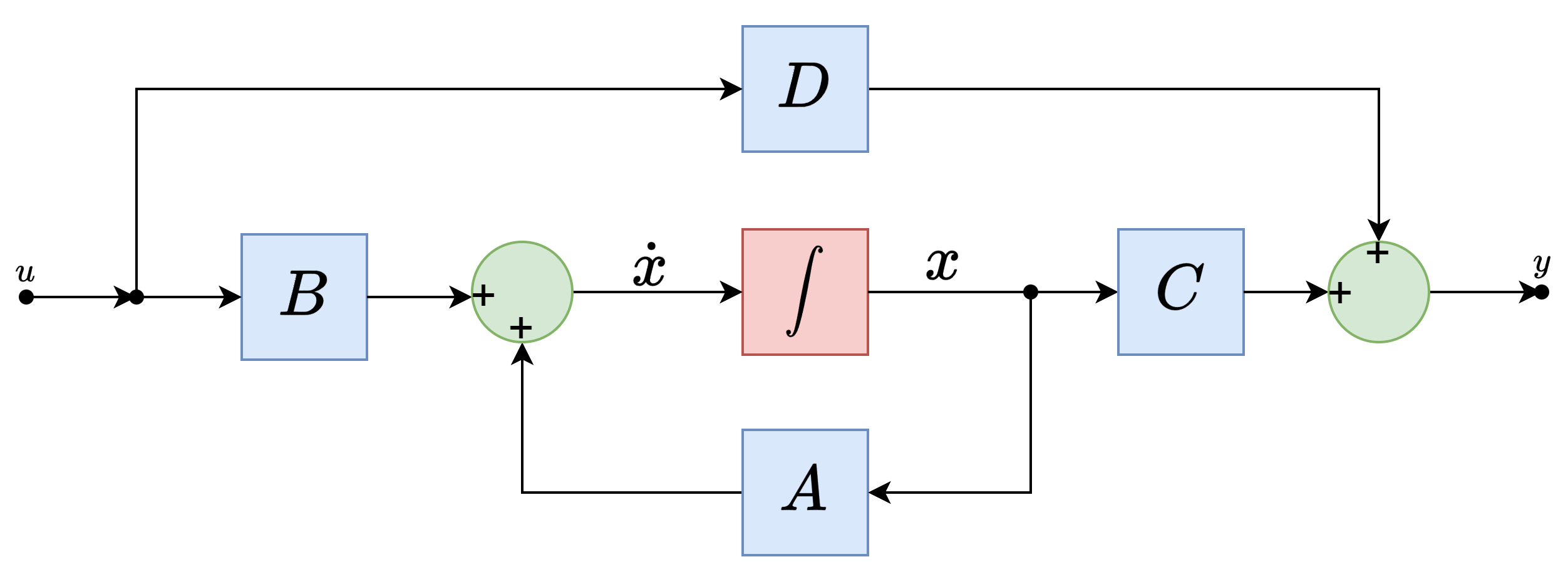}  
   % \vspace{-0.in}
  \caption{Illustration depicting the concept of a state-space model, which describes the system dynamics through a series of first-order differential equations.}
  \label{fig:ssm} 
\end{figure}
\subsubsection{Definition}
Linear state-space equations provide a versatile framework for modeling discrete-time dynamical systems:

\[ x(t+1) = Ax(t) + Bu(t), \quad y(t) = Cx(t) + Du(t), \quad t=0,1,2,\ldots \]

Here, \( x(t) \) in \( \mathbf{R}^n \) represents the system's state at time \( t \), encapsulating its condition. \( u(t) \) in \( \mathbf{R}^p \) contains control variables, while \( y(t) \) in \( \mathbf{R}^k \) comprises specific outputs of interest. Matrices \( A \), \( B \), \( C \), and \( D \) are of appropriate sizes.

Essentially, a linear dynamical model assumes that the state at the next time step is a linear combination of the state at previous time steps and, potentially, other exogenous inputs. Additionally, it posits that the output is a linear function of the state and input vectors.

In contrast, a continuous-time model takes the form of a differential equation:

\[ \frac{d}{dt} x(t) = A x(t) + Bu(t), \quad y(t) = Cx(t) + Du(t), \quad t \geq 0. \]

Lastly, time-varying models incorporate matrices \( A \), \( B \), \( C \), \( D \) that change over time, offering a more flexible representation of dynamic systems.

\subsubsection{Model Formulation:}

To model a large sequence we use state space models instead of Multi-headed self-attention due to its complexity. The state space model\cite{gu2021efficiently,gu2023mamba} is commonly known as a linear time-invariant system that maps the input stimulation $x(t) \in \mathcal{R}^L $ to a response $y(t) \ $ through a hidden space $h(t) \in \mathcal{R}^N $. Structured state space sequence models (S4) are a recent class of sequence models for deep learning that are broadly related to RNNs, CNNs, and classical state space models. Mathematically,  The Continuous-time Latent State spaces can be modeled as linear ordinary differential equations that use evolution parameter $A \in \mathcal{R}^{N\times N} $ and projection parameter $B \in \mathcal{R}^{N\times 1} $ and $C \in \mathcal{R}^{N\times 1} $ as follows:

\begin{equation}
  \label{eq:1}
  \begin{aligned}
    \dot{x}(t) &= \bm{A}x(t) + \bm{B}u(t) \\
    y(t) &= \bm{C}x(t) + \bm{D}u(t)
  \end{aligned}
\end{equation}

\subsubsection{Discrete-time SSM: }
\label{sec:ss-recurrent}
The discrete form of SSM uses a time-scale parameter $\dt$ to transform continuous parameters A, B, and C to discrete parameters $\Bar{A}, \Bar{B}$ and $ \Bar{C} $ using fixed formula  $\Bar{A}=f_{A}(\dt,A), \Bar{B}=f_{B}(\dt,A,B)$. The pair $f_{A},f_{B}$ is the discretization rule that uses a zero-order hold (ZOH)  for this transformation. The equations are as follows.

\begin{equation}
  \label{eq:2}
\begin{aligned}
  x_{k} &= \bm{\overline{A}} x_{k-1} + \bm{\overline{B}} u_k &
  \bm{\overline{A}} &= (\bm{I} - \dt/2 \cdot \bm{A})^{-1}(\bm{I} + \dt/2 \cdot \bm{A}) &
  \\
  y_k &= \bm{\overline{C}} x_k &
  \bm{\overline{B}} &= (\bm{I} - \dt/2 \cdot \bm{A})^{-1} \dt \bm{B} &
  \bm{\overline{C}} &= \bm{C}
  .
\end{aligned}
\end{equation}
% Equation \eqref{eq:2} is now a \emph{sequence-to-sequence} map \( u_k \mapsto y_k \) instead of function-to-function. 

\subsubsection{ Convolutional Kernel Representation}
\label{sec:ss-convolution}
The discretized form of recurrent SSM in equation-\ref{eq:2} is not practically trainable due to its sequential nature. To get efficient representation, we model continuous convolution as discrete convolution as it is a linear time-invariant system. %The recurrent SSM \eqref{eq:2} is not practical for training on modern hardware due to its sequentiality. Instead, there is a well-known connection between linear time-invariant (LTI) SSMs such as \eqref{eq:1} and continuous convolutions. Correspondingly, \eqref{eq:2} can actually be written as a discrete convolution.
For simplicity let the initial state be \( x_{-1} = 0 \).
Then unrolling \eqref{eq:2} explicitly yields: %
\begin{align*}
  x_0 &= \bm{\overline{B}} u_0 &
  x_1 &= \bm{\overline{A}} \bm{\overline{B}} u_0 + \bm{\overline{B}} u_1 &
  x_2 &= \bm{\overline{A}}^2 \bm{\overline{B}} u_0 + \bm{\overline{A}} \bm{\overline{B}} u_1 + \bm{\overline{B}} u_2 & \dots
  \\
  y_0 &= \bm{\overline{C}} \bm{\overline{B}} u_0 &
  y_1 &= \bm{\overline{C}} \bm{\overline{A}} \bm{\overline{B}} u_0 + \bm{\overline{C}} \bm{\overline{B}} u_1 &
  y_2 &= \bm{\overline{C}} \bm{\overline{A}}^2 \bm{\overline{B}} u_0 + \bm{\overline{C}} \bm{\overline{A}} \bm{\overline{B}} u_1 + \bm{\overline{C}} \bm{\overline{B}} u_2
  & \dots
\end{align*}
This can be vectorized into a convolution \eqref{eq:convolution} with an explicit formula for the convolution kernel \eqref{eq:krylov}.
\begin{equation}
  \label{eq:convolution}
  \begin{split}
    y_k &= \bm{\overline{C}} \bm{\overline{A}}^k \bm{\overline{B}} u_0 + \bm{\overline{C}} \bm{\overline{A}}^{k-1} \bm{\overline{B}} u_1 + \dots + \bm{\overline{C}} \bm{\overline{A}} \bm{\overline{B}} u_{k-1} + \bm{\overline{C}}\bm{\overline{B}} u_k
    \\
    y &= \bm{\overline{K}} \ast u %
    .
  \end{split}
\end{equation}
\begin{equation}%
  \label{eq:krylov}
  \bm{\overline{K}} \in \mathbb{R}^L :=
  \mathcal{K}_L(\bm{\overline{A}}, \bm{\overline{B}}, \bm{\overline{C}}) := \left(\bm{\overline{C}} \bm{\overline{A}}^i \bm{\overline{B}}\right)_{i \in [L]} = (\bm{\overline{C}}\bm{\overline{B}}, \bm{\overline{C}}\bm{\overline{A}}\bm{\overline{B}}, \dots, \bm{\overline{C}}\bm{\overline{A}}^{L-1}\bm{\overline{B}})
  .
\end{equation}
\normalsize
 \( \bm{\overline{K}} \) in equation \eqref{eq:convolution} can be represented as a single (non-circular) convolution which can be computed very efficiently with FFTs.However, computing \( \bm{\overline{K}} \) in \eqref{eq:krylov} is non-trivial and is modelled as a \( \bm{\overline{K}} \) the \textbf{SSM convolution kernel} or filter.

Specifically, we model input sequence using the state-of-the-art state space model Mamba\cite{gu2023mamba}. Mamba identifies a critical weakness in existing models: their inability to perform content-based reasoning.
To address this, Mamba introduces selective state spaces (SSMs) that allow the model to selectively propagate or forget information along the sequence length dimension based on the current token.  While we apply the Mamba block for the vision task, we face the problem of stability issues (loss convergence issue) compared to other models like S4 or Hippo.  We are providing one type of solution for the instability issue by preserving only negative eigenvalue. To perform this we need an extra module which we call the channel mixing, which was missing in the mamba block. We combine the channel mixing module with the Sequence mixing module and make a simplified Mamba Based Architecture (SiMBA) as shown in SiMBA\cite{patro2024simba}.  The input token sequence $\mathbf{X}_{\mathtt{l}-1}$ is first normalized by the normalization layer. Next, we linearly project the normalized sequence to the $\mathbf{x}$ and $\mathbf{z}$ with dimension size $E$. Then, we process the $\mathbf{x}$ from the forward and backward directions. 
For each direction, we first apply the 1-D convolution to the $\mathbf{x}$ and get the $\mathbf{x}'_{o}$. We then linearly project the $\mathbf{x}'_{o}$ to the $\mathbf{B}_{o}$, $\mathbf{C}_{o}$, $\mathbf{\Delta}_{o}$, respectively. 
The $\mathbf{\Delta}_{o}$ is then used to transform the $\overline{\mathbf{A}}_{o}$, $\overline{\mathbf{B}}_{o}$, respectively.

\section{Recent Advances in  State Space Models}\label{ssm}
Attention-based transformers have revolutionized natural language processing and other sequence-to-sequence tasks. However, they encounter certain limitations, especially when dealing with long input sequences, especially when dependencies extend beyond the attention window size. This constraint is particularly crucial in applications such as high-resolution imagery analysis and genomics. Efforts to address these limitations have been surveyed by Efficient 360 \cite{patro2023efficiency}, focusing on optimizing and improving efficiency in terms of computational complexity. Various aspects of transformers, including spectral analysis, fairness considerations, approximation methods, robustness enhancements, and computational complexity optimizations, have been discussed. In this report, we discuss these limitations and explore state space models (SSMs) as an alternative approach. 

\begin{itemize}
    \item \textbf{Computational Complexity:\cite{sanford2024representational,thangavel2023limitations}} Transformers exhibit high computational demands, particularly with large models. This complexity poses challenges for both training and deploying them on resource-constrained devices.

  \item \textbf{Large Memory Requirements:\cite{sanford2024representational,thangavel2023limitations}} Transformers necessitate significant memory resources for storing embeddings and intermediate activations. This can hinder scalability, especially for very long sequences, as they may surpass available memory capacity.

  \item \textbf{Fixed Sequence Length:\cite{sanford2024representational,thangavel2023limitations}} Transformers rely on fixed-size input sequences due to positional embeddings. Efficiently handling variable-length inputs presents a notable challenge in transformer-based architectures.

  \item \textbf{Attention Mechanism Scalability}

While attention is a powerful mechanism, it has a quadratic scaling with input sequence length. This makes it less efficient for very long sequences\cite{sanford2024representational, peng2024limitations,thangavel2023limitations}.

  \item \textbf{Lack of Causality in Standard Attention}

The standard self-attention mechanism used in transformers doesn't inherently capture causality. It treats all positions equally, which can be problematic for tasks where causality matters\cite{thangavel2023limitations}.

\end{itemize}

\begin{figure}
\centering
  \includegraphics[width=0.85\linewidth]{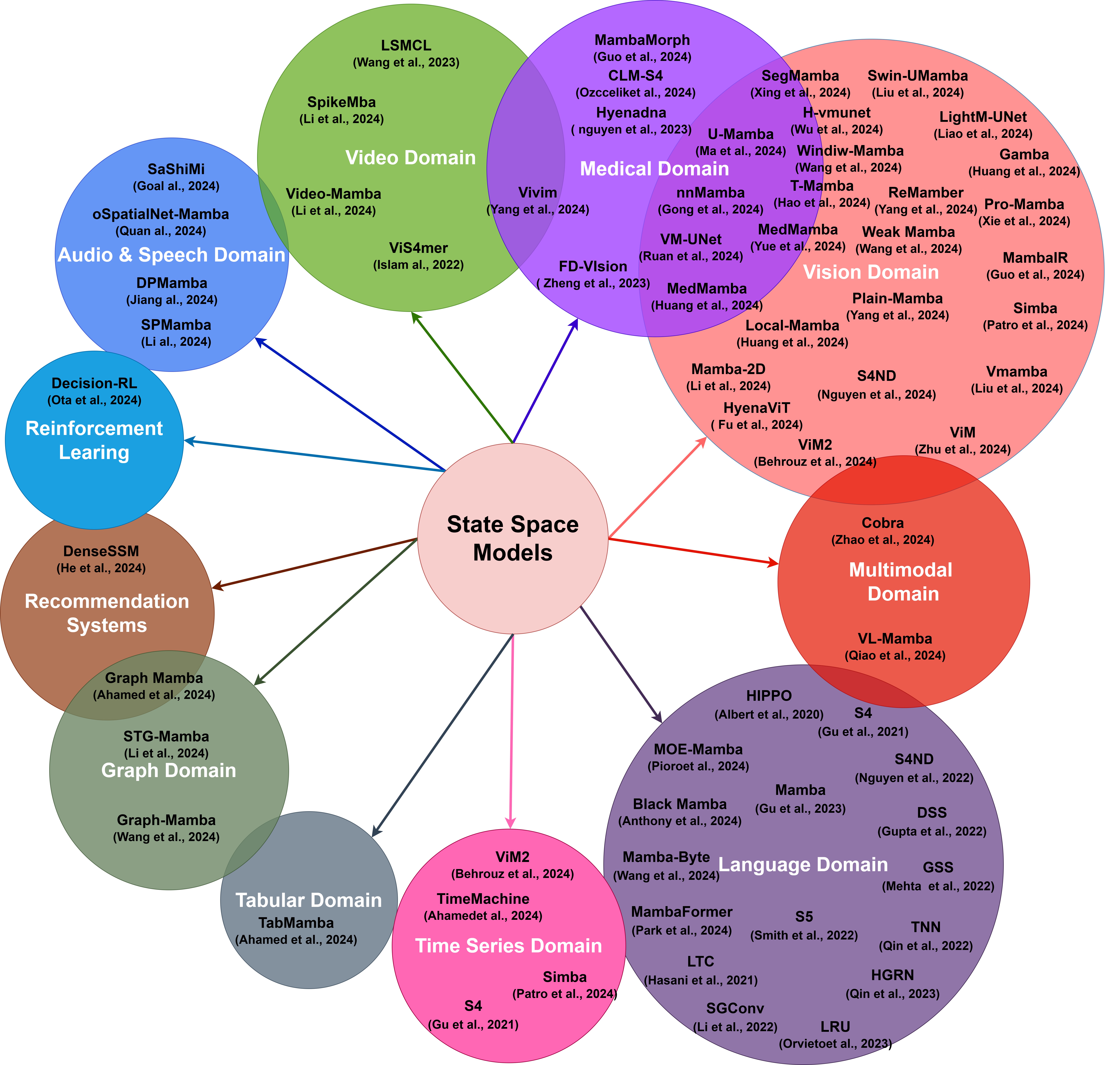}
  \caption{Application of State Space Models (SSMs) Across Various Domains.}
  \label{fig:SSM_categorization}
  % \vspace{-0.3in}
\end{figure}

Despite advancements, attention-based transformers still struggle with long sequences, leading to unsolved challenges in long-range benchmarks such as the path-X task. To address these limitations, State-space models(SSMs) offer a promising alternative approach to this issue, with pioneering models like S4 being among the first to effectively tackle the path-X problem. SSMs efficiently model long sequences while capturing long-term dependencies. In the following sections, we categorize and discuss key state space models from the literature. The categorization of key state space models is illustrated in Figure \ref{fig:SSM_categorization}:

\begin{itemize}
    \item \textbf{Structured SSMs}: These models, based on S4 and its variants, include Hippo, H3, HyenaHierarchy, Liquid-S4, S4nd, DSS, and Global convolutions and its variants including LongConv, FFTFlashConv and SG-Conv as well as certain foundational models such as LD-Stack and its derivative S5. They provide a principled way to handle long-range dependencies.
    
    \item \textbf{Recurrent SSMs}: These models are based on RNNs and their variants, such as RWKV, LRU, and HGRN, offering an alternative to attention-based approaches for sequence modeling.
    
    \item \textbf{Gated SSMs}: GSS, Mega, and TNN fall into this category, leveraging gating techniques to enhance performance on long sequences.
    
    \item \textbf{Miscellaneous SSMs}: MambaFormer, Mamba-Byte, and Mamba-MoE explore various techniques beyond the standard attention mechanism, combining ideas from different categories for efficient sequence modeling
\end{itemize}

It should however be noted that Mamba, for instance, is derived from both Hippo (a structured SSM) and is also incorporating gating technology. This is depicted in the diagram with arrows. Similarly, GSS, one of the foundational models in the structured category, is derived from DSS but also uses gating. Similarly, S5 is derived from one of the foundational models LDStack and S4.
State space models offer promising solutions for handling long sequences, and their efficiency and effectiveness make them valuable alternatives to attention-based transformers in certain scenarios.

\subsection{Structured State Space Models}
Structured State Space Models (SSMs) encompass various innovative approaches to sequence modeling, including S4, HiPPO, H3, and Liquid-S4. These models leverage sophisticated mechanisms such as polynomial projection operators, multi-input multi-output systems, and convolutional kernels to capture long-range dependencies efficiently. They demonstrate competitive performance across diverse benchmarks, showcasing their effectiveness in handling sequential data with improved computational efficiency.

 \subsubsection{Structured State Space Sequence (S4)\cite{gu2021efficiently}}
 S4\cite{gu2021efficiently} is a novel type of sequence model designed to capture long dependencies within sequences based on the state space model (SSM), typically continuous-time in nature. S4 introduces three important mechanisms: 1. Higher-Order Polynomial Project Operator (HiPPO): HiPPO operates on state and input transition matrices to memorize signal history effectively, enabling the model to capture long-term dependencies. 2. Diagonal Plus Low-Rank Parametrization: S4 conditions the SSM matrix (A) with a low-rank correction to stabilize it, ensuring diagonalizability and stability. 3. Efficient (convolutional) Kernel Computation: S4 leverages FFTs and iFFTs for efficient computation of transition matrices, reducing the overall complexity to $O(N\log(N))$. S4 introduces a new parameterization for the SSM which conditions matrix (A) with a low-rank correction, making it stable and diagonalizable. This transformation reduces the SSM computation to a well-studied Cauchy kernel. As a result, S4 models achieve strong empirical results while being computationally efficient. S4 was the first state space model to solve the path-X task in the LRA benchmark and reduces the computational complexity to $O(N\log(N))$. S4 performs exceptionally well across various benchmarks. It achieves 91\% accuracy on sequential CIFAR-10 without data augmentation or auxiliary losses. S4 closes the gap to transformers on image and language modeling tasks while being faster. Subsequent efforts, including Hippo and Long Convolutions, aimed to enhance state space models' efficiency but demonstrated a performance gap compared to state-of-the-art transformers.

\subsubsection{High-Order Polynomial Projection Operators (HiPPO)\cite{albert2020hippo}} HiPPO\cite{albert2020hippo} are applied to state and input transition matrices to effectively memorize signals' history. It was observed that there was no known mathematical interpretation of the specific matrix used in S4, originally defined for time-variant dynamical systems but employed in time-invariant state space models (SSMs). Hippo \cite{albert2020hippo} provided a mathematical interpretation of S4 as exponentially warped Legendre polynomials, explaining S4's capacity to capture long-range dependencies. The HiPPO framework is an intriguing concept that combines state space models (SSMs) with generalized orthogonal basis projections. Hippo comprises four variants, with one using the truncated Fourier basis polynomial known as Hippo-FouT, while the second variant uses LagT, based on Lagurre polynomials. The third variant uses LegT which is based on Legendre polynomials and the fourth is LegS, which uses Legendre polynomials with a sliding window. Hippo significantly enhances the S4's performance to 86\% on the long-range arena benchmark and 96\% on the most difficult task like path-X.

\subsubsection{Hungry Hungry HiPPO (H3) ~\cite{fu2022hungry}}
H3 ~\cite{fu2022hungry} identified two key challenges that prior state space models (SSMs) faced: one is difficulty in recalling earlier tokens,i.e,  traditional SSMs struggled to effectively retain information from earlier tokens within a sequence and the second is difficult in comparing the tokens across different sequences.  To overcome these limitations, H3 introduces a novel approach using three key components of their method, Stacked SSMs with Multiplicative Interactions, FlashConv for Training Efficiency, and a State-Passing Algorithm for Scaling. H3 employs a novel approach by stacking two SSMs with multiplicative interactions between their input and output projections. This design allows for enhanced memory retention and cross-sequence comparisons.  Additionally, H3 introduces FlashConv, a method aimed at enhancing training efficiency on modern hardware. FlashConv utilizes a fused block Fast Fourier Transform (FFT) algorithm specifically designed for sequences up to 8K in length. Moreover, to scale SSMs beyond the 8K sequence length limit, H3 incorporates a state-passing algorithm. This algorithm effectively splits the input into the largest possible chunks that can fit within the SRAM memory of the GPU. Despite these advancements, there remains a perplexity gap between H3 and transformers (with 1.3B parameters). However, H3 exhibits superior performance over transformers in zero-shot and few-shot learning scenarios, particularly evident on superglue benchmark tasks. Impressively, H3 achieves a 2× speedup on the long-range arena benchmark and enables hybrid language models to generate text 2.4× faster than transformers.

\subsubsection{Global Convolution}
Long Convolution \cite{fu2023simple} authors conjecture that SSMs rely on sophisticated mathematical structures to train effectively in deep networks. They generate a convolutional kernel as long as the input sequence, by repeatedly multiplying a hidden state matrix. This process leads to instability and needs hand-crafted initializations and hyper-parameter tuning.  To address these challenges, the authors introduce the main contribution of parametrizing long convolutional kernels directly. Typical implementations of long convolutions utilize Fast Fourier Transform (FFT), which can sometimes be slower than quadratic attention due to certain system constraints.  Long Convolutions solve these problems by using simple techniques like regularization (smoothening and squashing) and an IO-aware algorithm known as FlashButterfly. Impressively, Long Convolution outperforms transformers on the WikiText103 benchmark by reducing perplexity by 0.2 with 30\% fewer parameters.
Furthermore, Long Convolution claims to be 7.2 times faster than transformers on the LRA speed benchmark. Another related effort is FlashFFTConv \cite{fu2023flashfftconv}, which utilizes two sparse convolutional algorithms (frequency-sparse convolution and partial convolution). FlashFFTConv employs a matrix decomposition technique that computes FFT using matrix multiplication, enabling kernel fusion for long sequences by reducing I/O operations. Similar approaches include Structural Global Convolution (SGConv) \cite{Li2022WhatMC} and Hyena Hierarchy \cite{poli2023hyena}.

\subsubsection{Hyena Hierarchy (HH)\cite{poli2023hyena}}
HH \cite{poli2023hyena} is a state space model (SSM) proposed to address the perplexity gap with attention-based transformers. Attention mechanisms incur quadratic costs in sequence length, limiting the accessibility of context. While existing methods employ sub-quadratic layers based on low-rank and sparse approximations, they often require dense attention layers to match the performance of transformers. HH addresses this by demonstrating that previous sub-quadratic attention alternatives exhibit a perplexity gap with attention, particularly evident in long text sequences. Hyena employs a sub-quadratic approach that replaces attention with implicitly parametrized long convolutions and data-gating. HH has shown that the tipping point between HH and attention occurs at approximately 6K sequence length - at 100K sequence length, HH is 100 times faster than Flash attention, a highly optimized form of attention. Additionally, HH has nearly closed the performance gap with transformers on the pile benchmark, showcasing its effectiveness in handling long sequences efficiently while maintaining competitive performance with attention-based models.

\subsubsection{RWKV \cite{peng2023rwkv}}
RWKV \cite{peng2023rwkv} is a recent recurrent neural network (RNN) designed for language modeling, incorporating a linear attention approximation mechanism known as the "WKV" mechanism. RWKV utilizes LTI (linear time-invariant) recurrences and is conceptualized as the ratio of two state space models (SSMs). RWKV is based on a linear attention approximation, which differs from the traditional self-attention mechanism used in transformers.
The authors of RWKV observe that while transformers have revolutionized natural language processing (NLP) and various other domains, they suffer from quadratic computational and memory complexity. In contrast, RNNs offer linear scaling in both memory and computational complexity but face performance gaps compared to transformers. However, RNNs often lag behind transformers in terms of parallelization and scalability. RWKV claims to be a hybrid model combining the strengths of both RNNs and transformers. However, it's essential to note that, to the best of our knowledge, RWKV is essentially a transformer with a linear attention mechanism and does not possess recurrent properties typical of traditional RNNs.

\subsubsection{LDStack \cite{kaul2020linear}}
LDStack \cite{kaul2020linear} demonstrates how a recurrent neural network (RNN) can be represented as a multiple-input multiple-output (MIMO) Linear Dynamical System (LDS). LDStack demonstrates that an RNN can be effectively represented as an LDS. This viewpoint allows for a deeper understanding of RNN behavior and properties. Specifically, they consider the RNN as a system with multiple inputs and multiple outputs, similar to an LDS. To address computational challenges, LDStack introduces a technique called parallel scan. This involves approximating the MIMO LDS by aggregating it into a Single Input Multiple Outputs (SIMO) LDS. This approximation simplifies computations while preserving essential characteristics. The authors highlight that many discrete-time LDS can be modeled using state space equations. This implies that LDS can be viewed as equivalent to time-varying state space models \footnote{For reference, any dynamical system can be represented using state space equations of the form $x(t+1) = Ax(t) + Bu(t)$ and $y(t) = Cx(t) + Du(t)$, as illustrated in \url{https://inst.eecs.berkeley.edu/~ee127/sp21/livebook/l_mats_apps_state.html}}. LDStack's approach has demonstrated state-of-the-art performance, particularly in copy memory tasks \cite{arjovsky2016unitary}. This equivalence provides a bridge between the RNN framework and the more formal state space modeling. LDStack achieves state-of-the-art performance in copy memory problems, showcasing the effectiveness of their approach. Notably, this work contributes to the broader understanding of RNNs and their connections to linear dynamical systems. LDStack offers valuable insights into the relationship between RNNs and LDSs, opening up new avenues for research and practical applications.

\subsubsection{S5 \cite{smith2022simplified}}

S5 \cite{smith2022simplified} extends the principles of LDStack, which modeled RNNs as multiple-input multiple-output Linear Dynamical Systems (LDS), to state space models (SSMs), offering a generalized approach. Unlike LDStack, which handles independent SSMs sequentially, the S5 layer processes multiple inputs and outputs concurrently. Performance evaluations indicate that S5 achieves notable results, reaching 87.2\% accuracy on the Long-Range Arena (LRA) benchmark and an impressive 98.5\% accuracy on the path-X task within LRA. While S4 comprises numerous independent single-input, single-output SSMs utilizing the HiPPO framework, the S5 layer adopts a single multi-input, multi-output SSM, facilitating parallelization. This parallel processing significantly enhances computational efficiency. S5 represents an intriguing fusion of ideas from LDS-based modeling and state space models, emphasizing parallel processing and computational efficiency.

\subsubsection{S4nd\cite{nguyen2022s4nd}}
The primary goal of S4nd\cite{nguyen2022s4nd}  is to extend the applicability of SSMs beyond sequential data (such as Text data and time series) to continuous data domains. S4nd \cite{nguyen2022s4nd} addresses the challenge of applying state space models (SSMs) to continuous data such as images or videos.  S4nd contributes a new deep learning layer that transforms a standard SSM, which typically represents a 1-dimensional Ordinary Differential Equation (ODE), into a multi-dimensional Partial Differential Equation (PDE). This transformation allows SSMs to capture spatial dependencies across multiple dimensions. Moreover, S4nd demonstrates that a multi-dimensional SSM can be equivalently represented as ND continuous convolution, wherein each dimension undergoes 1D SSM convolution. To evaluate its efficacy, S4nd was tested on the ImageNet dataset using ConvNeXt as the backbone model, resulting in improved performance over traditional methods. Notably, it improved the performance of Vision Transformers (ViTs). Additionally, S4nd was found to outperform ConvNeXt on video recognition tasks, particularly on datasets such as HMDB \cite{kuehne2011hmdb}. S4nd extends the capabilities of SSMs to handle continuous data, bridging the gap between sequential and spatial modeling.

% Diagonal State Spaces (DSS)~\cite{gupta2022diagonal}: S4 can be viewed as a parametrization of state space matrices via diagonal plus low-rank correction. DSS~\cite{gupta2022diagonal} proved that they can match S4 performance by using only the diagonal state space matrices, without the need for low-rank correction. DSS simply removes the low-rank correction component of the Hippo matrix and still matches S4 performance on raw speech classification benchmarks as well as LRA. A DSS variant \cite{gu2022parameterization} outlines how specific initializations of the state space matrix based on approximating S4's matrix, can lead to good performance on long sequence modeling. 

\subsubsection{Diagonal State Spaces (DSS) \cite{gupta2022diagonal}}
DSS \cite{gupta2022diagonal} explores the parametrization of state space matrices through diagonal matrices plus low-rank correction. Remarkably, DSS demonstrates that it can achieve performance comparable to S4 without the need for low-rank correction. By solely utilizing diagonal state space matrices, DSS matches S4's performance on various benchmarks, including raw speech classification and the Long-Range Arena (LRA) benchmark. Furthermore, a variant of DSS proposed in \cite{gu2022parameterization} provides insights into specific initializations of the state space matrix. These initializations, based on approximating S4's matrix, result in effective performance for long sequence modeling tasks. This variant showcases the flexibility and potential of DSS in achieving strong performance across different applications and scenarios. The key innovation of DSS lies in removing the low-rank correction component from the Hippo matrix (used in S4).

\subsubsection{Liquid Structural State Spaces (Liquid-S4) \cite{hasani2022liquid}}
Liquid time-constant networks (LTCs) \cite{hasani2021liquid} are causal continuous-time neural networks characterized by input-specific state transitions.  Integrating LTCs with S4, researchers have introduced Liquid-S4 \cite{hasani2022liquid}, given by the following dynamics:$$ \dot{x} = (\bm{A} +\bm{B} u) x +\bm{B} u, y = \bm{C}x$$  which is different from S4 dynamics,  $$ \dot{x} = \bm{A} x +\bm{B} u, y = \bm{C}x$$.  The primary motivation behind Liquid-S4 is to enhance SSMs by leveraging the properties of LTCs. Liquid-S4 constructs a convolutional kernel based on a linearized version of LTCs. This kernel structure considers similarities among input sequence samples during both training and inference. Liquid-S4 introduces an input-dependent state-transition module, enabling adaptation to varying inputs. Additionally, Liquid-S4 constructs a convolutional kernel corresponding to a linearized version of LTC.  Liquid-S4 achieves state-of-the-art generalization across various sequence modeling tasks with long-term dependencies (including image, text, audio, and medical time series). Notably, Liquid-S4 achieves an average performance of 87.32\% on the Long-Range Arena (LRA) benchmark, with only Gated State Space models (GSS) achieving comparable results. Liquid-S4 outperforms ConvNeXt on the **raw Speech Command recognition dataset**, achieving 96.78\% accuracy with a 30\% reduction in parameter counts compared to S4. Furthermore, a Convolutional Representation of Liquid-SSMs is provided, similar to the formulation presented in (\ref{eq:convolution}). The Liquid-SSM is first unrolled in time to construct a convolutional kernel of it. By assuming $x_{-1} = 0$ :
\begingroup
\footnotesize
\begin{align}
    \label{eq:unroll_liquid_ssm} \nonumber
     x_0 &= \overline{\textbf{B}} u_0,~~~~~y_0 = \overline{\textbf{C}}\overline{\textbf{B}} u_0 \\
     x_1 &= \overline{\textbf{A}}\overline{\textbf{B}} u_0 + \overline{\textbf{B}} u_1 {\color{blue} +~ \overline{\textbf{B}}^2 u_0 u_1},~~~~~y_1 = \overline{\textbf{C}}\overline{\textbf{A}}\overline{\textbf{B}} u_0 + \overline{\textbf{C}}\overline{\textbf{B}} u_1 {\color{blue} + \overline{\textbf{C}}\overline{\textbf{B}}^2 u_0 u_1} \\ \nonumber
     x_2 &= \overline{\textbf{A}}^2\overline{\textbf{B}} u_0 + \overline{\textbf{A}}\overline{\textbf{B}} u_1 + \overline{\textbf{B}} u_2 {\color{blue} +~ \overline{\textbf{A}}\overline{\textbf{B}}^2 u_0 u_1 +~ \overline{\textbf{A}}\overline{\textbf{B}}^2 u_0 u_2 +~ \overline{\textbf{B}}^2 u_1 u_2 +~ \overline{\textbf{B}}^3 u_0 u_1 u_2 } \\ \nonumber
     y_2 &= \overline{\textbf{C}}\overline{\textbf{A}}^2\overline{\textbf{B}} u_0 + \overline{\textbf{C}}\overline{\textbf{A}}\overline{\textbf{B}} u_1 + \overline{\textbf{C}}\overline{\textbf{B}} u_2 {\color{blue} +~ \overline{\textbf{C}}\overline{\textbf{A}}\overline{\textbf{B}}^2 u_0 u_1 +~ \overline{\textbf{C}}\overline{\textbf{A}}\overline{\textbf{B}}^2 u_0 u_2 +~ \overline{\textbf{C}}\overline{\textbf{B}}^2 u_1 u_2 +~ \overline{\textbf{C}}\overline{\textbf{B}}^3 u_0 u_1 u_2 },~~ \dots \nonumber
\end{align}
\endgroup
Liquid-S4 combines the strengths of LTCs and SSMs, resulting in improved performance and adaptability across diverse sequence modeling tasks.

\subsubsection{State Space Augmented Transformer (SPADE) \cite{zuo2022efficient}}
The  SPADE\cite{zuo2022efficient} addresses the challenge of efficiently capturing both global and local information from long sequences. While attention optimizations in transformers enhance computational efficiency, they often struggle to capture global context from long sequences effectively. On the other hand, state space models (SSMs) excel at capturing global information from long sequences but may not efficiently capture local dependencies. SPADE proposes a novel approach by integrating an SSM, specifically S4, as the bottom layer of a transformer. This integration allows SPADE to leverage the strengths of both SSMs and transformers, efficiently capturing global information. Additionally, SPADE incorporates a window and chunk-based local attention layers to effectively capture local dependencies. By combining global information captured by the SSM layer with local information obtained from attention mechanisms, SPADE achieves significant performance gains across various benchmarks, including the Long-Range Arena (LRA), WikiText, and Glue benchmarks. This integration of SSMs and transformers in SPADE represents a promising direction for enhancing the performance and efficiency of sequence modeling tasks.

\subsection{Gated SSMs}
Gated State Spaces (GSS), Toeplitz Neural Network (TNN), and Mamba represent innovative approaches within the realm of Gated SSMs. GSS leverages gating units to optimize FFT operations, achieving efficient sequence processing and competitive performance. TNN introduces a position-encoded Toeplitz matrix for token mixing, significantly reducing space-time complexity while maintaining state-of-the-art results. Mamba addresses computational inefficiencies in traditional SSMs by incorporating a gated MLP and hardware-aware algorithms, demonstrating linear-time complexity and improved efficiency over conventional Transformers.

\subsubsection{Mega \cite{ma2022mega}}
Mega \cite{ma2022mega} identifies two deficiencies in transformers: weak inductive bias and quadratic attention complexity  with respect to the sequence length. The weak inductive bias refers to transformers' lack of assumptions regarding the patterns of interaction or dependencies between tokens.  It treats all positions equally, which can be suboptimal for modeling long sequences. To address this, Mega introduces a novel mechanism called Moving Average Equipped-Gated Attention. The gated attention mechanism of Mega employs a single-head gated attention mechanism. This mechanism integrates classical Exponential Moving Average (EMA) with single-headed gated attention, effectively incorporating positional-level local dependency into positional-agnostic attention.  By doing so, it introduces a more meaningful inductive bias into the attention mechanism. Additionally, Mega offers a variant known as Mega-Chunk, which addresses the quadratic attention complexity by introducing linear space and time complexity while minimizing quality loss. It achieves this by efficiently dividing the entire sequence into fixed-length chunks. By leveraging the strengths of EMA and gated attention, Mega enhances the inductive bias of transformers and mitigates the computational complexity associated with attention mechanisms. These innovations represent significant advancements toward more efficient and effective sequence modeling. The experiments conducted by Ma et al. demonstrate that Mega outperforms other sequence models, including various Transformer variants, on a wide range of benchmarks, such as neural machine translation, language modeling, and image and speech classification.

\subsubsection{Gated State Spaces (GSS) \cite{mehta2022long}}
GSS \cite{mehta2022long} builds upon the prior work of Lia\cite{liu2021pay} et al., who observed that replacing the MLP layer of the transformer with gating units can reduce dimensionality during token mixing. Expanding on this insight, GSS introduces a gating layer designed to decrease dimensionality during Fast Fourier Transform (FFT) operations in state space models (SSMs). This enables GSS to operate efficiently on longer input sequences.  GSS exhibits zero-shot generalization to longer input sequences while maintaining straightforward implementation. Furthermore, GSS provides additional insights by initializing the state space variables to random values. Surprisingly, GSS outperforms models like Hippo, S4, and DSS, which rely on special initializations using linear algebra techniques (e.g., Hippo matrix initializations). Additionally, GSS reduces the perplexity gap with Block Recurrent Transformers \cite{hutchins2022block} across various benchmarks such as PG-19, Arxiv, and GitHub, while also exhibiting higher efficiency. These findings highlight the effectiveness and versatility of GSS in sequence modeling tasks.  GSS trains significantly faster than the diagonal version of S4 (DSS) on Tensor Processing Units (TPUs). It achieves competitive performance with several well-tuned Transformer-based baselines.

\subsubsection{Toeplitz Neural Network (TNN)\cite{qin2022toeplitz}}
TNN \cite{qin2022toeplitz} has been recently proposed to address the two important aspects of transformers, namely the attention mechanism, which learns pair-wise correlations of input tokens and second, the positional embedding, which learns positional inductive bias. TNN uses a position-encoded Toeplitz matrix to capture relations between input token pairs as a token mixer and uses a Toeplitz matrix-vector production technique to reduce space-time complexity to O(N*logN). The authors claim that TNN is more efficient than transformers in that TNN is log-linear complexity, while transformers are quadratic. A key feature of TNN is its use of a Relative Position Encoder (RPE) to generate relative positional coefficients with a fixed parameter budget. RPE uses relative positional parameters to ensemble the Toeplitz matrix and makes TNN's parameters independent of sequence length. TNN consists of a Gates Toeplitz Unit (GTU) and the Gated Linear Unit (GLU).  The GTU employs a Toeplitz Neural Operator (TNO) for token mixing using the Toeplitz matrix, while the RPE utilizes a fully connected network encoded with position information to generate relative position coefficients. Empirical evaluations demonstrate that TNN achieves state-of-the-art performance on benchmarks such as Glue and Long-Range Arena (LRA). These findings underscore the efficacy and efficiency of TNN in comparison to traditional transformer architectures.

\subsubsection{Mamba \cite{gu2023mamba}}
Mamba \cite{gu2023mamba} highlights the quadratic computational and memory complexity of transformers, along with the perplexity gap that traditional state space models (SSMs) exhibit in comparison. Prior to Mamba, SSMs faced inefficiencies in addressing tasks such as selective copying and induction head \cite{olsson2022context}. However, observations on Liquid-S4's input-dependent state-transition module suggest its potential for solving these tasks effectively. Mamba addresses these challenges by introducing a novel parametrization approach for SSMs based on input characteristics and incorporating a simple selection mechanism. Additionally, Mamba presents an efficient hardware-aware algorithm based on selective scan. Similar to Gated State Spaces (GSS), Mamba employs a gated technique to reduce the dimensionality of global kernel operations. Moreover, Mamba combines a gated MLP\cite{liu2021pay} with the SSM module to further enhance its capabilities. Mamba’s linear-time complexity makes it more efficient than traditional Transformers.

\subsection {Recurrent SSMs}
Recurrent SSMS encompass the Linear Recurrent Unit (LRU) and the Hierarchically Gated Recurrent Neural Network (HGRN). The LRU, along with its extensions Griffin and Hawk, highlights the effectiveness of linear recurrence, MLP blocks, and attention mechanisms in enhancing long-sequence modeling. Conversely, the HGRN introduces dynamic forget gates to linear RNNs, leading to significant efficiency improvements and competitive performance across a range of benchmarks.

\subsubsection{Linear Recurrent Unit (LRU)\cite{orvieto2023resurrecting}}
LRU \cite{orvieto2023resurrecting} observes that while SSMs achieve good performance on long sequence modeling tasks, the exact reasoning behind the same is unclear. The initial reasoning behind good S4 performance, for instance, was the specific initialization(HiPPO\cite{albert2020hippo}) of the state matrices as well as the discretization of the continuous-time system of differential equations. However, as we have noted above, the GSS~\cite{mehta2022long}, and DSS ~\cite{gupta2022diagonal} have shown that the specific initializations mentioned above are not crucial, even random initializations can lead to significant performance gains. So,LRU takes a fresh approach by starting with Recurrent Neural Networks (RNNs) and exploring whether their performance can match that of SSMs in long sequence modeling tasks. They make several improvements to RNNs, including 1. Linear Recurrence: the removal of non-linearities in the recurrence and stacking up linear RNN layer with non-linear MLP blocks. 2. Parametrization: parametrizing the RNNs to a complex diagonal form, which allows for parallelization of training, inspired by DSS and S5. 3. Stable Exponential Parametrization: stable exponential parametrization of the diagonal recurrent matrix, which leads to simplifying the training process 4. Normalization: normalization of hidden activations on the forward pass, which enables training time stability enforcement. These improvements enable LRU to achieve performance levels comparable to other SSMs and transformers on benchmarks such as the Long-Range Arena (LRA). By combining insights from both traditional SSMs and neural network architectures, LRU sheds light on new avenues for improving long-sequence modeling tasks.

Griffin and Hawk, both developed by the same authors \cite{de2024griffin}, extend the capabilities of the Linear Recurrent Unit (LRU) by incorporating additional layers and mechanisms for improved sequence modeling. In Griffin, the authors interleave LRU layers with Multilayer Perceptron (MLP) blocks and local attention mechanisms \cite{beltagy2020longformer}. This architecture utilizes a residual block, an MLP block, and a temporal-mixing block. The temporal-mixing block is validated with three alternatives: 1) global multi-query attention \cite{shazeer2019fast}, 2) local (sliding window) attention, and 3) LRU-based recurrent blocks. On the other hand, Hawk, another model from the same authors, takes a slightly different approach by interleaving LRU layers with MLP blocks. Similar to Griffin, Hawk also incorporates a temporal-mixing block, which offers alternatives such as global multi-query attention, local attention, and LRU-based recurrent blocks. Both Griffin and Hawk undergo performance comparisons with LLama 2 and Mamba, demonstrating their effectiveness in sequence modeling tasks. These models showcase the potential of combining recurrent and attention mechanisms for improved long-sequence modeling.

\subsubsection{Hierarchically Gated Recurrent Neural Network (HGRN) \cite{qin2024hierarchically}}
HGRN \cite{qin2024hierarchically} is a recent effort in modifying RNNs that uses gated linear RNNs with forget gates that have learnable weights that propagate from the lower layer to the upper layers. This enables HGRNs to handle short-term dependencies, which are more local to the model, in the lower layers. HGRNs handle long-term dependencies in the upper layer, which captures global information. HGRNs address the shortcomings of RNNs, namely, the complexity of hidden state updates, which typically involves full matrix multiplication and the presence of non-linearity within the recurrent unit, that prevents parallel computation. HGRNs use an element-wise linear recurrent layer, which removes non-linearity in the recurrence, leading to parallelized training. HGRNs use element-wise multiplication for hidden state updates. Linear RNNs have used two methods, namely the Exponential Moving Average (EMA) and gating schemes. It must be noted that typical state space models such as S4, RWKV, S4nd, Mega, and LRU all use EMA methods, where decay rates are data independent, which implies they are static in all time steps. HGRN\cite{qin2024hierarchically} and Liquid-S4 \cite{hasani2022liquid} are the only models to use data-dependent or dynamic decay rates. HGRN uses forget gates to achieve these dynamic decay rates, while Liquid-S4 uses a dynamics transition matrix, which is a limited form of FFT. HGRN achieves significant performance results on Wikitext data, Glue, Long Range Arena, and the Pile benchmarks. However, while HGRN gains significant efficiencies compared to other SSMs and transformers, HGRNs overcome the perplexity gap with attention-based transformers with only TNN achieving lower perplexity scores. 

\subsection{Miscellaneous SSMs}

\subsubsection{Mixture of Experts (MoE)}
MoE has emerged as a prominent approach to enhance the performance of Large Language Models (LLMs). Several Efforts have been made to integrate MoE with state space models, leading to innovations such as BlackMamba \cite{anthony2024blackmamba}, MoE-Mamba \cite{pioro2024moe}, and Jamba \cite{lieber2024jamba}. BlackMamba innovatively replaces the self-attention mechanism of transformer architecture with Mamba State Space Models (SSMs)\cite{gu2023mamba} for sequence mixing. Additionally, it combines this with a MoE-Transformer architecture, where multiple Multilayer Perceptrons (MLPs) are employed for channel mixing. The MoE component in BlackMamba enables the activation of only one sparse subset of parameters in a single forward pass. Moreover, BlackMamba incorporates a router mechanism that dynamically learns to route tokens to the appropriate expert. The model has demonstrated promising results in lowering the perplexity gap observed with attention-based transformers. Similarly, MoE-Mamba follows a similar architecture to BlackMamba, leveraging the combination of Mamba SSMs and MoE-Transformer architecture. This model also showcases competitive performance, narrowing the perplexity gap with attention-based transformers. On the other hand, Jamba adopts a hybrid architecture, interleaving Mamba and transformer layers while integrating MoE with select layers. Jamba has been specifically tailored for common sense reasoning tasks and has demonstrated competitive performance, even outperforming Mixtral-8X-7B on certain benchmarks. Additionally, Jamba supports input token lengths of up to 256K, showcasing its scalability and versatility in handling long sequences. These advancements highlight the effectiveness of integrating MoE with state space models, paving the way for improved performance and scalability in large language models.

\textbf{MambaByte \cite{wang2024mambabyte}}
MambaByte \cite{wang2024mambabyte} represents a recent endeavor aimed at introducing hardware-efficient Mamba architectures tailored for efficient sequence processing. In MambaByte, the Mamba State Space Model (SSM) is utilized alongside a fixed-size memory state and optimized decoding techniques. The rationale behind MambaByte stems from the observation that Mamba possesses a substantial fixed-size memory state, akin to the hidden state in Recurrent Neural Networks (RNNs). This memory state remains independent of the context length, enabling efficient processing of long sequences. Furthermore, MambaByte incorporates a speculative decoding algorithm specifically tailored for byte-level models. Through experiments, MambaByte demonstrates superior performance compared to Mamba, particularly on long sequences spanning up to 524K in length. Additionally, MambaByte showcases its competitiveness against transformer models such as MegaByte \cite{yu2024megabyte} on the PG19 dataset \cite{rae2019compressive}, an open vocabulary language modeling benchmark derived from books. These advancements in MambaByte highlight its efficacy in achieving lower perplexity and improved performance, particularly in processing long sequences, thus contributing to the evolution of hardware-efficient Mamba architectures.

%In-context learning refers to a model’s ability to condition on a prompt sequence that includes in-context examples (input-output pairs related to a specific task) along with a new query input. The model then generates the corresponding output.
\subsubsection {SSMs and In-Context Learning}

In the exploration of in-context learning (ICL), the focus lies on the relationship between successful task performance and the information present within the training data. A significant inquiry revolves around whether models can be trained to effectively engage in in-context learning of specific function classes, such as linear functions, utilizing data derived from instances within that class.  Recent studies have discussed this question, particularly scrutinizing the capabilities of both standard Transformers and specialized architectures like Mamba. For instance, Garg et al. \cite{garg2022can} have demonstrated that standard Transformers can indeed be trained to adeptly engage in in-context learning of linear functions. Importantly, this learning occurs exclusively at inference time, without any parameter updates during training. Similarly, Grazzi et al. \cite{grazzi2024mamba} have investigated the feasibility of in-context learning with Mamba. Their findings reveal that Mamba exhibits comparable performance to Transformers in in-context learning tasks across standard scenarios such as skewed linear regression, ReLU neural network, and decision trees. Furthermore, the paper on MambaFormer \cite{park2024can} extends this exploration, showcasing the model's adeptness in in-context learning across additional tasks, including Vector-valued MQAR. Notably, MambaFormer demonstrates success where Mamba alone may falter, although it's important to note that the computational complexity of MambaFormer remains quadratic due to the integration of attention mechanisms with Mamba in each layer. These investigations collectively shed light on the potential of both Transformers and specialized architectures like Mamba and MambaFormer in engaging in effective in-context learning, contributing to a deeper understanding of their capabilities and limitations in this domain.

% The trained model is able to learn unseen linear functions from in-context examples with performance comparable to the optimal least squares estimator. Remarkably, in-context learning remains effective even under two forms of distribution shift:1)
% Between the training data and inference-time prompts. 2)
% Between the in-context examples and the query input during inference.

\begin{table*}[hbt]
%{r}{0.5\textwidth}
    \centering
    % \label{lm}
    \small  
    \caption{\textbf{Results on Wikitext-103\cite{merity2016pointer}} . $\downarrow$ means \textit{lower is better}.}
     \setlength{\tabcolsep}{1.1cm}
     \label{tab:wikitext}
    \begin{tabular}{l|lll}
    
    \hline
        Model & \makecell[c]{PPL \\(val)$\downarrow$} & \makecell[c]{PPL \\(test)$\downarrow$} &  \makecell[c]{Params\\(M)} \\ \hline
        \textit{Attn-based}  \\ \hline
        Transformer~\cite{vaswani2017attention} & 24.40 & {24.78} & 44.65 \\ 
        FLASH~\cite{hua2022transformer} & 25.92 & 26.70 & 42.17 \\ 
        1+elu~\cite{katharopoulos2020transformers} & 27.44 & 28.05 & 44.65 \\ 
        Performer~\cite{choromanski2020rethinking} & 62.50 & 63.16 & 44.65 \\
        cosFormer~\cite{qin2021cosformer} & 26.53 & 27.06 & 44.65 \\ \hline
        \textit{MLP-based}  \\ \hline
        Syn(D)~\cite{tay2021synthesizer} & 31.31 & 32.43 & 46.75 \\ 
        Syn(R)~\cite{tay2021synthesizer} & 33.68 & 34.78 & 44.65 \\ 
        gMLP\cite{liu2021pay} & 28.08 & 29.13 & 47.83 \\ \hline
        \textit{SSM-based}  \\ \hline
        S4~\cite{gu2021efficiently} & 38.34 & 39.66 & 45.69 \\ 
        DSS~\cite{gupta2022diagonal} & 39.39 & 41.07 & 45.73 \\ 
        GSS~\cite{mehta2022long} & 29.61 & 30.74 & 43.84 \\ 
        RWKV~\cite{peng2023rwkv} & 24.31 & 25.07 & 46.23\\
        LRU~\cite{orvieto2023resurrecting} & 29.86 & 31.12 & 46.24\\
        
        HGRN~\cite{qin2023hierarchically} & {24.14} & 24.82 &46.25 \\ 
        TNN~\cite{qin2022toeplitz} & {23.98} & {24.67} & 48.68 \\ 
        \hline
    \end{tabular}
\end{table*}

\begin{table*}[htb]
    \footnotesize
    \setlength{\tabcolsep}{0.28cm}
    \centering
    \caption{\textbf{Comparison of sequence modeling performance on the GLUE benchmark. MNLI results are reported separately for the match and mismatch splits. MRPC performance is measured using the F1 score, CoLA using the Matthews correlation coefficient, and all other tasks using accuracy. The best result is highlighted in **bold**, and the second-best is underlined. "-" denotes unconverted values. "Attn" refers to Attention models, "SSM" to State space models, "Trans" to Transformer models, and "LS" to Transformer-LS models.}}
    \begin{tabular}{lllllllll}
    \hline
        Method & MNLI &  QNLI &  QQP &  SST-2 &  MRPC &  CoLA & AVG & Params(m) \\ \hline
        \textit{Attn-based} & ~ & ~ & ~ & ~ & ~ & ~ & ~ & ~ \\ \hline
        Transr~\cite{vaswani2017attention} & 79.37/79.07 & 87.79  & 88.04  & 90.25  & 88.35  & 38.63  & \textbf{78.79}  & 124.70 \\
       LS~\cite{zhu2021long} & 77.01/76.78 & 84.86  & 86.85  & 90.25  & 82.65  & 40.65  & 77.01  & 128.28 \\
        FLASH~\cite{hua2022transformer} & 79.45/80.08 & 87.10  & 88.83  & 90.71  & 82.50  & 29.40  & 76.87  & 127.12 \\
        1+elu~\cite{katharopoulos2020transformers} & 74.87/75.37 & 82.59  & 86.90  & 87.27  & 83.03  & -  & 70.00  & 124.70 \\
        Performer~\cite{choromanski2020rethinking} & 58.85/59.52 & 63.44  & 79.10  & 81.42  & 82.11  & 19.41  & 63.41  & 124.70 \\
        cosFormer~\cite{qin2021cosformer} & 75.10/75.95 & 82.61  & 86.12  & 89.45  & 81.93  & 33.03  & 74.88  & 124.70 \\ \hline
        \textit{MLP-based} & ~ & ~ & ~ & ~ & ~ & ~ & ~ & ~ \\ \hline
        Syn(D)~\cite{tay2021synthesizer} & 50.93/51.02 & 62.80  & 81.33  & 82.34  & 81.79  & -  & 58.60  & 131.00 \\
        Syn(R)~\cite{tay2021synthesizer} & 52.82/52.13 & 62.29  & 78.11  & 82.22  & 81.38  & 4.63  & 59.08  & 129.42 \\
        gMLP~\cite{liu2021pay} & 73.30/73.60 & 80.56  & 86.48  & 90.25  & 82.30  & 36.06  & 74.65  & 131.08 \\ \hline
        \textit{FFT-based} & ~ & ~ & ~ & ~ & ~ & ~ & ~ & ~ \\ \hline
        FNet~\cite{lee2021fnet} & 62.45/64.71 & 73.31  & 79.43  & 81.88  & 82.91  & -  & 63.53  & 124.70 \\
        GFNett~\cite{lee2021fnet}  & 66.75/67.45 & 65.42  & 80.25  & 84.40  & 82.44  & 9.62  & 65.19  & 130.06 \\
        AFNOt~\cite{lee2021fnet} & 68.79/69.28 & 73.20  & 85.12  & 88.88  & 82.35  & 36.19  & 71.97  & 121.57 \\ \hline
        \textit{SSM-based} & ~ & ~ & ~ & ~ & ~ & ~ & ~ & ~ \\ \hline
        S4\cite{gu2021efficiently} & 68.45/68.42 & 72.14  & 84.61  & 87.04  & 83.36  & 23.01  & 69.58  & 131.79 \\
        DSS\cite{gupta2022diagonal} & 35.46/35.22 & 50.80  & 65.18  & 65.37  & 80.95  & 6.14  & 48.45  & 123.76 \\
        GSS\cite{mehta2022long} & 50.53/51.58 & 62.58  & 80.98  & 85.67  & 82.11  & 6.56  & 60.00  & 122.80 \\ 
        % \textit{Ours} & ~ & ~ & ~ & ~ & ~ & ~ & ~ & ~ \\ \hline
        % \rowcolor{lightgray}
        TNN\cite{qin2022toeplitz} & 76.72/76.06 & 85.06  & 88.30  & 90.60  & 82.96  & 49.85  & \underline{78.51}  & 126.40 \\
        \hline
    \end{tabular}
    \vspace{-3mm}
    % \vspace{-8mm}
    \label{exp:glue_benchmark}
\end{table*}

\section{Applications of State Space Models}\label{application_ssm}

SSMs have been primarily proposed to handle long input sequences as opposed to transformers and thus, have wide applicability in a number of domains that need long sequences to be processed.

\subsection{Language Domain (long sequence) }
In the domain of natural language processing (NLP), transformers have traditionally been the preferred choice for modeling text data due to their adeptness at capturing intricate dependencies through attention mechanisms \cite{vaswani2017attention}. However, their efficiency suffers from a quadratic $O(N^2)$ complexity, especially noticeable when handling long sequences. As sequence length increases, so do memory and compute requirements, rendering training on lengthy inputs impractical. To address these inefficiencies, a slew of State Space Models (SSMs) has emerged, including S4, s4nd, S5, Hippo, Hyena, H3, LDStack, Liquid-S4, DSS, GSS, Mega, LRU, HGRN, TNN, Mamba among others \cite{gu2021efficiently}. Unlike transformers, which rely on attention mechanisms, SSMs compress input data into a fixed-size latent state. This static memory allocation remains constant during sequence generation, rendering SSMs more efficient for handling long inputs.
 
However, there's a trade-off. While SSMs excel in efficiency, they sacrifice the capability to retrieve and copy portions of the input context, crucial for tasks like few-shot learning and retrieval. Transformers, on the other hand, shine in these domains. In our performance analysis, we compare Glue scores in table~\ref{exp:glue_benchmark}, Wikitext benchmarks in table~\ref{tab:wikitext}, and Pile benchmarks in table~\ref{tab:pile}  for long sequence text data across both transformers and SSMs. The ongoing debate between the two persists, each showcasing distinct strengths and limitations in the domain of NLP \cite{kenton2019bert}.

\begin{table*}[htb]
\scriptsize
  % \captionsetup{justification=centering}
  \caption{Test accuracy on the LRA benchmark tasks~\cite{tay2020long}. \xmark\ indicates the model did not exceed random guessing. Citations refer to the original model.  The results for models ranging from Transformer to Performer are sourced from Tay et al. (2020)~\cite{tay2020long}. We compiled this table using data from the HGRN paper by Qin et al. (2023)\cite{qin2024hierarchically} and the S5 paper by Smith et al. (2022)\cite{smith2022simplified}, consolidating the results into a unified presentation below. }
  \label{tab:lra}
  \centering
    \begin{tabular}{@{}lccccccc@{}}
    \toprule
    Model               & \texttt{ListOps}                  & \texttt{Text}                 & \texttt{Retrieval}            & \texttt{Image}                & \texttt{Pathfinder}       & \texttt{Path-X}     & Avg.                              \vspace{2pt}\\ 
    (Input length)      & (2,048)                           & (4,096)                       & (4,000)                       & (1,024)                       & (1,024)                   & (16,384)                                          \\ \midrule    

    Transformer  \cite{vaswani2017attention}            & 36.37             & 64.27             & 57.46              & 42.44             & 71.40               & \xmark          & 53.66             \\
        Local Attention       & 15.82             & 52.98             & 53.39              & 41.46             & 66.63               & \xmark          & 46.71             \\
        Sparse Trans.         & 17.07             & 63.58             & 59.59              & 44.24             & 71.71               & \xmark          & 51.03             \\
        Longformer ~\cite{beltagy2020longformer}           & 35.63             & 62.85             & 56.89              & 42.22             & 69.71               & \xmark          & 52.88             \\
        Linformer ~\cite{wang2020linformer}            & 35.70             & 53.94             & 52.27              & 38.56             & 76.34               & \xmark          & 51.14             \\
        Reformer  \cite{kitaev2019reformer}              & \underline{37.27} & 56.10             & 53.40              & 38.07             & 68.50               & \xmark          & 50.56             \\
        Sinkhorn Trans. \cite{tay2020sparse}      & 33.67             & 61.20             & 53.83              & 41.23             & 67.45               & \xmark          & 51.23             \\
        Synthesizer\cite{tay2021synthesizer}           & 36.99             & 61.68             & 54.67              & 41.61             & 69.45               & \xmark          & 52.40             \\
        BigBird  \cite{zaheer2020big}               & 36.05             & 64.02             & 59.29              & 40.83             & 74.87               & \xmark          & 54.17             \\
        Linear Trans.\cite{katharopoulos2020transformers}         & 16.13             & \underline{65.90} & 53.09              & 42.34             & 75.30               & \xmark          & 50.46             \\
        Performer\cite{choromanski2020rethinking}              & 18.01             & 65.40             & 53.82              & 42.77             & 77.05               & \xmark          & 51.18             \\
        cosFormer~\cite{qin2021cosformer} & 36.50 & 67.70 & 83.15 & 51.23 & 71.96 & - & 51.76  \\ %\hline
        FLASH~\cite{hua2022transformer} & 38.70 & 64.10 & 86.10 & 47.40 & 70.25 & - & 51.09  \\ %\hline
        \midrule
    FNet   \cite{lee2021fnet}         & 35.33                             & 65.11                         & 59.61                         & 38.67                         & 77.80                     & \xmark                & 54.42                            \\
    Nystr\"omformer \cite{xiong2021nystromformer} & 37.15                             & 65.52                         & 79.56                         & 41.58                         & 70.94                     & \xmark                & 57.46                            \\
    Luna-256  \cite{ma2021luna}      & 37.25                             & 64.57                         & 79.29                         & 47.38                         & 77.72                     & \xmark                & 59.37                            \\
    H-Transformer-1D \cite{zhu2021h} & 49.53                             & 78.69                         & 63.99                         & 46.05                         & 68.78                     & \xmark               & 61.41                             \\ 
    CCNN \cite{romero2022towards}  & 43.60                             & 84.08                         & \xmark                         & 88.90                         & 91.51                     & \xmark               & 68.02                             \\  \midrule

S4\cite{gu2021efficiently}& 58.35 & 76.02&87.09&87.26&86.05&88.10   &80.48    \\
DSSEXP~\cite{gupta2022diagonal} & 59.70 & 84.60 & 87.60 & 84.90 & 84.70 & 85.60 & 81.18  \\ %\hline
    DSS\textsubscript{SOFTMAX} \cite{gupta2022diagonal} & 60.60                             & 84.80                         & 87.80                         & 85.70                         & 84.60                     & 87.80  & 81.88                                             \\
                
        % DSSEXP-NO-SCALE~\cite{dss} & 59.30 & 82.40 & 86.00 & 81.20 & 81.30 & - & 65.03  \\ %\hline

    S4D-LegS & 60.47                    & 86.18                         & 89.46                         & 88.19                & 93.06                     & 91.95            & 84.89                        \\
        Mega-chunk ($\mathcal{O}(L)$) \cite{ma2022mega} & 58.76                             & \underline{90.19}                         & 90.97                         & 85.80                         & 94.41                     & 93.81               & 85.66                             \\  
    S4-LegS  & 59.60                    & 86.82                         & 90.90                         & 88.65                & 94.20                     & 96.35            & 86.09\\
               TNN~\cite{qin2022toeplitz} & 61.04 & 87.90 & 90.97 & 88.24 & 93.00 & 96.10 & 86.21  \\ 
       
    LRU~\cite{orvieto2023resurrecting} & 60.20 & 89.40 & 89.90 & 89.00 & 95.10 & 94.20 & 86.30  \\
     HGRN\cite{qin2023hierarchically} & 59.95 & 88.14 & 94.23 & 88.69 & 92.92 & 97.50 & 86.91 \\
    SGConv~\cite{Li2022WhatMC} & 61.45 & 89.2 & 91.11 & 87.97 & 95.46 & 97.83 & 87.17 \\ 
    Liquid-S4\cite{hasani2021liquid}  & \underline{62.75}                    & 89.02                         & 91.20                         & \underline{89.50}                & 94.8                     & 96.66            & 87.32                        \\ \midrule
    {S5\cite{smith2022simplified}}                  & 62.15                             & 89.31                & \textbf{91.40}                & 88.00                         & \underline{95.33}            & \textbf{98.58}          & \underline{87.46} \\

    Mega ($\mathcal{O}(L^2)$) \cite{ma2022mega}  & \textbf{63.14}                             & \textbf{90.43}                         & 91.25                         & \textbf{90.44}                         & \textbf{96.01}                     & \underline{97.98}               & \textbf{88.21}                             \\
   
    \bottomrule
    \end{tabular}
\end{table*}

\begin{table}
\label{tab:pile}
% \small
% \vspace{-4mm}
    \caption{\textbf{ Results on the Pile: Model Sizes and Perplexity Scores.} The table below presents different language models trained on varying token counts (5 billion to 100 billion) from the Pile dataset. Lower perplexity (PPL) scores indicate better performance in language modeling tasks.  These results were adapted from the HRGN paper\cite{qin2023hierarchically} and the Hyena Hierarchy\cite{poli2023hyena}. }
    % \vspace{-2mm}
    \centering
    \begin{tabular}{l|llll}
    \hline
        Model & 5B &10B & 15B& 100B\\ \hline
        GPT (125M) &13.3& 11.9& 11.2&\\ 
        Hyena-2 (153M)& 13.3& 11.8& 11.1 &\\\hline 
        GPT (355M) &11.4& 9.8 &9.1 &\\ 
        Hyena-2 (355M) &11.3& 9.8 &9.2& \\\hline
        Transformer  (1000M) &-&-&-& 4.56  \\
        LRU (1000M) &-&-&-& 5.07  \\
        HGRN (1000M)&-&-&-& {4.14}  \\  \hline
    \end{tabular}
\end{table}

\subsection{Vision domain:}
State space models (SSMs) have demonstrated remarkable efficiency in various computer vision tasks, including image classification, segmentation, and object detection. Several efforts, such as SiMBA \cite{patro2024simba}, V-Mamba \cite{liu2024vmamba}, Vim \cite{zhu2024vision}, localMamba \cite{huang2024localmamba}, plainmamba \cite{yang2024plainmamba}, and Vim2 \cite{behrouz2024mambamixer}, have adapted SSMs for image classification tasks. Similarly, U-Mamba \cite{ma2024u}, SegMamba \cite{xing2024segmamba}, Swin U-Mamba \cite{liu2024swin}, P-Mamba (for ventricular segmentation) \cite{ye2024p}, and VM-UNet \cite{ruan2024vm} have been proposed for segmentation tasks in computer vision.

Vision-specific adaptations of the Mamba architecture, such as Vision Mamba \cite{zhu2024vision}, V-Mamba\cite{liu2024vmamba} and  SiMBA ~\cite{patro2024simba}, utilize bidirectional and visual state space models for computer vision tasks. However, there exists a performance gap between these models and state-of-the-art transformer models like SpectFormer \cite{patro2023spectformer}, SVT \cite{patro2023scattering}, WaveViT \cite{yao2022wave}, Volo \cite{yuan2022volo}, and SCT \cite{patro2024spectral}. SiMBA \cite{patro2024simba} aims to bridge this gap by incorporating Mamba for token mixing, replacing attention networks, and leveraging Einstein FFT (EinFFT) for channel mixing. SiMBA introduces the Einstein blending method for channel mixing, providing a novel approach without the constraints of requiring perfect square dimensions for sequence length and channel dimensions. Furthermore, SiMBA adopts the pyramid version of the transformer architecture, leading to significant performance improvements, especially on benchmarks like ImageNet and time series tasks. However, there remains a performance gap compared to state-of-the-art transformer models like SCT \cite{patro2024spectral}. Table~\ref{tab:imagenet_sota} summarizes results for the image classification task on the ImageNet dataset, showcasing SiMBA as the state-of-the-art model in the State Space architecture for image recognition tasks. RSMamba \cite{chen2024rsmamba} focuses on remote sensing image classification, while Res-VMamba \cite{chen2024res} targets fine-grained food category visual classification using selective state space models with deep residual learning. Additionally, efforts like SiMBA, V-Mamba, and Vim have demonstrated promising performance in object detection tasks.\

U-Mamba \cite{ma2024u} focuses on enhancing long-range dependencies for biomedical image segmentation tasks. By leveraging state space models, U-Mamba effectively captures fine-grained relationships and dependencies within biomedical images, particularly addressing long-range sequential modeling challenges. Similarly, SegMamba \cite{xing2024segmamba} extends the capabilities of Mamba for 3D medical image segmentation tasks. By incorporating the Mamba architecture, SegMamba can effectively model sequential dependencies within volumetric medical image data. MambaMorph \cite{guo2024mambamorph} tackles the problem of deformable MR-CT registration by integrating contrastive feature learning within the Mamba framework. This approach enables robust and accurate registration of medical images by leveraging the capabilities of state space models. VM-UNet \cite{ruan2024vm}, on the other hand, utilizes Vision Mamba UNet for medical image segmentation. By combining the Mamba architecture with UNet, VM-UNet achieves superior performance in segmenting medical images, leveraging the strengths of both architectures. nnMamba \cite{gong2024nnmamba} extends state space models to various biomedical image analysis tasks, including 3D biomedical image segmentation, classification, and landmark detection. By incorporating Mamba architecture, nnMamba demonstrates improved performance and efficiency in handling complex biomedical image data. FD-Vision \cite{zheng2024fd} utilizes Mamba architecture to tackle endoscopic exposure correction, enhancing robustness in medical image analysis tasks. Weak-Mamba-UNet \cite{wang2024weak} improves scribble-based medical image segmentation by integrating Mamba with CNN and ViT architectures. This integration enables more effective utilization of weak supervision signals and achieves better segmentation accuracy. MedMamba \cite{yue2024medmamba} investigates the effectiveness of Mamba architecture specifically for medical image classification tasks, demonstrating its capability to handle diverse medical imaging data effectively. LightM-UNet \cite{liao2024lightm} optimizes medical image segmentation using the lightweight Mamba architecture, emphasizing efficiency and effectiveness in handling medical imaging tasks. Large Window-based Mamba UNet \cite{wang2024large} enhances medical image segmentation by exploring innovative approaches beyond traditional convolutional and self-attention mechanisms, leading to improved segmentation performance.

H-vmunet \cite{wu2024h} introduces a high-order vision Mamba UNet architecture tailored for medical image segmentation tasks. By leveraging high-order dependencies and the Mamba architecture, H-vmunet achieves improved segmentation accuracy and robustness. ProMamba \cite{xie2024promamba} specializes in polyp segmentation by incorporating prompt-based techniques within the Mamba framework, achieving state-of-the-art results in this specific medical imaging task. CMViM \cite{yang2024cmvim} explores contrastive masked Vim autoencoders for 3D multi-modal representation learning in Alzheimer's disease classification. By leveraging Mamba architecture, CMViM achieves enhanced representation learning capabilities, contributing to improved classification performance. Gamba \cite{shen2024gamba} combines Gaussian splatting with Mamba for single-view 3D reconstruction tasks, leveraging the strengths of both techniques to achieve accurate and robust reconstruction results. ReMamber \cite{yang2024remamber} focuses on image segmentation tasks using Mamba Twister, demonstrating superior segmentation performance compared to traditional methods. MambaIR \cite{guo2024mambair} serves as a straightforward baseline for image restoration tasks utilizing state-space models. T-mamba \cite{hao2024t} enhances tooth segmentation accuracy in 3D imaging by integrating frequency-based features and gated long-range dependencies into Vision Mamba.  This integration results in improved segmentation accuracy, particularly in challenging imaging conditions such as noise, low contrast, and artifacts. Both Convolutional Neural Networks (CNNs) and transformers are popular for image segmentation. However, their handling of long-range dependencies is limited due to locality or computational complexity.  It is difficult to get efficient tooth segmentation in 3D imaging (critical for orthodontic diagnosis) face noise, low contrast, and artifacts in CBCT images.  T-Mamba integrates shared positional encoding and frequency-based features into Vision Mamba. It addresses spatial position preservation and feature enhancement in the frequency domain. T-Mamba introduces frequency-based features into vision Mamba using Gate Selection Unit which adaptively integrates two spatial domain features and one frequency domain feature.  T-mamba achieves state-of-the-art results on public Tooth CBCT datasets, outperforming previous methods significantly in various evaluation metrics such as IoU, SO, DSC, HD, and ASSD.

\begin{table}[H]
\centering
\scriptsize
% \tiny
\caption{\textbf{SOTA on ImageNet-1K}The table shows the performance of various vision backbones on the ImageNet1K\cite{deng2009imagenet} dataset for image recognition tasks.  $\star$ indicates additionally trained with the Token Labeling~\cite{wang2022scaled} for patch encoding. We have grouped the vision models into three categories based on their GFLOPs (Small, Base, and Large). The GFLOP ranges: Small (GFLOPs$<$5), Base (5$\leq$GFLOPs$<$10), and Large (10$\leq$GFLOPs$<$30). This table is adapted from the SiMBA paper \cite{patro2024simba}}
\label{tab:imagenet_sota}
\begin{tabular}{c|ccc|c}
\toprule
Method & Image Size & \#Param. & FLOPs  & Top-1 acc. \\

\toprule
\multicolumn{4}{c}{\textbf{Convnets}} \\
\midrule

ResNet-101 & $224^{2}$& 45M&-  & 77.4\\
\rowcolor{gray!15}RegNetY-8G~\cite{radosavovic2020designing} & 224$^2$ & 39M & 8.0G  & 81.7 \\
\midrule
ResNet-152 &$224^{2}$ & 60M&-  & 78.3\\
\rowcolor{gray!15}RegNetY-16G~\cite{radosavovic2020designing} & 224$^2$ & 84M & 16.0G  & 82.9 \\
\toprule
\multicolumn{4}{c}{\textbf{Transformers}} \\

\midrule
DeiT-S~\cite{touvron2021training} & 224$^2$ & 22M & 4.6G  & 79.8 \\
Swin-T~\cite{liu2022swin} & 224$^2$ & 29M & 4.5G  & 81.3 \\
EffNet-B4~\cite{li2022efficientformer} & 380$^2$ & 19M & 4.2G  & 82.9\\
WaveViT-H-S$^\star$~\cite{yao2022wave} & 224$^2$ & 22.7M & 4.1G  & 82.9\\
SpectFormer-H-S$^\star$~\cite{patro2023spectformer} & 224$^2$ & 22.2M & 3.9G  & 84.3\\
SVT-H-S$^\star$~\cite{patro2023scattering} & 224$^2$ & 22M & 3.9G  & 84.2\\
\rowcolor{gray!15}SCT-H-S$^\star$~\cite{patro2024spectral} & 224$^2$ & 21.7M & 4.1G  & 84.5\\
\midrule

EffNet-B5~\cite{li2022efficientformer} & 456$^2$ & 30M & 9.9G & 83.6 \\
Swin-S~\cite{liu2022swin} & 224$^2$ & 50M & 8.7G  & 83.0 \\
CMT-B ~\cite{guo2022cmt}& 224$^2$&45M&9.3G& 84.5\\
MaxViT-S~\cite{tu2022maxvit}& 224$^2$& 69M& 11.7G& 84.5\\
iFormer-B\cite{si2022inception}& 224$^2$ & 48M& 9.4G & 84.6 \\
Wave-ViT-B$^\star$  \cite{yao2022wave}& 224$^2$& 33M& 7.2G & 84.8  \\
{SpectFormer-H-B$^\star$}\cite{patro2023spectformer} & 224$^2$& {33.1M} & {6.3G}& {85.1} \\
{SVT-H-B$^\star$}\cite{patro2023scattering} & 224$^2$& {32.8M} & {6.3G}& {85.2} \\
\rowcolor{gray!15}{SCT-H-B$^\star$}\cite{patro2024spectral} & 224$^2$& {32.5M} & {6.5G}& {85.2} \\
% Swin-B~\cite{liu2022swin} & 384$^2$ & 88M & 47.0G & 84.7 & 84.5 \\
\midrule
% ViT-b + Monarch~\cite{fu2024monarch} &224$^2$ &33M &-& 78.9 \\%& 94.2 &  \\
M2-ViT-b~\cite{fu2024monarch}  &224$^2$ &45M  &- & {79.5}  \\
DeiT-B~\cite{touvron2021training} & 224$^2$ & 86M & 17.5G  & 81.8 \\
Swin-B~\cite{liu2022swin} & 224$^2$ & 88M & 15.4G  & 83.5 \\
M2-Swin-B~\cite{fu2024monarch}  & 224$^2$&50M & -& 83.5  \\ 
EffNet-B6~\cite{li2022efficientformer} & 528$^2$ & 43M & 19.0G  & 84.0 \\
MaxViT-B ~\cite{tu2022maxvit}& 224$^2$& 120M& 23.4G&85.0\\
% VOLO-D2$^\star$ \cite{yuan2022volo}& 224$^2$ & 58M& 14.1G & 85.2 \\
VOLO-D3$^\star$ \cite{yuan2022volo} & 224$^2$ & 86M& 20.6G & 85.4  \\
Wave-ViT-L$^\star$ \cite{yao2022wave} & 224$^2$ & 57M& 14.8G & 85.5  \\
{SpectFormer-H-L$^\star$}\cite{patro2023spectformer} & 224$^2$& {54.7M} & {12.7G} & {85.7} \\
{SVT-H-L$^\star$}\cite{patro2023scattering} & 224$^2$& {54.0M} & {12.7G} & {85.7} \\
\rowcolor{gray!15}{SCT-H-L$^\star$}\cite{patro2024spectral} & 224$^2$& {54.1M} & {13.4G} & {85.9} \\

\toprule
\multicolumn{4}{c}{\textbf{SSMs}} \\

\midrule
Vim-Ti\cite{zhu2024vision} &$224^{2}$ & 7M&- & 76.1 \\ %
PlainMamba-L1 ~\cite{yang2024plainmamba} & 224$^2$& 7M &3.0G& 77.9 \\
VMamba-T\cite{liu2024vmamba} & 224$^2$ & 22M & 5.6G  & 82.2 \\
% \rowcolor{gray!15}SiMBA-S(EinFFT) (Ours) & 224$^2$ & 26M & 5.2G  & 84.0 \\
SiMBA-S(Monarch) ~\cite{patro2024simba} & 224$^2$ & 18.5M & 3.6G  & 81.1 \\
Mamba-2D-S  ~\cite{li2024mamba}&  224 $^2$& 24M&- & 81.7\\
SiMBA-S(EinFFT)  ~\cite{patro2024simba} & 224$^2$ & 15.3M &2.4G  & 81.7 \\
LocalVMamba-T ~\cite{huang2024localmamba}  & 224$^2$& 26M & 5.7G& 82.7\\
ViM2-T  ~\cite{behrouz2024mambamixer} & 224$^2$ & 20M &- &82.7\\
\rowcolor{gray!15}SiMBA-S(MLP)  ~\cite{patro2024simba} & 224$^2$ & 26.5M & 5.0G & 84.0 \\

\midrule
Vim-S\cite{zhu2024vision} & $224^{2}$ & 26M &-& 80.5 \\
PlainMamba-L2 ~\cite{yang2024plainmamba} &  224$^2$ &25M& 8.1G &81.6\\
SiMBA-B(Monarch)  ~\cite{patro2024simba} & 224$^2$ & 26.9M & 6.3G  & 82.6 \\
Mamba-2D-B ~\cite{li2024mamba} & 224$^2$&  92M&- &  83.0\\
SiMBA-B(EinFFT)  ~\cite{patro2024simba} & 224$^2$ & 22.8M &5.2G  & 83.5 \\
VMamba-S\cite{liu2024vmamba} & 224$^2$ & 44M & 11.2G  & 83.5 \\
LocalVMamba-S ~\cite{huang2024localmamba} &  224$^2$ &50M & 11.4G& 83.7\\
ViM2-S  ~\cite{behrouz2024mambamixer} & 224$^2$ & 43M&- & 83.7\\

\rowcolor{gray!15}SiMBA-B(MLP)  ~\cite{patro2024simba} & 224$^2$ & 40.0M & 9.0G  & 84.7 \\
\midrule
HyenaViT-B~\cite{fu2024monarch}  &224$^2$ &88M &- & 78.5 \\
S4ND-ViT-B~\cite{nguyen2022s4nd} & 224$^2$ & 89M & -  & 80.4 \\
PlainMamba-L3 ~\cite{yang2024plainmamba} &  224$^2$& 50M& 14.4G& 82.3\\
VMamba-B\cite{liu2024vmamba} & 224$^2$ & 75M & 18.0G & 83.2 \\

SiMBA-L(Monarch) ~\cite{patro2024simba} & 224$^2$ & 42M & 10.7G & 83.8 \\
ViM2-B  ~\cite{behrouz2024mambamixer} & 224$^2$ & 74M&- & 83.9\\
\rowcolor{gray!15}SiMBA-L(EinFFT)  ~\cite{patro2024simba} & 224$^2$ & 36.6M & 9.6G & 84.4 \\
% SiMBA-L(MLP)$^\dagger$ ~\cite{patro2024simba} & 224$^2$ & 66.6M & 16.3G & 49.4 \\
\bottomrule
\end{tabular}
% \normalsize
 % \vspace{-0.40in}
\end{table}

\subsection{Medical Domains:} 

Mamba-based models are at the forefront of cutting-edge research in both genomics and drug design. In the biological domain, researchers leverage Mamba to analyze genomic sequences, decode genetic variations, and understand hereditary diseases. While Nucleotide Transformers and BERT-based transformers have been explored, SSMs have shown promise for genomic sequence modeling. In the chemical domain, Mamba plays a crucial role in molecular exploration and drug design. The Chemical Language Model (CLM) generates bio-active designs and diverse molecules. Recently, S4 has been adapted to enhance CLMs, overcoming the limitations of traditional transformers. From our genetic blueprint to novel compounds, Mamba is shaping the future of medicine and discovery!
\begin{itemize}

\item \textbf{Biological Domain (Genome):}
Genomics is the study of the structure, function, evolution, mapping, and editing of the genomes of an organism. The human genome has approximately about 3.1 billion base pairs. Studying the genome can result in understanding the risk for common diseases like cancer and diabetes. Several efforts have been made to process the long sequences in genomics, including the Nucleotide Transformer \cite{dalla2023nucleotide}, BERT-based transformers for DNA sequencing such as \cite{liang2023rethinking}. Transformers may, however, have complexity issues in genomics and further, limited attention window constraints on the range of dependency that can be handled. HyenaDNA \cite{nguyen2023hyenadna} has shown that SSMs can perform better than transformers for genomic sequence modeling. We show the top-1 accuracy of the genomic benchmark in table \ref{tab:genomic-benchmark} adapted from HyenaDNA paper \cite{} and another 18 dataset benchmark from Nucleotide transformer \ref{tab:nuctran}.
\begin{table} 
\setlength{\tabcolsep}{5pt}
    \caption{{\bf Nucleotide Transformer (NT) Benchmarks} 
      The Matthews correlation coefficient (MCC) is used as the performance metric for the enhancer and epigenetic marks dataset, and the F1-score is used for the promoter and splice site dataset.}
    \label{tab:nuctran}
    \vspace{5mm}
    \centering
        \begin{tabular}{lcccc}
            \toprule
            \textsc{Model} 
            & \textsc{NT} & \textsc{NT} & \textsc{NT} & {$\sf HyenaDNA$} \\
            \textsc{Params} & 500M & 2.5B & 2.5B & 1.6M \\
            \textsc{\# of Genomes} & 1 & 3,202 & 850 & 1 \\
            \midrule
            Enhancer%$^1$ %& 0.51 
            & 53.5 & 59.3 & 58.0 & \textbf{62.6} \\
            Enhancer types%$^1$ %& 0.4 
            & 48.5 & 50.0 & 47.4 & \textbf{55.7} \\
            %\midrule
            H3%$^2$  %& 0.73 
            & 73.7 & 77.6 & 81.4 & \textbf{81.7} \\%& 500 \\
            H3K4me1%$^2$  %& 0.44 
            & 35.8 & 44.5 & 55.9 & \textbf{57.1} \\%& 500 \\
            H3K4me2%$^2$  %& 0.39 
            & 28.1 & 30.0 & 32.6 & \textbf{53.9} \\%& 500 \\
            H3K4me3%$^2$  %& 0.46 
            & 26.3 & 28.1 & 42.1 & \textbf{61.2} \\%& 500 \\
            H3K9ac%$^2$  %& 0.54 
            & 46.2 & 50.8 & 57.5 & \textbf{65.1} \\%& 500 \\
            H3K14ac%$^2$  %& 0.51 
            &  37.7 & 47.1 & 55.0 & \textbf{66.3} \\%& 500 \\
            H3K36me3%$^2$  %& 0.54 
            & 46.7 & 53.3 & 63.2 & \textbf{65.3} \\%& 500 \\
            H3K79me3%$^2$  %& 0.65 
            & 57.7 & 59.2 & 64.2 & \textbf{71.6} \\%& 500 \\
            H4%$^2$  %& 0.76 
            & 76.2 & 78.9 & \textbf{82.2} & 79.6 \\%& 500 \\
            H4ac%$^2$  %& 0.5 
            & 34.4 & 42.3 & 50.1 & \textbf{63.7} \\%& 500 \\
            Promoter all%$^3$ %& 0.96 
            &  95.4 & 96.6 & \textbf{97.4} & 96.5 \\%& 300 \\
            Promoter non-TATA%$^3$ %& 0.96 
            & 95.6 & 96.9 &\textbf{97.7} & 96.6 \\%& 300 \\
            Promoter TATA%$^3$ & 0.94 
            & 94.8 & 95.8 & 96.4 & \textbf{96.7} \\%& 300 \\
            Splice acceptor%$^4$ % 0.94 
            & 96.5 & 98.5 & \textbf{99.0} & 96.6 \\%& 600 \\
            Splice donor%$^4$ & 0.95 
            & 97.2 & 98.2 & \textbf{98.4} & 97.3 \\%& 600 \\
            Splice all %$^4$ & 0.95 
            & 97.2 & 97.8 & \textbf{98.3} & 97.9 \\%& 600 \\
            \bottomrule
        \end{tabular}
\end{table}

\item \textbf{Chemical Domain (Drug Design):}
Designing molecules from scratch requires navigating the space of chemical molecules, which may be of order $(10^60)$ \cite{bohacek1996art}. A generic language model known as the Chemical Language Model (CLM)\cite{yuan2017chemical,merk2018novo,ballarotto2023novo,grisoni2021combining,grisoni2023chemical} is capable of generating experimentally validated bio-active designs and acting as molecular generators. CLMs were traditionally implemented using LSTMs, but have also been implemented using Transformers. However, the quadratic complexity of transformers limits their chemical space exploration. The authors of \cite{ozccelik2024chemical} have adapted S4 to make it work as a generative CLM in designing valid and diverse molecules. 
\end{itemize}
\begin{table} 
      \small
      \caption{{\bf GenomicBenchmarks}
        Top-1 accuracy (\%) for pretrained {$\sf HyenaDNA$}, DNABERT and Transformer (GPT ), and the previous SotA baseline CNN (scratch).
      }
      \label{tab:genomic-benchmark}
    \vspace{4mm}
    \centering
    \setlength{\tabcolsep}{4pt} % col spacing
        \begin{tabular}{lcccc}
            \toprule
            \textsc{Dataset} & \textsc{CNN} & \textsc{DNABERT} & \textsc{GPT} & \textsc{HyenaDNA} \\
            \midrule
            Mouse Enhancers & 69.0 & 66.9 & 80.1 & \textbf{85.1} \\
            Coding vs Intergenomic & 87.6 & \textbf{92.5 }& 88.8 & 91.3 \\
            Human vs Worm & 93.0 & 96.5 & 95.6 & \textbf{96.6} \\
            Human Enhancers Cohn & 69.5 & 74.0 & 70.5 & \textbf{74.2} \\
            Human Enhancers Ensembl & 68.9 & 85.7 & 83.5 & \textbf{89.2} \\
            Human Regulatory & 93.3 & 88.1 & 91.5 & \textbf{93.8} \\
            Human Nontata Promoters & 84.6 & 85.6 & 87.7 & \textbf{96.6} \\
            Human OCR Ensembl & 68.0 & 75.1 & 73.0 & \textbf{80.9} \\
            \bottomrule
        \end{tabular}
\end{table}

\subsection{Video Domain:}
Modern video processing techniques typically operate on short video clips (5-10-second clips). However, long-term video understanding problems\cite{wu2021towards} may require operating on longer video clips and modeling long-range dependencies in videos. They propose  Object Transformers, which are modeling long-range dependencies in videos. However, recent efforts such as ViS4mer\cite{islam2022long} have shown that Mamba-based architectures can be adapted to long-term video understanding tasks as well including long-term video classification. The Structured State-Space Sequence (S4) model has garnered attention as a promising solution for modeling complex spatiotemporal dependencies in long-form videos. Its linear complexity makes it an attractive choice. However, a limitation arises when treating all image tokens equally, which can impact efficiency and accuracy. Unlike previous mask-based token reduction methods, our Selective S4 (S5) model employs a lightweight mask generator. It selectively chooses informative image tokens, leading to more efficient and accurate modeling of long-term dependencies in videos. Importantly, we avoid the dense self-attention calculation by leveraging guidance from the momentum-updated S4 model. However, like other token reduction techniques, there’s a risk of incorrectly dropping informative image tokens. To enhance the robustness and temporal context, we propose a novel approach called long-short masked contrastive learning (LSMCL)\cite{wang2023selective}. This enables our model to predict longer temporal dependencies using shorter input videos.  LSMCL is evaluated extensive comparative results on challenging long-form video understanding datasets (LVU, COIN, and Breakfast) demonstrate that our approach consistently outperforms the previous state-of-the-art S4 model by up to 9.6\% accuracy, all while reducing its memory footprint by 23\%.

The VideoMamba~\cite{li2024videomamba} is a state space model (SSM)-based approach designed specifically for efficient video understanding. VideoMamba tackles the dual challenges of local redundancy and global dependencies in video data.  It achieves this by innovatively adapting the Mamba~\cite{gu2023mamba} to the video domain. The existing 3D convolutional neural networks (CNNs) and video transformers have limitations, while VideoMamba aims to overcome these limitations by leveraging the linear-complexity operator inherent in the Mamba. This operator enables efficient long-term modeling, which is crucial for understanding high-resolution, lengthy videos. VideoMamba demonstrates scalability in the visual domain without extensive datasets pertaining to using the self-distillation technique. It showcases robustness in multi-modal contexts, making it compatible with other modalities. By combining these distinct advantages, VideoMamba sets a new benchmark for video understanding on various benchmarking datasets such as Kinetics (K400)\cite{kay2017kinetics},  SthSthV2 (SSV2)\cite{goyal2017something}, Breakfast\cite{kuehne2014language},  Long-form Video Understanding (LUV)\cite{wu2021towards} and COIN\cite{tang2019coin} datasets. Table-\ref{tab:results_k400},  table-\ref{tab:results_ssv2}, table-\ref{tab:results_lvu}, and table-\ref{tab:results_breakfast_coin} shows state pf the art results in K400, SSV2, LUV and breakfast and COIN dataset respectively.  SpikeMba \cite{li2024spikemba} is a novel approach for temporal video grounding, a critical task in video content understanding. Temporal video grounding involves identifying specific moments or segments within a video that correspond to a given textual query or description. For example, if you have a video of a soccer match and a query like "goal scored by Messi," temporal video grounding aims to pinpoint the exact moment when Messi scores that goal. Existing methods struggle to capture fine-grained relationships between different modalities (such as video frames, audio, and text). SpikeMba integrates Spiking Neural Networks (SNNs) and state space models (SSMs). Contextual Moment Reasoner (CMR) balances contextual information preservation and semantic relevance exploration.  Vivim \cite{yang2024vivim} is a Video Vision Mamba for Medical Video Object Segmentation. Mamba architecture enhances sequential modeling, improving accuracy.

\begin{table}[H]
     \caption{\textbf{Comparison with the state-of-the-art on scene-related Kinetics-400.} 
    ``\textit{iso.}'' means isotropic architecture without downsampling layers.
    Masked modeling~\cite{li2023unmasked} also works for Mamba,
    but the inconsistent architecture leads to inferior alignment. This table is adapted from VideoMamba~\cite{li2024videomamba} paper.
    }
    \label{tab:results_k400}
    % \vspace{-0.6cm}
    \centering
    \setlength\tabcolsep{2pt}
    \resizebox{0.95\linewidth}{!}{
        \begin{tabular}{l|l|c|l|r|r|r|cc}
        \Xhline{1.0pt}
            \multirow{2}*{\textbf{Arch.}} & \multirow{2}*{\textbf{Model}} & \multirow{2}*{\textit{\textbf{iso.}}} & \textbf{Extra} & \textbf{Input} & \textbf{\#Param} & \textbf{FLOPs} & \multicolumn{2}{c}{\textbf{K400}} \\
            ~ & ~ & ~ & \textbf{Data} & \textbf{Size} & \textbf{(M)} & \textbf{(G)} & \textbf{Top-1} & \textbf{Top-5} \\
            \Xhline{0.8pt}
             \multicolumn{9}{l}{\textit{\textbf{\color{blue}{Supervised:}} \color{blue}{Those models with extra data are under supervised training.}}} \\
             \multirow{3}{*}{\textit{\textbf{CNN}}} & SlowFast$_{R101+NL}$~\cite{feichtenhofer2019slowfast} & \xmark &  & 80$\times$224$^2$ & 60 & 234$\times$3$\times$10 & 79.8 & 93.9 \\
             ~ & X3D-M~\cite{feichtenhofer2020x3d} & \xmark &  & 16$\times$224$^2$ & 4 & 6$\times$3$\times$10 & 76.0 & 92.3 \\
             ~ & X3D-XL~\cite{feichtenhofer2020x3d} & \xmark &  & 16$\times$312$^2$ & 20 & 194$\times$3$\times$10 & 80.4 & 94.6 \\
             \hline
             \multirow{3}{*}{\textit{\textbf{Trans.}}} & Swin-T~\cite{liu2022swin} & \xmark & IN-1K & 32$\times$224$^2$ & 28 & 88$\times$3$\times$4 & 78.8 & 93.6 \\
             ~ & Swin-B~\cite{liu2022swin} & \xmark & IN-1K & 32$\times$224$^2$ & 88 & 88$\times$3$\times$4 & 80.6 & 94.5 \\
             ~ & Swin-B~\cite{liu2022swin} & \xmark & IN-21K & 32$\times$224$^2$ & 88 & 282$\times$3$\times$4 & \underline{82.7} & \textbf{95.5} \\
             \hline
             \multirow{5}{*}{\textbf{\textit{\makecell[l]{CNN+\\Trans.}}}} & MViTv1-B~\cite{li2022mvitv2} & \xmark &  & 32$\times$224$^2$ & 37 & 70$\times$1$\times$5 & 80.2 & 94.4 \\
             ~ & MViTv2-S~\cite{li2022mvitv2} & \xmark &  & 16$\times$224$^2$ & 35 & 64$\times$1$\times$5 & 81.0 & 94.6 \\
             ~ & UniFormer-S~\cite{li2022uniformer} & \xmark & IN-1K & 16$\times$224$^2$ & 21 & 42$\times$1$\times$4 & 80.8 & 94.7 \\
             ~ & UniFormer-B~\cite{li2022uniformer} & \xmark & IN-1K & 16$\times$224$^2$ & 50 & 97$\times$1$\times$4 & 82.0 & 95.1 \\
             ~ & UniFormer-B~\cite{li2022uniformer} & \xmark & IN-1K & 32$\times$224$^2$ & 50 & 259$\times$3$\times$4 & \textbf{83.0} & \underline{95.4} \\
             \Xhline{0.8pt}
             \multirow{4}{*}{\textit{\textbf{Trans.}}} & STAM~\cite{sharir2021image} & \cmark & IN-21K & 64$\times$224$^2$ & 121 & 1040$\times$1$\times$1 & 79.2 & - \\
             ~ & TimeSformer-L~\cite{bertasius2021space} & \cmark & IN-21K & 96$\times$224$^2$ & 121 & 2380$\times$3$\times$1 & 80.7 & 94.7 \\
             ~ & ViViT-L~\cite{arnab2021vivit} & \cmark & IN-21K & 16$\times$224$^2$ & 311 & 3992$\times$3$\times$4 & \underline{81.3} & 94.7 \\
             ~ & Mformer-HR~\cite{patrick2021keeping} & \cmark & IN-21K & 16$\times$336$^2$ & 311 & 959$\times$3$\times$10 & 81.1 & \underline{95.2} \\
             \hline
             \multirow{9}{*}{\textit{\textbf{SSM}}} & {VideoMamba-Ti} & {\cmark} & {IN-1K} & {16$\times$224$^2$} & {7} & {17$\times$3$\times$4} & {78.1} & {93.5} \\
             ~ & {VideoMamba-Ti}~\cite{li2024videomamba} & {\cmark} & {IN-1K} & {32$\times$224$^2$} & {7} & {34$\times$3$\times$4} & {78.8} & {93.9} \\
             ~ & {VideoMamba-Ti}~\cite{li2024videomamba} & {\cmark} & {IN-1K} & {64$\times$384$^2$} & {7} & {202$\times$3$\times$4} & {80.3} & {94.8} \\

             ~ & {VideoMamba-S} ~\cite{li2024videomamba}& {\cmark} & {IN-1K} & {16$\times$224$^2$} & {26} & {68$\times$3$\times$4} & {80.8} & {94.8} \\
             ~ & {VideoMamba-S}~\cite{li2024videomamba} & {\cmark} & {IN-1K} & {32$\times$224$^2$} & {26} & {135$\times$3$\times$4} & {81.5} & {95.2} \\
             ~ & {VideoMamba-S}~\cite{li2024videomamba} & {\cmark} & {IN-1K} & {64$\times$384$^2$} & {26} & {395$\times$3$\times$4} & {82.7} & {95.6} \\

             ~ & {VideoMamba-M}~\cite{li2024videomamba} & {\cmark} & {IN-1K} & {16$\times$224$^2$} & {74} & {202$\times$3$\times$4} & {81.9} & {95.4} \\
             ~ & {VideoMamba-M}~\cite{li2024videomamba} & {\cmark} & {IN-1K} & {32$\times$224$^2$} & {74} & {403$\times$3$\times$4} & {82.4} & {95.7} \\
             ~ & {VideoMamba-M} ~\cite{li2024videomamba}& {\cmark} & {IN-1K} & {64$\times$384$^2$} & {74} & {2368$\times$3$\times$4} & {\textbf{83.3}} & {\textbf{96.1}} \\
             
             \Xhline{1.0pt}
             \multicolumn{9}{l}{\textit{\textbf{\color{blue}{Self-supervised:}} \color{blue}{For UMT, the CLIP-400M is used in pretrained teacher.}}} \\
             \multirow{5}{*}{\textit{\textbf{Trans.}}} & BEVT-B$_{800e}$~\cite{wang2022bevt} & \xmark & IN-1K & 32$\times$224$^2$ & 88 & 282$\times$3$\times$4 & 81.1 & - \\
             \cline{2-9}
             ~ & ST-MAE-B$_{1600e}$~\cite{feichtenhofer2022masked} & \cmark &  & 16$\times$224$^2$ & 87 & 180$\times$3$\times$7 & 81.3 & 94.9 \\
             ~ & VideoMAE-S$_{2400e}$~\cite{tong2022videomae} & \cmark &  & 16$\times$224$^2$ & 22 & 57$\times$3$\times$5 & 79.0 & 93.8 \\
             ~ & VideoMAE-B$_{1600e}$~\cite{tong2022videomae} & \cmark &  & 16$\times$224$^2$ & 87 & 180$\times$3$\times$5 & 81.5 & 95.1 \\
             ~ & UMT-B$_{800e}$~\cite{li2023unmasked} & \cmark & \color{blue}{CLIP-400M} & 8$\times$224$^2$ & 87 & 180$\times$3$\times$5 & \textbf{85.7} & \textbf{97.0} \\

             \hline
             \multirow{4}{*}{\textit{\textbf{SSM}}} & {VideoMamba-M$_{800e}$}~\cite{li2024videomamba} & {\cmark} & {\color{blue}{CLIP-400M}} & {8$\times$224$^2$} & {74} & {101$\times$3$\times$4} & {82.0} & {95.4} \\
             ~ & {VideoMamba-M$_{800e}$}~\cite{li2024videomamba} & {\cmark} & {\color{blue}{CLIP-400M}} & {16$\times$224$^2$} & {74} & {202$\times$3$\times$4} & {83.4} & {95.9} \\
             ~ & {VideoMamba-M$_{800e}$}~\cite{li2024videomamba} & {\cmark} & {\color{blue}{CLIP-400M}} & {32$\times$224$^2$} & {74} & {403$\times$3$\times$4} &{83.9} & {96.2} \\
             ~ & {VideoMamba-M$_{800e}$} ~\cite{li2024videomamba} & {\cmark} & {\color{blue}{CLIP-400M}} & {64$\times$384$^2$} & {74} & {2368$\times$3$\times$4} &{\underline{85.0}} & {\underline{96.9}} \\
        \Xhline{1.0pt}	
        \end{tabular}
    }

\end{table}

\begin{table}[H]
    % \vspace{-0.3cm}
     \caption{\textbf{Comparison with the state-of-the-art on temporal-related SthSth V2.} 
    ``\textit{iso.}'' means isotropic architecture without downsampling layers.
    Masked modeling~\cite{li2023unmasked} also works for Mamba, 
    and it performs better than VideoMAE\cite{tong2022videomae}. This table is adapted from VideoMamba\cite{li2024videomamba} paper
    }
    \label{tab:results_ssv2}
    % \vspace{-0.6cm}
    \centering
    \setlength\tabcolsep{2pt}
    \resizebox{0.96\linewidth}{!}{
        \begin{tabular}{l|l|c|l|r|r|r|cc}
        \Xhline{1.0pt}
            \multirow{2}*{\textbf{Arch.}} & \multirow{2}*{\textbf{Model}} & \multirow{2}*{\textit{\textbf{iso.}}} & \textbf{Extra} & \textbf{Input} & \textbf{\#Param} & \textbf{FLOPs} & \multicolumn{2}{c}{\textbf{SSV2}} \\
            ~ & ~ & ~ & \textbf{Data} & \textbf{Size} & \textbf{(M)} & \textbf{(G)} & \textbf{Top-1} & \textbf{Top-5} \\
            \Xhline{0.8pt}
             \multicolumn{9}{l}{\textit{\textbf{\color{blue}{Supervised:}} \color{blue}{Those models with extra data are under supervised training.}}} \\
             \multirow{3}{*}{\textit{\textbf{CNN}}} & SlowFast$_{R101}$~\cite{feichtenhofer2019slowfast} & \xmark & K400 & 32$\times$224$^2$ & 53 & 106$\times$3$\times$1 & 63.1 & 87.6 \\
             ~ & CT-Net$_{R50}$~\cite{li2020ct} & \xmark & IN-1K & 16$\times$224$^2$ & 21 & 75$\times$1$\times$1 & 64.5 & 89.3 \\
             ~ & TDN$_{R50}$~\cite{wang2021tdn} & \xmark & IN-1K & 16$\times$224$^2$ & 26 & 75$\times$1$\times$1 & 65.3 & 91.6 \\
             \hline
             \multirow{1}{*}{\textit{\textbf{Trans.}}} & Swin-B~\cite{liu2022swin} & \xmark & K400 & 32$\times$224$^2$ & 89 & 88$\times$3$\times$1 & 69.6 & 92.7 \\
             \hline
             \multirow{6}{*}{\textbf{\textit{\makecell[l]{CNN+\\Trans.}}}} & MViTv1-B~\cite{li2022mvitv2} & \xmark & K400 & 16$\times$224$^2$ & 37 & 71$\times$3$\times$1 & 64.7 & 89.2 \\
             ~ & MViTv1-B~\cite{li2022mvitv2} & \xmark & K400 & 32$\times$224$^2$ & 37 & 170$\times$3$\times$1 & 67.1 & 90.8 \\
             ~ & MViTv2-S~\cite{li2022mvitv2} & \xmark & K400 & 16$\times$224$^2$ & 35 & 65$\times$3$\times$1 & 68.2 & 91.4 \\
             ~ & MViTv2-B~\cite{li2022mvitv2} & \xmark & K400 & 32$\times$224$^2$ & 51 & 225$\times$3$\times$1 & \textbf{70.5} & \underline{92.7} \\
             ~ & UniFormer-S~\cite{li2022uniformer} & \xmark & \scriptsize{IN-1K+K400} & 16$\times$224$^2$ & 21 & 42$\times$3$\times$1 & 67.7 & 91.4 \\
             ~ & UniFormer-B~\cite{li2022uniformer} & \xmark & \scriptsize{IN-1K+K400} & 16$\times$224$^2$ & 50 & 97$\times$3$\times$1 & \underline{70.4} & \textbf{92.8} \\
             \Xhline{0.8pt}
             \multirow{3}{*}{\textit{\textbf{Trans.}}} & TimeSformer-HR~\cite{bertasius2021space} & \cmark & IN-21K & 16$\times$224$^2$ & 121 & 1703$\times$3$\times$1 & 62.5 & - \\
             ~ & ViViT-L~\cite{arnab2021vivit} & \cmark & \scriptsize{IN-21K+K400} & 16$\times$224$^2$ & 311 & 3992$\times$3$\times$4 & 65.4 & 89.8 \\
             ~ & Mformer-HR~\cite{patrick2021keeping} & \cmark & \scriptsize{IN-21K+K400} & 16$\times$336$^2$ & 311 & 1185$\times$3$\times$1 & \underline{68.1} & \underline{91.2} \\
             \hline
             \multirow{9}{*}{\textit{\textbf{SSM}}} & {VideoMamba-Ti} ~\cite{li2024videomamba} & {\cmark} & {IN-1K} & {8$\times$224$^2$} & {7} & {9$\times$3$\times$2} & {65.1} & {89.1} \\
             ~ & {VideoMamba-Ti}  ~\cite{li2024videomamba}& {\cmark} & {IN-1K} & {16$\times$224$^2$} & {7} & {17$\times$3$\times$2} & {66.0} & {89.6} \\
             ~ & {VideoMamba-Ti} ~\cite{li2024videomamba} & {\cmark} & {IN-1K} & {16$\times$288$^2$} & {7} & {28$\times$3$\times$2} & {66.2} & {90.0} \\

             ~ & {VideoMamba-S}  ~\cite{li2024videomamba}& {\cmark} & {IN-1K} & {8$\times$224$^2$} & {26} & {34$\times$3$\times$2} & {66.6} & {90.4} \\
             ~ & {VideoMamba-S}  ~\cite{li2024videomamba}&{\cmark} & {IN-1K} & {16$\times$224$^2$} & {26} & {68$\times$3$\times$2} & {67.6} & {90.9} \\
             ~ & {VideoMamba-S} ~\cite{li2024videomamba} &{\cmark} & {IN-1K} & {16$\times$288$^2$} & {26} & {112$\times$3$\times$2} & {68.1} & {91.2} \\

             ~ & {VideoMamba-M} ~\cite{li2024videomamba} & {\cmark} & {IN-1K} & {8$\times$224$^2$} & {74} & {101$\times$3$\times$4} & {67.3} & {91.0} \\
             ~ & {VideoMamba-M}  ~\cite{li2024videomamba}& {\cmark} & {IN-1K} & {16$\times$224$^2$} & {74} & {202$\times$3$\times$4} & {68.3} & {91.4} \\
             ~ & {VideoMamba-M} ~\cite{li2024videomamba} & {\cmark} & {IN-1K} & {16$\times$288$^2$} & {74} & {333$\times$3$\times$4} & {\textbf{68.4}} & {\textbf{91.6}} \\
             
             \Xhline{1.0pt}
             \multicolumn{9}{l}{\textit{\textbf{\color{blue}{Self-supervised:}} {For UMT, the CLIP-400M is used in pretrained teacher.}}} \\
             \multirow{4}{*}{\textit{\textbf{Trans.}}} & BEVT-B$_{800e}$~\cite{wang2022bevt} & \xmark & \scriptsize{IN-1K+K400} & 32$\times$224$^2$ & 88 & 321$\times$3$\times$1 & 70.6 & - \\
             \cline{2-9}
             ~ & VideoMAE-S$_{2400e}$~\cite{tong2022videomae} & \cmark &  & 16$\times$224$^2$ & 22 & 57$\times$3$\times$2 & 66.8 & 90.3 \\
             ~ & VideoMAE-B$_{2400e}$~\cite{tong2022videomae} & \cmark &  & 16$\times$224$^2$ & 87 & 180$\times$3$\times$2 & \underline{70.8} & 92.4 \\
             ~ & UMT-B$_{800e}$~\cite{li2023unmasked} & \cmark & \color{blue}{CLIP-400M} & 8$\times$224$^2$ & 87 & 180$\times$3$\times$2 & \underline{70.8} & \underline{92.6} \\

             \hline
             \multirow{3}{*}{\textit{\textbf{SSM}}} & {VideoMamba-M$_{800e}$} & {\cmark} & {\color{blue}{CLIP-400M}} & {8$\times$224$^2$} & {74} & {101$\times$3$\times$2} & {70.2} & {92.6} \\
             ~ & {VideoMamba-M$_{800e}$}  ~\cite{li2024videomamba}& {\cmark} & {\color{blue}{CLIP-400M}} & {16$\times$224$^2$} & {74} & {202$\times$3$\times$2} & {71.0} & {92.7} \\
             ~ & {VideoMamba-M$_{800e}$} ~\cite{li2024videomamba} & {\cmark} & {\color{blue}{CLIP-400M}} & {16$\times$288$^2$} & {74} & {333$\times$3$\times$2} & {\textbf{71.4}} & {\textbf{92.9}} \\
        \Xhline{1.0pt}	
        \end{tabular}
    }
   
\end{table}

\begin{table}[H]
    \centering
     \caption{\textbf{Comparison with the state-of-the-art on LVU.} 
    ``\textit{e2e}'' means end-to-end methods without exhausting feature extraction.
    ``Rel.'', ``Dir.'' and ``Wtr.'' refers to ``Relation'', ``Director'' and ``Writer'', respectively. This table is adapted from VideoMamba\cite{li2024videomamba} paper
    }
    \label{tab:results_lvu}
    \setlength\tabcolsep{2pt}
    \resizebox{0.98\linewidth}{!}{
        \begin{tabular}{l|c|l|ccc|cccc|cc}
        \Xhline{1.0pt}
            \multirow{2}*{\textbf{Method}} & \multirow{2}*{\textit{\textbf{e2e}}} & \multirow{2}*{\textit{\textbf{Backbone}}} & \multicolumn{3}{c|}{\textbf{Content($\uparrow$)}} & \multicolumn{4}{c|}{\textbf{Metadata($\uparrow$)}} & \multicolumn{2}{c}{\textbf{User($\downarrow$)}} \\
            ~ & ~ & ~ & \textbf{Rel.} & \textbf{Speak} & \textbf{Scene} & \textbf{Dir.} & \textbf{Genre} & \textbf{Wtr.} & \textbf{Year} & \textbf{Like} & \textbf{View} \\
            \Xhline{0.8pt}
            VideoBERT~\cite{sun2019videobert} & \xmark & S3D & 52.80 & 37.90 & 54.90 & 47.30 & 51.90 & 38.50 & 36.10 & 0.32 & 4.46 \\
            Object Trans.\cite{wu2021towards} & \xmark & ResNet &  53.10 & 39.40 & 56.90 & 51.20 & 54.60 & 34.50 & 39.10 & \textbf{0.23} & \underline{3.55} \\
            LST~\cite{islam2022long} & \xmark & ViT-L &  52.38 & 37.31 &	62.79 &	56.07 &	52.70 &	42.26 &	39.16 &	0.31 &	3.83 \\
            Performer~\cite{islam2022long} & \xmark & ViT-L & 50.00 &	38.80 &	60.46 &	58.87 &	49.45 &	48.21 &	41.25 &	0.31 &	3.93 \\
            Orthoformer~\cite{islam2022long} & \xmark & ViT-L &  50.00 &	39.30 &	66.27 &	55.14 &	\underline{55.79} &	47.02&	43.35 &	0.29 &	3.86 \\  \Xhline{0.8pt}
            ViS4mer~\cite{islam2022long} & \xmark & ViT-L &  \underline{57.14} & \textbf{40.79} & \underline{67.44} & \underline{62.61} & 54.71 & \underline{48.80} & \underline{44.75} & \underline{0.26} & 3.63 \\       
            
            VideoMamba$_{f32}$ ~\cite{li2024videomamba} & \cmark & VM-Ti & \textbf{62.50} & \underline{40.43} & \textbf{70.37} & \textbf{67.29} & \textbf{65.24} & \textbf{52.98} & \textbf{48.23} & \underline{0.26} & \textbf{2.90} \\
        \Xhline{1.0pt}	
        \end{tabular}
    }
   
    % \vspace{-0.5cm}
\end{table}

\subsection{Tabular Domain}
 While CNNs and transformers have been shown to perform well on tabular data, they may require significant computing resources, data pre-processing as well as tuning. They may not handle incremental feature learning, where features may get added sequentially. MambaTab \cite{ahamed2024mambatab} has shown that it can model incremental feature learning efficiently and has shown to perform comparably with CNNs and transformers while being more parameter efficient. Unlike traditional models, MambaTab can adapt to datasets with increasing features over time without requiring complete retraining along with that it requires minimal preprocessing and utilizes fewer computational resources. 

 This work develops an approach based on a structured state-space model (SSM), MambaTab, for tabular data. SSMs have strong capabilities for efficiently extracting effective representations from data with long-range dependencies. MambaTab leverages Mamba, an emerging SSM variant, for end-to-end supervised learning on tables. Compared to state-of-the-art baselines, MambaTab delivers superior performance while requiring significantly fewer parameters and minimal preprocessing, as empirically validated on diverse benchmark datasets. MambaTab's efficiency, scalability, generalizability, and predictive gains signify it as a lightweight, "out-of-the-box" solution for diverse tabular data with promise for enabling wider practical applications.

\subsection{Audio and Speech Domain:}

In the domain of speech processing, Recurrent Neural Networks (RNNs) have conventionally been the primary choice for various tasks. However, the advent of attention-based transformers spurred endeavors to apply them to speech datasets. While attention networks excel at modeling long-range dependencies, they often struggle with handling local dependencies, necessitating a combination of convolutions. This fusion is exemplified in models like Conformer \cite{49414} and BranchFormer \cite{BranchFormer}. A pioneering effort in applying State Space Models (SSMs) to speech tasks is SaShiMi \cite{goel2022s}. Recognizing the numerical instability of S4 for auto-regressive tasks, SaShiMi enforces the state matrix to be Hurwitz, rendering it suitable for tasks like audio generation. Multi-head SSM (MH-SSM) introduces an MH-SSM layer as a replacement for the attention layer in transformers, branding the hybrid model as StateFormer. StateFormer achieves state-of-the-art performance on speech datasets such as LibriSpeech \cite{LibriSpeech}.

SP-Mamba \cite{li2024spmamba} underscores the inadequacy of attention-based transformers for modeling lengthy speech signals due to their quadratic complexity. Instead, SP-Mamba adopts a bi-directional Mamba architecture, leveraging both time-domain and frequency-domain features. It replaces Bidirectional Long Short-Term Memory (BLSTM) with multiheaded self-attention for the time-frequency domain module. Evaluation of datasets like LibriSpeech \cite{LibriSpeech} and WHAM \cite{wichern2019wham} validates SP-Mamba's effectiveness. Concurrently, Dual-Path Mamba (DPMamba) \cite{jiang2024dual} splits long speech signals into manageable chunks and applies Mamba across all chunks in both temporal directions. DPMamba demonstrates superior performance over transformers in tasks like speech separation, as evidenced by evaluations on datasets like WSJ0-2mix \cite{hershey2016deep}. Additionally, Multichannel Long-Term Streaming Neural Speech Enhancement for Static and Moving Speakers \cite{quan2024multichannel} proposes a Mamba-based variant of SpatialNet to replace self-attention, showcasing the versatility of SSMs in enhancing speech-related tasks.

\subsection{Time Series Domain:}
In the domain of time series modeling, traditional approaches have relied on statistical models like ARIMA(AutoRegressive Integrated Moving Average) for forecasting and analysis. However, recent advancements have witnessed a shift towards employing transformers, originally designed for natural language processing, in the time series domain. Models such as AutoFormer\cite{wu2021autoformer}, Informer\cite{zhou2021informer}, FEDFormer\cite{zhou2022fedformer}, TFT\cite{lim2021temporal}, WaveNet\cite{van2016wavenet}, and PatchTST\cite{nie2022time} have been developed to adapt transformers to time series data. Despite their effectiveness, many of these models grapple with two key challenges: attention complexity and the ability to capture long-range dependencies inherent in time series data. To address these issues, recent research has turned to integrating state space models like S4 \cite{gu2021efficiently} into time series analysis.

State-of-the-art approaches for time series modeling now include innovative models such as Timemachine \cite{ahamed2024timemachine}, SiMBA \cite{patro2024simba}, and MambaMix \cite{behrouz2024mambamixer}. These models leverage the power of state space models to efficiently capture temporal dependencies and patterns in time series data. To provide a comprehensive comparison between transformers and state space models for time series analysis, we collate benchmark results from both domains, as outlined in Table \ref{exp:mts} (adapted from the SiMBA paper \cite{patro2024simba}). Through such comparative analysis, researchers gain valuable insights into the strengths and weaknesses of each approach, aiding further advancements in time series modeling.

\linespread{1.2}
\begin{table*}[thb]
\scriptsize
	\centering
	\resizebox{\linewidth}{!}{
		\begin{tabular}{cc|c|cc|cc|cc|cc|cc|cc|cc|cc|cc|ccc}
			\cline{2-23}
			&\multicolumn{2}{c|}{Models}& \multicolumn{2}{c|}{Simba}& \multicolumn{2}{c|}{TimesNet}& \multicolumn{2}{c|}{Crossformer}& \multicolumn{2}{c|}{PatchTST}& \multicolumn{2}{c|}{ETSFormer}& \multicolumn{2}{c|}{DLinear}& \multicolumn{2}{c|}{FEDFormer}& \multicolumn{2}{c}{Autoformer}&\multicolumn{2}{c}{Pyraformer}&\multicolumn{2}{c}{MTGNN}& \\
			\cline{2-23}
			&\multicolumn{2}{c|}{Metric}&MSE&MAE&MSE&MAE&MSE&MAE&MSE&MAE&MSE&MAE&MSE&MAE&MSE&MAE&MSE&MAE&MSE&MAE&MSE&MAE\\
			\cline{2-23}
			&\multirow{4}*{\rotatebox{90}{ETTm1}}& 96 &\textbf{0.324}&\textbf{0.360} & \uline{0.338} & \uline{0.375} & 0.349 & 0.395 & 0.339 & 0.377 & 0.375 & 0.398 & 0.345 & 0.372 & 0.379 & 0.419 & 0.505 & 0.475 & 0.543 & 0.510 & 0.379 & 0.446 \\
            &\multicolumn{1}{c|}{}& 192 & \textbf{0.363}&\textbf{0.382} & \uline{0.374} & \uline{0.387} & 0.405 & 0.411 & 0.376 & 0.392 & 0.408 & 0.410 & 0.380 & 0.389 & 0.426 & 0.441 & 0.553 & 0.496 & 0.557 & 0.537 & 0.470 & 0.428 \\
            &\multicolumn{1}{c|}{}& 336 & \textbf{0.395} &\textbf{0.405}  & 0.410 & \uline{0.411} & 0.432 & 0.431 & \uline{0.408} & 0.417 & 0.435 & 0.428 & 0.413 & 0.413 & 0.445 & 0.459 & 0.621 & 0.537 & 0.754 & 0.655 & 0.473 & 0.430 \\
            &\multicolumn{1}{c|}{}& 720 & \textbf{0.451} &\textbf{0.437}  & \uline{0.478} & \uline{0.450} & 0.487 & 0.463 & 0.499 & 0.461 & 0.499 & 0.462 & 0.474 & 0.453 & 0.543 & 0.490 & 0.671 & 0.561 & 0.908 & 0.724 & 0.553 & 0.479 \\
			\cline{2-23}
			&\multirow{4}*{\rotatebox{90}{ETTm2}}& 96 &\textbf{0.177} & \textbf{0.263}& \uline{0.187}& \uline{0.267}& 0.208& 0.292& 0.192& 0.273& 0.189 & 0.280 & 0.193 & 0.292 & 0.203 & 0.287 & 0.255 & 0.339 & 0.435 & 0.507 & 0.203 & 0.299 \\
            &\multicolumn{1}{c|}{} & 192 &\textbf{0.245} &\textbf{0.306} & \uline{0.249}& \uline{0.309}& 0.263& 0.332& 0.252& 0.314& 0.253 & 0.319 & 0.284 & 0.362 & 0.269 & 0.328 & 0.281 & 0.340 & 0.730 & 0.673 & 0.265 & 0.328 \\ 
            &\multicolumn{1}{c|}{}& 336 &\textbf{0.304} &\textbf{0.343} & 0.321& \uline{0.351}& 0.337& 0.369& \uline{0.318}& 0.357& 0.314 & 0.357 & 0.369 & 0.427 & 0.325 & 0.366 & 0.339 & 0.372 & 1.201 & 0.845 & 0.365 & 0.374 \\
            &\multicolumn{1}{c|}{}&720 & \textbf{0.400}&\textbf{0.399} &\uline{ 0.408}& \uline{0.403}& 0.429& 0.430& 0.413& 0.416& 0.414 & 0.413 & 0.554 & 0.522 & 0.421 & 0.415 & 0.433 & 0.432 & 3.625 & 1.451 & 0.461 & 0.459 \\
            \cline{2-23}
			&\multirow{4}*{\rotatebox{90}{ETTh1}}& 96 & \textbf{0.379} & \textbf{0.395} & \uline{0.384} & \uline{0.402} & 0.384 & 0.428 & 0.385 & 0.408 & 0.494 & 0.479 & 0.386 & 0.400 & 0.376 & 0.419 & 0.449 & 0.459 & 0.664 & 0.612 & 0.515 & 0.517 \\
			&\multicolumn{1}{c|}{}&192 & \uline{0.432} &\textbf{0.424 } & 0.436 & \uline{0.429} & 0.438 & 0.452 &\textbf{ 0.431} & 0.432 & 0.538 & 0.504 & 0.437 & 0.432 & 0.420 & 0.448 & 0.500 & 0.482 & 0.790 & 0.681 & 0.553 & 0.522 \\
			&\multicolumn{1}{c|}{}& 336 & \textbf{0.473} & \textbf{0.443} & 0.491 & 0.469 & 0.495 & 0.483 & \uline{0.485} & \uline{0.462} & 0.574 & 0.521 & 0.481 & 0.459 & 0.459 & 0.465 & 0.521 & 0.496 & 0.891 & 0.738 & 0.612 & 0.577 \\
			&\multicolumn{1}{c|}{}& 720 & \textbf{0.483} &\textbf{0.469}  & 0.521 & 0.500 & 0.522 & 0.501 & \uline{0.497} & \uline{0.483} & 0.562 & 0.535 & 0.519 & 0.516 & 0.506 & 0.507 & 0.514 & 0.512 & 0.963 & 0.782 & 0.609 & 0.597 \\
			\cline{2-23}
			&\multirow{4}*{\rotatebox{90}{ETTh2}}& 96 & \textbf{0.290} & \textbf{0.339} & \uline{0.340} & \uline{0.374} & 0.347 & 0.391 & 0.343 & 0.376 & 0.340 & 0.391 & 0.333 & 0.387 & 0.358 & 0.397 & 0.346 & 0.388 & 0.645 & 0.597 & 0.354 & 0.454 \\ 
            &\multicolumn{1}{c|}{} & 192 & \textbf{0.373} &\textbf{ 0.390} & \uline{0.402} & \uline{0.414} & 0.419 & 0.427 & 0.405 & 0.417 & 0.430 & 0.439 & 0.477 & 0.476 & 0.429 & 0.439 & 0.456 & 0.452 & 0.788 & 0.683 & 0.457 & 0.464 \\
            &\multicolumn{1}{c|}{}& 336 & \textbf{0.376} & \textbf{0.406} & 0.452 & \uline{0.452} & 0.449 & 0.465 &\uline{ 0.448} & 0.453 & 0.485 & 0.479 & 0.594 & 0.541 & 0.496 & 0.487 & 0.482 & 0.486 & 0.907 & 0.747 & 0.515 & 0.540 \\
            &\multicolumn{1}{c|}{}& 720 &\textbf{0.407}  & \textbf{0.431} & \uline{0.462} & 0.468 & 0.479 & 0.505 & 0.464 & \uline{0.483} & 0.500 & 0.497 & 0.831 & 0.657 & 0.463 & 0.474 & 0.515 & 0.511 & 0.963 & 0.783 & 0.532 & 0.576 \\
			\cline{2-23}
			&\multirow{4}*{\rotatebox{90}{Electricity}}& 96 & \uline{0.165} &\textbf{ 0.253} & 0.168 & 0.272 & 0.185 & 0.288 & \textbf{0.159} & \uline{0.268} & 0.187 & 0.304 & 0.197 & 0.282 & 0.193 & 0.308 & 0.201 & 0.317 & 0.386 & 0.449 & 0.217 & 0.318\\
            &\multicolumn{1}{c|}{}& 192 &\textbf{0.173 } & \textbf{0.262} & 0.198 & 0.300 & 0.211 & 0.312 & \uline{0.195} & \uline{0.296} & 0.212 & 0.329 & 0.209 & 0.301 & 0.214 & 0.329 & 0.231 & 0.338 & 0.376 & 0.443 & 0.260 & 0.348 \\
            &\multicolumn{1}{c|}{}& 336 & \textbf{0.188} & \textbf{0.277} & 0.198 & 0.300 & 0.211 & 0.312 & \uline{0.195} & \uline{0.296} & 0.212 & 0.329 & 0.209 & 0.301 & 0.214 & 0.329 & 0.231 & 0.338 & 0.376 & 0.443 & 0.260 & 0.348 \\
            &\multicolumn{1}{c|}{}& 720 & \textbf{0.214} & \textbf{0.305} & 0.220 & 0.320 & 0.223 & 0.335 & \uline{0.215} & \uline{0.317} & 0.233 & 0.345 & 0.245 & 0.333 & 0.246 & 0.355 & 0.254 & 0.361 & 0.376 & 0.445 & 0.290 & 0.369 \\
			\cline{2-23}
			&\multirow{4}*{\rotatebox{90}{Traffic}}& 96 & \textbf{0.468} & \textbf{0.268} & 0.593 & 0.321 & 0.591 & 0.329 &  \uline{0.583} &  \uline{0.319} & 0.607 & 0.392 & 0.650 & 0.396 & 0.587 & 0.366 & 0.613 & 0.388 & 0.867 & 0.468 & 0.660 & 0.437 \\
            &\multicolumn{1}{c|}{}&192 &\textbf{0.413}  &\textbf{ 0.317} & 0.617 & 0.336 & 0.607 & 0.345 & \uline{ 0.591} &  \uline{0.331} & 0.621 & 0.399 & 0.598 & 0.370 & 0.604 & 0.373 & 0.616 & 0.382 & 0.869 & 0.467 & 0.649 & 0.438 \\
            &\multicolumn{1}{c|}{}& 336 & \textbf{0.529} & \textbf{0.284} & 0.629 & 0.336 & 0.613 & 0.339 &  \uline{0.599} &  \uline{0.332} & 0.622 & 0.396 & 0.605 & 0.373 & 0.621 & 0.383 & 0.622 & 0.337 & 0.881 & 0.469 & 0.653 & 0.472 \\
            &\multicolumn{1}{c|}{}& 720 & \textbf{0.564} & \textbf{0.297} & 0.640 & 0.350 & 0.620 & 0.348 &  \uline{0.601} &  \uline{0.341} & 0.632 & 0.396 & 0.645 & 0.394 & 0.626 & 0.382 & 0.660 & 0.408 & 0.896 & 0.473 & 0.639 & 0.437 \\
			\cline{2-23}
			&\multirow{4}*{\rotatebox{90}{Weather}}&96 & 0.176 & \textbf{0.219} & \uline{0.172} & \uline{0.220} & 0.191 & 0.251 & \textbf{0.171} & 0.230 & 0.197 & 0.281 & 0.196 & 0.255 & 0.217 & 0.296 & 0.266 & 0.336 & 0.622 & 0.556 & 0.230 & 0.329 \\ 
            &\multicolumn{1}{c|}{}& 192 & 0.222 & \textbf{0.260} & \textbf{0.219} & \uline{0.261} & \uline{0.219} & 0.279 & 0.219 & 0.271 & 0.237 & 0.312 & 0.237 & 0.296 & 0.276 & 0.336 & 0.307 & 0.367 & 0.739 & 0.624 & 0.263 & 0.322 \\
            &\multicolumn{1}{c|}{}& 336 & \textbf{0.275} & \textbf{0.297} &  \uline{0.280} &  \uline{0.306} & 0.287 & 0.332 & 0.277 & 0.321 & 0.298 & 0.353 & 0.283 & 0.335 & 0.339 & 0.380 & 0.359 & 0.395 & 1.004 & 0.753 & 0.354 & 0.396 \\ 
            &\multicolumn{1}{c|}{}&720 &\textbf{ 0.350} & \textbf{0.349} &  \uline{0.365} &  \uline{0.359} & 0.368 & 0.378 & 0.365 & 0.367 & 0.352 & 0.288 & 0.345 & 0.381 & 0.403 & 0.428 & 0.419 & 0.428 & 1.420 & 0.934 & 0.409 & 0.371 \\ 
			\cline{2-23}
		\end{tabular}
	}
 \caption{\textbf{Multivariate long-term forecasting results:} It uses prediction lengths  $T \in \{96, 192, 336, 720\}$ for all the datasets for lookup window 96. The best results are in \textbf{bold} and the second best is \uline{underlined}. This table is adapted from the SiMBA paper \cite{patro2024simba}}
\label{tab:simba_MTS}
 % \vspace{-0.3in}
\end{table*}

\subsection{Recommendation Systems:}
Recommendation systems can be modeled as heterophilic\cite{platonov2022critical} graphs (the opposite of homophilous graphs, where edges tend to connect to nodes of the same class). Graph Neural Networks (GNNs) typically work on homophelic graphs, while GraphMamba\cite{behrouz2024graph} has shown that it can perform well on heterophilic graphs on tasks such as product rating prediction (Nodes are products such as books, music, VDs, DVD, etc., while edges connect products that are frequently bought together). GraphMamba adapts SSMs to work on heterophilic graphs by using steps such as neighborhood tokenization, token ordering, selective scan, local, positional, and structural encoding.  DenseSSM\cite{he2024densemamba} bridges the gap between SSMs and Transformers, offering efficient large language models with improved performance. Large language models (LLMs) based on the Transformer architecture face computational and memory constraints. DenseSSM: A novel approach that enhances information flow between SSM layers by selectively integrating shallow-layer hidden states into deeper layers. State Space Models (SSMs): These offer lower computational complexity but haven’t fully matched Transformer performance. it retains fine-grained information crucial for the final output while maintaining training parallelizability and inference efficiency.

\subsection{Graph Domain:}
Graph-Mamba\cite{wang2024graph} is a novel method that integrates a Mamba block with a node selection mechanism to enhance long-range context modeling in graph networks. The objective of this paper is to address the challenge of modeling long-range dependencies in graph data using efficient methods. The core of Graph-Mamba is the Graph-Mamba block (GMB), which combines the selection mechanism of the Mamba module with a node prioritization approach. The Mamba block is known for its effectiveness in modeling long-range dependencies in sequential data while the node selection mechanism prioritizes and permutes nodes in a graph-centric manner. Graph-Mamba enhances context-aware reasoning by combining Mamba’s selection mechanism with graph-centric strategies. By formulating graph-centric node prioritization and permutation strategies, Graph-Mamba achieves substantial improvements in context-aware reasoning and predictive performance. Notably, it outperforms state-of-the-art methods in long-range graph prediction tasks, all while requiring only a fraction of the computational cost in terms of FLOPs and GPU memory consumption. Spatio Temporal Graph Mamba (STG-Mamba) \cite{li2024stg} is a parallel effort that also leverages Mamba for spatio-temporal graph data modeling.

\subsection{Multimodal Systems:}
In the domain of multi-modal systems integrating both vision and language datasets, recent efforts have focused on enhancing traditional architectures with innovative modules to handle diverse data types effectively. VL-Mamba \cite{qiao2024vl} extends the vanilla Mamba architecture by incorporating a dedicated multi-modal connector module, specifically designed for multi-modal tasks that involve both vision and language data. VL-Mamba incorporates a multi-modal connector module, which consists of a vision selection scan and two linear layers. This module facilitates effective fusion of visual and textual information. VL-Mamba's performance has been rigorously evaluated on standard multi-modal benchmark datasets such as VQA (Visual Question Answering), GQA (Visual Grounding Question Answering), and SQA (Science Question Answering). Comparative analysis demonstrates that VL-Mamba achieves performance levels comparable to other state-of-the-art language and vision models like  standard large language models (LLMs) such as Lava-1.5 and LavaVA-Phi. Notably, VL-Mamba addresses the efficiency challenge posed by the quadratic computation complexity of the Transformer network, which is commonly used in downstream tasks.

In a similar work, Cobra \cite{zhao2024cobra} introduces an extension to the Mamba architecture by incorporating visual information through an image encoder. The goal is to create an efficient multi-modal large language model (MLLM). By integrating visual information through an image encoder and employing a carefully crafted training recipe, Cobra emerges as a powerful multi-modal language model. Evaluation of vision-language benchmarks, including VQA and GQA, showcases Cobra's competitive performance compared to established transformer models like Lava-Phi. The comprehensive analysis presented in Table \ref{tab:vL-results} (adapted from VL-Mamba \cite{qiao2024vl}) provides valuable insights into the efficacy of both transformers and SSMs in the multi-modal domain, guiding further advancements in this burgeoning field.

\subsection{Reinforcement Learning:}
Behavior Cloning (BC) plays a crucial role in online reinforcement learning (RL), as it directly learns the mapping from states to actions from available datasets. However, BC faces challenges when sufficient expert demonstrations are not available. To address this limitation, return-conditioned BC has been introduced as a solution. Decision Transformer (DT)\cite{chen2021decision} is a groundbreaking approach that models reinforcement learning as a sequence modeling task. It leverages return-conditioned BC to enhance RL performance. DT applies Transformer architectures to RL, allowing it to capture complex dependencies in sequences of states, actions, and rewards. The core of DT lies in its causal self-attention network, which adeptly models sequences comprising states, actions, and rewards.

In the pursuit of capturing temporal dependencies and intricate patterns inherent in sequence modeling tasks, Decision Mamba (DMamba) \cite{ota2024decision}  focuses on modeling temporal dependencies and intricate patterns in sequence modeling tasks. Unlike DT, DMamba explores replacing the causal self-attention mechanism with the Mamba framework. DMamba employs Mamba as a token mixing module within a typical transformer architecture, including feed-forward networks and layer normalization.  Extensive evaluations demonstrate DMamba's superior performance over DT across D4RI \cite{fu2020d4rl} benchmarks and Atari \cite{agarwal2020optimistic} datasets. Notably, it showcases enhanced capabilities in capturing complex temporal dependencies, thus showcasing its efficacy in RL tasks. It's worth mentioning that Meta-RL \cite{rimon2024mamba}, although named Mamba, Meta-RL takes a different approach.  It proposes a meta-reinforcement learning strategy rather than being a state space model. Meta-RL aims to improve RL performance by learning across multiple tasks.

% Point mamba\cite{liu2024point}

% Transformers are Multi-State RNNs\cite{oren2024transformers}

\section{State Of The Arts Results}\label{sota}
We consolidate results from multiple sources and discuss the analysis in this section. We include Long Range Arena (LRA) benchmark results first for long sequence processing, as LRA captures text data processing tasks such as ListOps or even vision tasks such as PathFinder \cite{tay2020long}. We also include four text benchmarks, namely, GLUE, Pile and Wikitext and report the results from several SSM papers.  We also have included vision benchmarks, namely the ImageNet classification and instance segmentation with MS Coco dataset. We also include seven-time series benchmark datasets in the performance comparison. 
\subsection{Dataset Details}
\subsubsection{LRA}
We offer additional context and details for each of the LRA (Long Range Aerena) datasets introduced by Tay et al. (2020)\cite{tay2020long} and the Speech Commands dataset introduced by Warden et al. (2018). Our data preprocessing steps align with those outlined by Gu et al. (2021), which we present here for completeness.

\begin{itemize}
\item \texttt{ListOps}: This dataset is an extension of the dataset introduced by Nangia et al. (2018) known as ListOps. It involves evaluating nested mathematical operations, such as \texttt{min} and \texttt{max}, performed on integer operands ranging from zero to nine. The expressions are presented in prefix notation with brackets. The goal is to compute the integer result of the mathematical expression. Each character is encoded as a one-hot vector, with 17 unique values including operators and brackets. Sequences vary in length and are padded with a fixed indicator value to a maximum length of 2,000. A special end-of-sequence token is appended. There are 10 different classes representing the integer result of the expression. The dataset consists of 96,000 training sequences, 2,000 validation sequences, and 2,000 test sequences, without any normalization applied.

\item \texttt{Text}: This dataset is derived from the iMDB sentiment dataset introduced by Maas et al. (2011). It involves classifying movie reviews as positive or negative. Movie reviews are encoded as sequences of integer tokens, with 129 unique values representing characters. Sequences are padded to a maximum length of 4,096, and there are 25,000 training examples and 25,000 test examples. No validation set is provided, and no normalization is applied.

\item \texttt{Retrieval}: This dataset is based on the ACL Anthology network corpus presented by Radev et al. (2009). It aims to classify whether two textual citations are equivalent. Citations are encoded as sequences of integer tokens, and each pair of citations is compressed separately before being passed into a final classifier layer. The decoder head then utilizes the encoded representation to perform the classification task. Characters are encoded into one-hot vectors with 97 unique values. Paired sequences may have unequal lengths, with a maximum length of 4,000. The dataset consists of 147,086 training pairs, 18,090 validation pairs, and 17,437 test pairs, with no normalization applied.

\item \texttt{Image}: This dataset utilizes the CIFAR-10 dataset introduced by Krizhevsky et al. (2009). It involves classifying $32 \times 32$ grayscale CIFAR-10 images into one of ten classes. Images are represented as one-dimensional raster scans of length 1,024. There are 45,000 training examples, 5,000 validation examples, and 10,000 test examples. RGB pixel values are converted to grayscale intensities and normalized to have zero mean and unit variance across the entire dataset.

\item \texttt{Pathfinder}: This dataset is based on the Pathfinder challenge introduced by Linsley et al. (2018). It aims to classify whether a dashed line or path joins the start and end points in a $32 \times 32$ grayscale image. There are two classes indicating the presence or absence of a valid path. Sequences are of equal length (1,024), and there are 160,000 training examples, 20,000 validation examples, and 20,000 test examples. The data is normalized to the range [-1, 1].

\item \texttt{Path-X}: An extension of the Pathfinder challenge where images are $128 \times 128$ pixels, resulting in sequences that are sixteen times longer. Otherwise, it is identical to the Pathfinder challenge.
 
\end{itemize}

\subsubsection{Multi Varitate Time Series  Benchmark Datasets}
The evaluation of the SSM model, conducted across seven benchmark standard datasets commonly used for Multivariate Time Series Forecasting, showcases its robust performance compared to a range of state-of-the-art models. Here's a detailed elaboration: The evaluation was performed on seven benchmark datasets widely used in the field of Multivariate Time Series Forecasting. These datasets cover various domains including Electricity, Weather, Traffic, and four datasets from the Energy Time Series Forecasting (ETT) domain: ETTh1, ETTh2, ETTm1, and ETTm2.

\subsubsection{Video Understanding Datasets}
To evaluate VideoMamba's performance in both short-term and long-term video understanding, we conducted experiments on six diverse datasets:

\begin{itemize}
    \item Short-term Video Understanding:
We assessed VideoMamba's capabilities on scene-related and temporal-related tasks using two widely-used datasets:
\begin{itemize}
    \item Kinetics-400: This dataset consists of videos with an average length of 10 seconds, focusing on scene-related actions.
    \item Something-Something V2: These videos have an average duration of 4 seconds and emphasize temporal-related actions.
\end{itemize}

\item Long-term Video Understanding:
For long-term video comprehension, we conducted rigorous evaluations on three comprehensive datasets:
\begin{itemize}
    \item Breakfast: This dataset comprises 1,712 videos, covering 10 intricate cooking activities spanning 77 hours.
\item COIN: Featuring 11,827 videos across 180 procedural tasks, with an average duration of 2.36 minutes, COIN provides a diverse range of video content for evaluation.

\item Long-form Video Understanding (LVU): LVU is a benchmark consisting of approximately 30,000 movie clips, lasting between 1 to 3 minutes each. It encompasses nine tasks across three primary categories: content understanding, metadata prediction, and user engagement.
\end{itemize}
\end{itemize}
By conducting evaluations on these datasets, we aim to comprehensively assess VideoMamba's performance across various aspects of video understanding, including scene recognition, temporal reasoning, and long-term comprehension of complex video content.

\subsubsection{Multimodal Large Langauge Model Datasets}

We conducted a comprehensive evaluation of our model across a diverse range of 8 benchmark datasets:

\begin{enumerate}
    \item \textbf{VQA-v2 \cite{goyal2017making}}: This dataset focuses on assessing models' capability to comprehend and reason about images and accompanying questions.
    \item \textbf{GQA~\cite{Hudson_2019_CVPR} }: Designed to evaluate spatial understanding and multi-step inference in real-world images.
    \item \textbf{ScienceQA-IMG~\cite{NEURIPS2022_11332b6b}}: Provides multimodal multiple-choice questions on scientific topics, emphasizing common sense reasoning.
    \item \textbf{TextVQA~\cite{singh2019towards}}: Contains questions related to text within images, evaluating the model's optical character recognition (OCR) and inference abilities.
    \item \textbf{POPE ~\cite{li2023evaluating}}: A benchmark for evaluating object hallucinations, involving a binary classification task determining the presence of objects.
    \item \textbf{MME ~\cite{yin2023survey}}: Evaluates perceptual and cognitive abilities, including OCR, object recognition, common sense reasoning, numerical calculations, text translation, and code reasoning.
    \item \textbf{MMBench~\cite{liu2023mmbench}}: Features 3,000 single-choice questions across 20 dimensions, utilizing a CircularEval strategy for robust evaluation, with ChatGPT matching model predictions to choices.
    \item \textbf{MM-Vet~\cite{yu2023mm}}: Identifies 16 emergent tasks from core visual and linguistic (VL) capabilities, including Recognition, Knowledge, OCR, Spatial awareness, Language generation, and Math.
\end{enumerate}

Each dataset presents unique challenges and evaluates various aspects of multimodal understanding, ranging from basic object recognition to complex reasoning and inference tasks across images and text.

\subsection{Results}
\subsubsection{LRA}
We adopted the table from the HGRN paper \cite{qin2023hierarchically} and the S5 paper \cite{smith2022simplified} and consolidated the results into the table below \ref{tab:lra}. This result shows that transformers such as Local Attention, Sparse Attention, LongFormer, LinFormer, Reformer, Sinkhorn, Synthesizer, BigBird, Linear Transformer, Performer, CosFormer, FNet, NystromFormer, Luna, H-Transformer-ID and CCNN do not perform too well on LRA benchmarks, on many tasks ListOps, Text, Retrieval, Image, Pathfinder and Path-X - this is due to the quadratic complexity of attention and lack of inductive bias of transformers. State space models outperform transformers on LRA benchmarks. Among SSMs, we found that S5 and Mega are the top-performing SSMs across all the tasks. S4, one of the pioneering models, is outperformed by its variants such as DSS, S4nd, Liquid-S4, etc. TNN, HGRN and LRU are in the middle and outperform S4 and its variants. However, the exact reason as to why S5 and Mega should be top-performing SSMs remains to be explored. One could possibly analyze the SSMs from an explainability perspective to get to the bottom of these results, which we leave for future work. 

\subsubsection{Language Domain}
We show results from three benchmarks, including GLUE captured in table \ref{exp:glue_benchmark}. Glue is a combination of six tasks, MNLI, QNLI, QQP, SST2, MRPC, and Cola.  The table is divided into four categories, namely, attention based transformers, MLP-based networks, FFT based transformers, and state space models. We find that SSMs have a performance gap with all three architectures, while MLP-based networks and FFT based transformers get better scores than SSMs. The attention-based transformers are the top performers on this benchmark, with the Toeplitz Neural Network (TNN) being the only exception, with TNN outperforming all including attention-based transformers. 

Similarly, another benchmark we use for the language domain evaluation is the WikiText data \cite{merity2016pointer}. The table below \ref{tab:wikitext} captures the perplexity scores of the different architectures and models. We adopted the table from the TNN paper \cite{qin2022toeplitz} and HGRN paper \cite{qin2023hierarchically} and consolidated the results into the table below \ref{tab:wikitext}. The perplexity score measures the ability of a model to make a next-word prediction, with lower the perplexity score the better. This table captures results on validation and test sets. The table is divided into attention-based transformers, MLP-based networks, and SSMs.  Again, it can be observed that attention-based transformers outperform other architectures, including SSMs. However, the HGRN SSM outperforms all others and has the lowest perplexity scores. 

The last benchmark in the language domain is the Pile benchmark. Pile is an 825GB benchmark that comprises several smaller datasets combined together, namely, PubMed Central, ArXiv, GitHub, the FreeLaw Project, Stack Exchange, the US Patent and Trademark Office, PubMed, Ubuntu IRC, HackerNews, YouTube, PhilPapers, and NIH ExPorter. We give below the results from the leaderboard of the Pile benchmark and other sources in table \ref{tab:pile}. We find that HGRN is the only SSM that performs comparably with attention-based transformers, especially if they have been trained on the Pile datasets.

%%%%%%%%%%%%%%%%%%%%%%%%%%%%%%%%%%%%%%%%%%%%%%%%%%%%%%%%%%%%%%%%%%%%%%%%%%%%%%%%%%%%%%%%%%%%%%%%%%%%%%%%%%%%%%%%%%%%%%%%%%%%%%%%%%%%%%%%%%%%%%%%%%%%%%%%%%%%%%

\subsubsection{Image  Domain}

We show evaluation on the ImageNet-1K dataset showcases the performance of various vision backbones across different architectural categories and computational complexities as reported in table-\ref{tab:imagenet_sota}. Among CNNs, RegNetY-8G and RegNetY-16G stand out with accuracies of 81.7\% and 82.9\%, respectively. These results underscore the effectiveness of SSM models in capturing image representations, with LocalVMamba-S and PlainMamba-L3 showcasing competitive performance. Additionally, transformer models such as Swin-T and EffNet-B4 demonstrate strong performance, while RegNetY-8G and RegNetY-16G emerge as top performers among CNNs.  In comparison, transformer models such as Swin-T and EffNet-B4 achieve accuracies of 81.3\% and 82.9\%, respectively.  Notably, RegNetY-8G and RegNetY-16G stand out among convnets, achieving top-1 accuracies of 81.7\% and 82.9\%, respectively. Among transformers, SpectFormer-H-B, SVT-H-B, and and SCT-H-B exhibit impressive performance, achieving top-1 accuracies of 85.1\%, 85.2\%, and 85.2\%, respectively. Among SSM models, LocalVMamba-S demonstrates strong performance with a top-1 accuracy of 83.7\%, while PlainMamba-L3 achieves competitive results with an accuracy of 82.3\%. Vim-Ti and VMamba-T also exhibit noteworthy performance, achieving accuracies of 76.1\% and 82.2\%, respectively. For SSMs, SiMBA-S(MLP) emerges as a strong contender, achieving a top-1 accuracy of 84.0\%, while SiMBA-B(MLP) and SiMBA-L(EinFFT) attain top-1 accuracies of 84.7\% and 83.9\%, respectively. These results underscore the effectiveness of diverse architectural approaches across different computational complexities in addressing image recognition tasks.

\begin{table}[htb]
\centering
\scriptsize
% \tiny
\caption{\textbf{SSM SOTA on ImageNet-1K} This table shows the performance of various SSM models for Image Recognition tasks on the ImageNet1K\cite{deng2009imagenet} dataset. We have grouped the vision models into three categories based on their GFLOPs (Small, Base, and Large). The GFLOP ranges: Small (GFLOPs$<$5), Base (5$\leq$GFLOPs$<$10), and Large (10$\leq$GFLOPs$<$30).}
\label{tab:imagenet_sota_ssm}
\begin{tabular}{c|ccc|c}
\toprule
Method & Image Size & \#Param. & FLOPs  & Top-1 acc. \\

\toprule
\multicolumn{4}{c}{\textbf{SSMs}} \\

\midrule

HyenaViT-B~\cite{fu2024monarch}  &224$^2$ &88M &- & 78.5 \\
S4ND-ViT-B~\cite{nguyen2022s4nd} & 224$^2$ & 89M & -  & 80.4 \\

TNN-T\cite{qin2022toeplitz} & - &  6.4M & -  & 72.29\\
TNN-S\cite{qin2022toeplitz} & - &  23.4M & - & 79.20\\

Vim-Ti\cite{zhu2024vision} &$224^{2}$ & 7M&- & 76.1 \\ 
Vim-S\cite{zhu2024vision} & $224^{2}$ & 26M &-& 80.5 \\

HGRN-T\cite{qin2024hierarchically} & - &  6.1M & - &  74.40 \\
HGRN-S\cite{qin2024hierarchically} & - &  23.7M & - &80.09 \\

PlainMamba-L1 ~\cite{yang2024plainmamba} & 224$^2$& 7M &3.0G& 77.9 \\
PlainMamba-L2 ~\cite{yang2024plainmamba} &  224$^2$ &25M& 8.1G &81.6\\
PlainMamba-L3 ~\cite{yang2024plainmamba} &  224$^2$& 50M& 14.4G& 82.3\\

Mamba-2D-S  ~\cite{li2024mamba}&  224 $^2$& 24M&- & 81.7\\
Mamba-2D-B ~\cite{li2024mamba} & 224$^2$&  92M&- &  83.0\\

VMamba-T\cite{liu2024vmamba} & 224$^2$ & 22M & 5.6G  & 82.2 \\
VMamba-S\cite{liu2024vmamba} & 224$^2$ & 44M & 11.2G  & 83.5 \\
VMamba-B\cite{liu2024vmamba} & 224$^2$ & 75M & 18.0G & 83.2 \\

LocalVMamba-T ~\cite{huang2024localmamba}  & 224$^2$& 26M & 5.7G& 82.7\\
LocalVMamba-S ~\cite{huang2024localmamba} &  224$^2$ &50M & 11.4G& 83.7\\

SiMBA-S(Monarch) ~\cite{patro2024simba} & 224$^2$ & 18.5M & 3.6G  & 81.1 \\
SiMBA-B(Monarch)  ~\cite{patro2024simba} & 224$^2$ & 26.9M & 6.3G  & 82.6 \\
SiMBA-L(Monarch) ~\cite{patro2024simba} & 224$^2$ & 42M & 10.7G & 83.8 \\

ViM2-T  ~\cite{behrouz2024mambamixer} & 224$^2$ & 20M &- &82.7\\
ViM2-S  ~\cite{behrouz2024mambamixer} & 224$^2$ & 43M&- & 83.7\\
ViM2-B  ~\cite{behrouz2024mambamixer} & 224$^2$ & 74M&- & 83.9\\

SiMBA-S(EinFFT)  ~\cite{patro2024simba} & 224$^2$ & 15.3M &2.4G  & 81.7 \\
SiMBA-B(EinFFT)  ~\cite{patro2024simba} & 224$^2$ & 22.8M &5.2G  & 83.5 \\
\rowcolor{gray!20}SiMBA-L(EinFFT)  ~\cite{patro2024simba} & 224$^2$ & 36.6M & 9.6G & 84.4 \\

\rowcolor{gray!20}SiMBA-S(MLP)  ~\cite{patro2024simba} & 224$^2$ & 26.5M & 5.0G & 84.0 \\
\rowcolor{gray!20}SiMBA-B(MLP)  ~\cite{patro2024simba} & 224$^2$ & 40.0M & 9.0G  & 84.7 \\
% SiMBA-L(MLP)$^\dagger$ ~\cite{patro2024simba} & 224$^2$ & 66.6M & 16.3G & 49.4 \\
\bottomrule
\end{tabular}
\end{table}

The table-\ref{tab:imagenet_sota_ssm} presents the state-of-the-art (SOTA) performance of various Structured State Space Models (SSMs) on the ImageNet-1K dataset for image classification tasks. These SSM models are grouped based on their computational complexity measured in GFLOPs (Small, Base, and Large). Notable models include HyenaViT-B, S4ND-ViT-B, Vim-S, HGRN-S, PlainMamba-L3, Mamba-2D-B, VMamba-S, LocalVMamba-S, SiMBA-L(Monarch), ViM2-B, and SiMBA-L(EinFFT), achieving top-1 accuracy ranging from 78.5\% to 84.5\%. These models demonstrate competitive performance while varying in parameter counts and computational complexity, showcasing the effectiveness of SSMs for image classification tasks with improved computational efficiency.

\begin{table*}[htb]
\caption{\textbf{Comparison with SoTA methods on 8 benchmarks.} 
Benchmark names are abbreviated due to space limits. VQA-v2~\cite{goyal2017making}; GQA~\cite{Hudson_2019_CVPR}; SQA$^\text{I}$: ScienceQA-IMG~\cite{NEURIPS2022_11332b6b}; VQA$^\text{T}$: TextVQA~\cite{singh2019towards}; POPE~\cite{li2023evaluating}; MME~\cite{yin2023survey}; MMB: MMBench~\cite{liu2023mmbench}; MM-Vet~\cite{yu2023mm}.
PT and IT indicate the number of samples in the pretraining and instruction tuning stages, respectively.  This table is adapted from VL-Mamba\cite{qiao2024vl} paper.
}
\label{tab:vL-results}
\centering
\vspace{3pt}
\renewcommand{\arraystretch}{1.25}
%\scalebox{0.5}{
\resizebox{\linewidth}{!}{
\begin{tabular}{ll cc | cccc | cccc }
\toprule
Method & LLM  & PT & IT & VQA$^\text{v2}$ & GQA & SQA$^\text{I}$ & VQA$^\text{T}$ & POPE & MME & MMB & MM-Vet \\
\midrule
BLIP-2~\cite{li2023blip} & Vicuna-13B  & 129M & - & 41.0 & 41.0 & 61.0 & 42.5 & {85.3} & 1293.8 & -- & 22.4 \\
MiniGPT-4~\cite{zhu2023minigpt}&Vicuna-7B&5M&5K&-&32.2&-&-&-&581.7&23.0&-\\
InstructBLIP~\cite{dai2024instructblip} & Vicuna-7B & 129M & 1.2M & -- & 49.2 & 60.5 & 50.1 & -- & -- & 36 & 26.2 \\ 
InstructBLIP~\cite{dai2024instructblip} & Vicuna-13B  & 129M & 1.2M & -- & 49.5 & 63.1 & 50.7 & 78.9 & 1212.8 & -- & 25.6 \\ 
Shikra~\cite{chen2023shikra} & Vicuna-13B  & 600K & 5.5M & 77.4 & -- & -- & -- & -- & -- & 58.8 & -- \\ 
Otter~\cite{li2023mimic} & LLaMA-7B & -& -&-&-&-&-&-& 1292.3 & 48.3 & 24.6\\
mPLUG-Owl~\cite{ye2023mplug} & LLaMA-7B  & 2.1M & 102K & -&-&-&-&-&967.3&49.4&-\\
IDEFICS-9B~\cite{laurenccon2024obelics} & LLaMA-7B & 353M & 1M & 50.9 & 38.4 & -- & 25.9 & -- & -- & 48.2 & -- \\ 
IDEFICS-80B~\cite{laurenccon2024obelics} & LLaMA-65B & 353M & 1M & 60.0 & 45.2 & -- & 30.9 & -- & -- & 54.5 & -- \\
Qwen-VL~\cite{bai2023qwen} & Qwen-7B  & 1.4B & 50M & 78.8 & 59.3 & 67.1 & 63.8 & -- & -- & 38.2 & -- \\ 
Qwen-VL-Chat~\cite{bai2023qwen}  & Qwen-7B & 1.4B & 50M & 78.2 & 57.5 & 68.2 & 61.5 & -- & 1487.5 & 60.6 & -- \\ 
LLaVA-1.5~\cite{liu2024visual} & Vicuna-7B & 558K&665K&78.5&62.0&66.8&58.2&85.9&1510.7&64.3&30.5\\
LLaVA-1.5~\cite{liu2024visual} & Vicuna-13B & 558K&665K&80.0&63.3&71.6&61.3&85.9&1531.3&67.7&35.4\\
\midrule
LLaVA-Phi~\cite{zhu2024llava} & Phi-2-2.7B & 558K& 665K& 71.4 & - &{68.4} & 48.6 & 85.0 & 1335.1 & {59.8} & 28.9\\
MobileVLM-3B~\cite{chu2023mobilevlm} & MobileLLaMA-2.7B  & 558K & 665K& - & {59.0} & 61.2 &47.5 & 84.9 & 1288.9 & 59.6 & -\\
\midrule
Cobra\cite{zhao2024cobra}& Mamba-2.8B& -& -&  75.9& 58.5& -& 46.0 &88.0& -& -&-\\

VL-Mamba\cite{qiao2024vl}& Mamba LLM{-2.8B} &558K & 665K&{76.6} & 56.2 & 65.4 & {48.9} &{84.4} &{1369.6}& 57.0& {32.6}\\
\bottomrule
\end{tabular}}
\end{table*}
\subsubsection{Multimodal Applications}
VL-Mamba, achieves notable performance across various benchmarks as shown in table-\ref{tab:vL-results}. With similar multimodal training data, VL-Mamba outperforms SQAI (65.4 vs. 61.2), VQAT (48.9 vs. 47.5), and MME (1369.6 vs. 1288.9). Remarkably, despite having fewer pretrained tokens (627B) compared to MobileVLM's backbone MobileLLaMA (1.3T), VL-Mamba exhibits superior performance. For instance, compared to LLaVA-Phi with the Phi-2-2.7B language model, VL-Mamba demonstrates better performance on VQA-v2 (76.6 vs. 71.4), MME (1369 vs. 1335.1), and MM-Vet (32.6 vs. 28.9). These results underscore the efficacy of VL-Mamba in multimodal learning tasks and highlight the potential of leveraging state-space models for such applications
\begin{table*}[thb]
\centering
\fontsize{9pt}{9pt}\selectfont
\centering
\caption{Univariate long sequence time-series forecasting results on four datasets (five cases). This table is adapted from the S4 paper \cite{gu2021efficiently}}
\label{tab:informer-s}
\resizebox{\linewidth}{!}{
\begin{tabular}{c|c|c|c|c|c|c|c|c|c|c|c}
\toprule[1.0pt]
\multicolumn{2}{c|}{Methods}              & \textbf{S4} & {Informer}                     & {Informer$^{\dag}$}            & {LogTrans}       & {Reformer}   & {LSTMa}      & {DeepAR}     & {ARIMA}                 & {Prophet}    \\
\midrule[0.5pt]
\multicolumn{2}{c|}{Metric}               & MSE~~MAE              & MSE~~MAE                       & MSE~~MAE                       & MSE~~MAE         & MSE~~MAE     & MSE~~MAE     & MSE~~MAE     & MSE~~MAE                & MSE~~MAE     \\
\midrule[1.0pt]
\multirow{5}{*}{\rotatebox{90}{ETTh$_1$}} & 24                    & \textbf{0.061}~~\textbf{0.191} & 0.098~~0.247                   & {0.092}~~{0.246} & 0.103~~0.259 & 0.222~~0.389 & 0.114~~0.272 & 0.107~~0.280            & 0.108~~0.284  & 0.115~~0.275 \\
                                          & 48                    & \textbf{0.079}~~\textbf{0.220} & {0.158}~~{0.319}               & 0.161~~0.322     & 0.167~~0.328 & 0.284~~0.445 & 0.193~~0.358 & 0.162~~0.327            & 0.175~~0.424  & 0.168~~0.330 \\
                                          & 168                   & \textbf{0.104}~~\textbf{0.258} & {0.183}~~{0.346}               & 0.187~~0.355     & 0.207~~0.375 & 1.522~~1.191 & 0.236~~0.392 & 0.239~~0.422            & 0.396~~0.504  & 1.224~~0.763 \\
                                          & 336                   & \textbf{0.080}~~\textbf{0.229} & 0.222~~0.387                   & {0.215}~~{0.369} & 0.230~~0.398 & 1.860~~1.124 & 0.590~~0.698 & 0.445~~0.552            & 0.468~~0.593  & 1.549~~1.820 \\
                                          & 720                   & \textbf{0.116}~~\textbf{0.271} & 0.269~~0.435                   & {0.257}~~{0.421} & 0.273~~0.463 & 2.112~~1.436 & 0.683~~0.768 & 0.658~~0.707            & 0.659~~0.766  & 2.735~~3.253 \\
\midrule[0.5pt]
\multirow{5}{*}{\rotatebox{90}{ETTh$_2$}} & 24                    & 0.095~~0.234                   & \textbf{0.093}~~\textbf{0.240} & 0.099~~0.241     & 0.102~~0.255 & 0.263~~0.437 & 0.155~~0.307 & 0.098~~0.263            & 3.554~~0.445  & 0.199~~0.381 \\
                                          & 48                    & 0.191~~0.346                   & \textbf{0.155}~~\textbf{0.314} & 0.159~~0.317     & 0.169~~0.348 & 0.458~~0.545 & 0.190~~0.348 & 0.163~~0.341            & 3.190~~0.474  & 0.304~~0.462 \\
                                          & 168                   & \textbf{0.167}~~\textbf{0.333} & {0.232}~~{0.389}               & 0.235~~0.390     & 0.246~~0.422 & 1.029~~0.879 & 0.385~~0.514 & 0.255~~0.414            & 2.800~~0.595  & 2.145~~1.068 \\
                                          & 336                   & \textbf{0.189}~~\textbf{0.361} & 0.263~~{0.417}                 & {0.258}~~0.423   & 0.267~~0.437 & 1.668~~1.228 & 0.558~~0.606 & 0.604~~0.607            & 2.753~~0.738  & 2.096~~2.543 \\
                                          & 720                   & \textbf{0.187}~~\textbf{0.358} & {0.277}~~{ 0.431}              & 0.285~~0.442     & 0.303~~0.493 & 2.030~~1.721 & 0.640~~0.681 & 0.429~~0.580            & 2.878~~1.044  & 3.355~~4.664 \\
\midrule[0.5pt]
\multirow{5}{*}{\rotatebox{90}{ETTm$_1$}} & 24                    & \textbf{0.024}~~\textbf{0.117} & {0.030}~~{0.137}               & 0.034~~0.160     & 0.065~~0.202 & 0.095~~0.228 & 0.121~~0.233 & 0.091~~0.243            & 0.090~~0.206  & 0.120~~0.290 \\
                                          & 48                    & \textbf{0.051}~~\textbf{0.174} & 0.069~~0.203                   & {0.066}~~{0.194} & 0.078~~0.220 & 0.249~~0.390 & 0.305~~0.411 & 0.219~~0.362            & 0.179~~0.306  & 0.133~~0.305 \\
                                          & 96                    & \textbf{0.086}~~\textbf{0.229} & 0.194~~{0.372}                 & {0.187}~~0.384   & 0.199~~0.386 & 0.920~~0.767 & 0.287~~0.420 & 0.364~~0.496            & 0.272~~0.399  & 0.194~~0.396 \\
                                          & 288                   & \textbf{0.160}~~\textbf{0.327} & {0.401}~~0.554                 & 0.409~~{0.548}   & 0.411~~0.572 & 1.108~~1.245 & 0.524~~0.584 & 0.948~~0.795            & 0.462~~0.558  & 0.452~~0.574 \\
                                          & 672                   & \textbf{0.292}~~\textbf{0.466} & {0.512}~~{0.644}               & 0.519~~0.665     & 0.598~~0.702 & 1.793~~1.528 & 1.064~~0.873 & 2.437~~1.352            & 0.639~~0.697  & 2.747~~1.174 \\
\midrule[0.5pt]
\multirow{5}{*}{\rotatebox{90}{Weather}}  & 24                    & 0.125~~0.254                   & \textbf{0.117}~~\textbf{0.251} & 0.119~~0.256     & 0.136~~0.279 & 0.231~~0.401 & 0.131~~0.254 & 0.128~~0.274            & 0.219~~0.355  & 0.302~~0.433 \\
                                          & 48                    & 0.181~~\textbf{0.305}          & \textbf{0.178}~~0.318          & 0.185~~0.316     & 0.206~~0.356 & 0.328~~0.423 & 0.190~~0.334 & 0.203~~0.353            & 0.273~~0.409  & 0.445~~0.536 \\
                                          & 168                   & \textbf{0.198}~~\textbf{0.333} & {0.266}~~{0.398}               & 0.269~~0.404     & 0.309~~0.439 & 0.654~~0.634 & 0.341~~0.448 & 0.293~~0.451            & 0.503~~0.599  & 2.441~~1.142 \\
                                          & 336                   & 0.300~~0.417                   & \textbf{0.297}~~\textbf{0.416} & 0.310~~0.422     & 0.359~~0.484 & 1.792~~1.093 & 0.456~~0.554 & 0.585~~0.644            & 0.728~~0.730  & 1.987~~2.468 \\
                                          & 720                   & \textbf{0.245}~~\textbf{0.375} & {0.359}~~{0.466}               & 0.361~~0.471     & 0.388~~0.499 & 2.087~~1.534 & 0.866~~0.809 & 0.499~~0.596            & 1.062~~0.943  & 3.859~~1.144 \\
\midrule[0.5pt]
\multirow{5}{*}{\rotatebox{90}{ECL}}      & 48                    & 0.222~~\textbf{0.350}          & 0.239~~0.359                   & 0.238~~0.368     & 0.280~~0.429 & 0.971~~0.884 & 0.493~~0.539 & \textbf{0.204}~~{0.357} & 0.879~~0.764  & 0.524~~0.595 \\
                                          & 168                   & 0.331~~\textbf{0.421}          & 0.447~~0.503                   & 0.442~~0.514     & 0.454~~0.529 & 1.671~~1.587 & 0.723~~0.655 & \textbf{0.315}~~{0.436} & 1.032~~0.833  & 2.725~~1.273 \\
                                          & 336                   & \textbf{0.328}~~\textbf{0.422} & 0.489~~0.528                   & 0.501~~0.552     & 0.514~~0.563 & 3.528~~2.196 & 1.212~~0.898 & {0.414}~~{0.519}        & 1.136~~0.876  & 2.246~~3.077 \\
                                          & 720                   & \textbf{0.428}~~\textbf{0.494} & {0.540}~~{0.571}               & 0.543~~0.578     & 0.558~~0.609 & 4.891~~4.047 & 1.511~~0.966 & 0.563~~0.595            & 1.251~~0.933  & 4.243~~1.415 \\
                                          & 960                   & \textbf{0.432}~~\textbf{0.497} & {0.582}~~{0.608}               & 0.594~~0.638     & 0.624~~0.645 & 7.019~~5.105 & 1.545~~1.006 & 0.657~~0.683            & 1.370~~0.982  & 6.901~~4.264 \\

\midrule[1.0pt]
% \multicolumn{2}{c|}{Count}                & {22}                  & {5}                            & {0}                            & {0}              & {0}          & {0}          & {2}          & {0}                     & {0}          \\
\bottomrule[1.0pt]

\end{tabular}%
}

\end{table*}

\subsubsection{Time Series Forecasting }
\label{exp:mts}

The evaluation of the SSM model, conducted across seven benchmark standard datasets commonly used for Multivariate Time Series Forecasting, showcases its robust performance compared to a range of state-of-the-art models.  The evaluation compared SiMBA against several state-of-the-art models, including Transformer-based methods such as PatchTST, CrossFormer, FEDFormer, ETSFormer, PyraFormer, and AutoFormer. It also included comparisons with CNN-based models like TimeNet, graph-based approaches such as MTGNN, and MLP-based models like DLinear. The performance of SiMBA and the comparison models was assessed using common evaluation metrics for time series forecasting tasks. These metrics typically include Mean Squared Error (MSE) and Mean Absolute Error (MAE), which are standard measures for quantifying the accuracy of forecasting models. SiMBA consistently outperformed existing state-of-the-art models across the evaluated datasets in terms of MSE and MAE. This indicates that SiMBA is adept at capturing temporal patterns and making accurate predictions across various time series forecasting tasks and modalities. The superior performance of SiMBA underscores its adaptability and efficacy in handling diverse time series forecasting challenges, establishing it as a prominent model in the field. Overall, the extensive evaluation of SiMBA across multiple benchmark datasets and comparison with state-of-the-art models demonstrates its robust performance and effectiveness in addressing a wide range of time series forecasting tasks. The findings highlight SiMBA as a promising model for practical applications in various domains requiring accurate and reliable time-series predictions.

\subsubsection{Video  Domain}

The table-\ref{tab:results_k400} provides a comparison of various models' performance on scene-related Kinetics-400 dataset. It includes architectures such as CNN, transformers (Trans.), and state space models (SSM), along with their corresponding models, input data characteristics, number of parameters, and FLOPs (floating point operations).  For each architecture type, the table lists different models, indicating whether they are supervised or self-supervised. The models are evaluated based on their performance in terms of Top-1 and Top-5 accuracy on the Kinetics-400 dataset. Additionally, the table highlights whether the models utilize isotropic architecture without downsampling layers, extra data, and pretrained teacher models.  Supervised CNN models like SlowFast, X3D-M, and X3D-XL achieve competitive performance. Transformer models such as Swin-T and Swin-B demonstrate strong performance, especially when trained with additional data from IN-21K.  Models combining CNN and Transformer components (CNN+Trans.) like MViTv1-B and UniFormer-B exhibit promising results. State space models (SSM) such as VideoMamba show competitive performance, with VideoMamba-M outperforming other models in terms of Top-1 and Top-5 accuracy, particularly when trained with extra data from CLIP-400M.

Table-\ref{tab:results_ssv2} presents a comparison of various models' performance on the temporal-related SthSth V2 dataset. It includes architectures such as CNN, transformers (Trans.), and state space models (SSM), along with their corresponding models, input data characteristics, number of parameters, and FLOPs (floating point operations). CNN models like SlowFast$_{R101}$, CT-Net$_{R50}$, and TDN$_{R50}$ demonstrate moderate performance on the SthSth V2 dataset. The transformer model Swin-B achieves competitive performance compared to CNN models. Models combining CNN and transformer components (CNN+Trans.) like MViTv1-B and UniFormer-B show promising results, with UniFormer-B outperforming other models in terms of Top-1 accuracy. Transformer models such as TimeSformer-HR and ViViT-L also exhibit reasonable performance on the dataset.  State space models (SSM) like VideoMamba demonstrate competitive performance, with VideoMamba-M showing the highest Top-1 accuracy among SSMs. Self-supervised transformer models like VideoMAE-B$_{2400e}$ and UMT-B$_{800e}$ perform well, particularly when utilizing the CLIP-400M pretrained teacher model. VideoMamba-M$_{800e}$, a state space model trained with CLIP-400M, achieves the highest Top-1 accuracy among SSMs.

The table-\ref{tab:results_lvu} compares various methods' performance on LVU (Large-scale Video Understanding) using different evaluation metrics related to content, metadata, and user engagement.  Methods like VideoBERT, Object Trans., LST, Performer, Orthoformer, ViS4mer, and VideoMamba$_{f32}$ employ different architectures and backbones to understand video content. VideoMamba$_{f32}$ achieves the highest scores in content-related metrics, particularly in scene understanding.  Metadata understanding involves attributes such as director, genre, writer, and release year. VideoMamba$_{f32}$ outperforms other methods in most metadata-related metrics, indicating its ability to extract useful information from video metadata. Some methods, like VideoMamba$_{f32}$, are end-to-end (e2e) approaches, which do not rely on exhausting feature extraction separately. This integrated approach helps in better understanding and utilizing both content and metadata for user engagement prediction.
\begin{table}[htb]
    % \vspace{-0.3cm}
    \centering
    \caption{\textbf{Comparison with the state-of-the-art on Breakfast and COIN.} ``\textit{e2e}'' means end-to-end methods without exhausting feature extraction.``\color{red}{$\dag$} '' marks the backbone with masked pretraining. This table is adapted from VideoMamba\cite{li2024videomamba} paper.
    }
    \label{tab:results_breakfast_coin}
    \setlength\tabcolsep{2pt}
    \resizebox{0.95\linewidth}{!}{
        \begin{tabular}{l|c|l|l|l|cc}
        \Xhline{1.0pt}
            \multirow{2}*{\textbf{Method}} & \multirow{2}*{\textit{\textbf{e2e}}} & \multirow{2}*{\textbf{Backbone}} & \multirow{2}*{\textbf{Neck Type}} & \textbf{Pretraining}  & \textbf{BF} & \textbf{COIN} \\
            ~ & ~ & ~ & ~ & \textbf{Dataset} & \textbf{Top-1} & \textbf{Top-1} \\
            \Xhline{0.8pt}
             Timeception~\cite{hussein2019timeception} & \xmark & 3D-ResNet & Conv. & IN-1K+K400 & 71.3 & - \\
             VideoGraph~\cite{hussein2019videograph} & \xmark & I3D & Conv.+Atten. & IN-1K+K400 & 69.5 & - \\
             GHRM~\cite{zhou2021graph} & \xmark & I3D & Graph Conv.. & IN-1K+K400 & 75.5 & - \\
             % TimeSformer~\cite{distant} & \xmark & TimeSformer & Attention & K400 & 81.1 & 83.5 \\
             % Distant Supervision~\cite{distant} & \xmark & TimeSformer & Atten. & IN-21K+HTM & 88.7 & 88.9 \\
             Distant Supervision~\cite{lin2022learning} & \xmark & TimeSformer & Atten. w/ KB & IN-21K+HTM & \textbf{89.9} & \textbf{90.0} \\
             ViS4mer~\cite{islam2022long} & \xmark & Swin-B & SSM & IN-21K+K600 & \underline{88.2} & \underline{88.4} \\
             \Xhline{0.8pt}
             Turbo$_{f32}$~\cite{han2022turbo} & \cmark & VideoMAE-B & - & K400 & 86.8 & 82.3 \\
             Turbo$_{f32}$~\cite{han2022turbo} & \cmark & VideoMAE-B &-  & K400+HTM-AA & \underline{91.3} & \underline{87.5} \\
             
             VideoMamba$_{f32}$ ~\cite{li2024videomamba}& \cmark & VideoMamba-Ti & - & K400 & 94.3 & 86.2 \\
             
             VideoMamba$_{f64}$ ~\cite{li2024videomamba}& \cmark & VideoMamba-Ti & - & K400 & 94.3 & 87.0 \\
             
             VideoMamba$_{f32}$~\cite{li2024videomamba} & \cmark & VideoMamba-S & - & K400 & 95.3 & 88.4 \\
             
             VideoMamba$_{f64}$~\cite{li2024videomamba} & \cmark & VideoMamba-S & - & K400 & 97.4 & 88.7 \\
             
             VideoMamba$_{f32}$ ~\cite{li2024videomamba}& \cmark & VideoMamba-M &  -& K400 & 94.8 & 88.3 \\
             
             VideoMamba$_{f64}$~\cite{li2024videomamba} & \cmark & VideoMamba-M & - & K400 & 95.8 & 89.5 \\
             
             VideoMamba$_{f32}$~\cite{li2024videomamba} & \cmark & VideoMamba-M\color{red}{$\dag$} &-  & K400 & \textbf{97.9} & 89.6 \\
             
             VideoMamba$_{f64}$ ~\cite{li2024videomamba}& \cmark & VideoMamba-M\color{red}{$\dag$} &  -& K400 & 96.9 & \textbf{90.4} \\
        \Xhline{1.0pt}	
        \end{tabular}
    }
    \vspace{-0.5cm}
\end{table}

The table-\ref{tab:results_breakfast_coin} presents a comparison of various methods' performance on the Breakfast (BF) and COIN datasets, focusing on end-to-end (e2e) approaches and different backbone architectures. The existing methods such as Timeception, VideoGraph, and GHRM are some of the existing methods that utilize different backbone architectures and pretraining datasets to achieve performance on BF and COIN datasets. Distant Supervision achieves the highest performance among existing methods, leveraging TimeSformer with attention and knowledge base (KB) pretraining.
 Turbo$_{f32}$ is an end-to-end approach using VideoMAE-B backbone, achieving competitive performance on both datasets.  VideoMamba$_{f32}$ and VideoMamba$_{f64}$ employ different variants of the VideoMamba backbone (Ti, S, M) and achieve high accuracy on both BF and COIN datasets.
 VideoMamba$_{f32}$ and VideoMamba$_{f64}$ with a masked pretraining backbone (marked with $\dag$) achieve even higher performance, especially VideoMamba$_{f32}$, which achieves the highest accuracy on both datasets.  VideoMamba backbone variants (Ti, S, M) with different precision (f32, f64) are evaluated, showing that higher precision (f64) generally leads to better performance. Among all variants, VideoMamba-M with f32 precision and a masked pretraining backbone achieves the highest accuracy on the BF dataset, while VideoMamba-M with f64 precision performs the best on the COIN dataset.

\section{Conclusions}
This paper has provided a survey of state space models for sequence processing - how they evolved from RNNs and how they are now competing with transformers on several language, vision, time series, video, audio tasks, and benchmarks. While there are still some tasks like copying and retrieving information from the context where transformers outperform SSMs, SSMs have lowered the performance gap with state-of-the-art transformers in several domains. We have classified the SSMs into three buckets, namely, the structured, gated, and recurrent - we have also identified the foundational models in each category and discussed the core contributions made by each. In addition, we have also consolidated the performance benchmarking of SSMs and compared them to state-of-the-art transformers in various domains and benchmark datasets. This has opened up the research possibilities in SSMs and closed the performance gap with transformers in different domains. For instance, the SiMBA combines transformers and Mamba architecturally and achieves state-of-the-art results on standard time series and vision benchmark datasets. Similarly, the other foundational SSMs can also be combined with the transformers to outperform state-of-art transformers. We also identify that scaling the SSMs to large network sizes is still an open issue, especially with Mamba which has certain stability issues at scale. The stability of SSMs at large network sizes is an open issue, especially in computer vision. Another possible open research problem is to solve advanced in-context learning tasks - though transformers and state space models have been trained to perform certain in-context learning tasks.

%%%%%%%%%%%%%%%%%%%%%%%%%%%%%%%%%%%%%%%%%%%%%%%%%%%%%%%%%%%%

% {\small
% \bibliographystyle{iccv2023AuthorKit/ieee_fullname}
% \bibliography{egbib}
% }
% \clearpage
% {\small 
\bibliographystyle{plain}
\bibliography{egbib}

\begin{thebibliography}{100}

\bibitem{agarwal2020optimistic}
Rishabh Agarwal, Dale Schuurmans, and Mohammad Norouzi.
\newblock An optimistic perspective on offline reinforcement learning.
\newblock In {\em International Conference on Machine Learning}, pages 104--114. PMLR, 2020.

\bibitem{ahamed2024mambatab}
Md~Atik Ahamed and Qiang Cheng.
\newblock Mambatab: A simple yet effective approach for handling tabular data.
\newblock {\em arXiv preprint arXiv:2401.08867}, 2024.

\bibitem{ahamed2024timemachine}
Md~Atik Ahamed and Qiang Cheng.
\newblock Timemachine: A time series is worth 4 mambas for long-term forecasting.
\newblock {\em arXiv preprint arXiv:2403.09898}, 2024.

\bibitem{anthony2024blackmamba}
Quentin Anthony, Yury Tokpanov, Paolo Glorioso, and Beren Millidge.
\newblock Blackmamba: Mixture of experts for state-space models.
\newblock {\em arXiv preprint arXiv:2402.01771}, 2024.

\bibitem{arjovsky2016unitary}
Martin Arjovsky, Amar Shah, and Yoshua Bengio.
\newblock Unitary evolution recurrent neural networks.
\newblock In {\em International conference on machine learning}, pages 1120--1128. PMLR, 2016.

\bibitem{arnab2021vivit}
Anurag Arnab, Mostafa Dehghani, Georg Heigold, Chen Sun, Mario Lu{\v{c}}i{\'c}, and Cordelia Schmid.
\newblock Vivit: A video vision transformer.
\newblock In {\em Proceedings of the IEEE/CVF international conference on computer vision}, pages 6836--6846, 2021.

\bibitem{bai2023qwen}
Jinze Bai, Shuai Bai, Shusheng Yang, Shijie Wang, Sinan Tan, Peng Wang, Junyang Lin, Chang Zhou, and Jingren Zhou.
\newblock Qwen-vl: A versatile vision-language model for understanding, localization, text reading, and beyond.
\newblock 2023.

\bibitem{ballarotto2023novo}
Marco Ballarotto, Sabine Willems, Tanja Stiller, Felix Nawa, Julian~A Marschner, Francesca Grisoni, and Daniel Merk.
\newblock De novo design of nurr1 agonists via fragment-augmented generative deep learning in low-data regime.
\newblock {\em Journal of Medicinal Chemistry}, 66(12):8170--8177, 2023.

\bibitem{behrouz2024graph}
Ali Behrouz and Farnoosh Hashemi.
\newblock Graph mamba: Towards learning on graphs with state space models.
\newblock {\em arXiv preprint arXiv:2402.08678}, 2024.

\bibitem{behrouz2024mambamixer}
Ali Behrouz, Michele Santacatterina, and Ramin Zabih.
\newblock Mambamixer: Efficient selective state space models with dual token and channel selection.
\newblock {\em arXiv preprint arXiv:2403.19888}, 2024.

\bibitem{beltagy2020longformer}
Iz~Beltagy, Matthew~E Peters, and Arman Cohan.
\newblock Longformer: The long-document transformer.
\newblock {\em arXiv preprint arXiv:2004.05150}, 2020.

\bibitem{bertasius2021space}
Gedas Bertasius, Heng Wang, and Lorenzo Torresani.
\newblock Is space-time attention all you need for video understanding?
\newblock In {\em ICML}, volume~2, page~4, 2021.

\bibitem{bohacek1996art}
Regine~S Bohacek, Colin McMartin, and Wayne~C Guida.
\newblock The art and practice of structure-based drug design: a molecular modeling perspective.
\newblock {\em Medicinal research reviews}, 16(1):3--50, 1996.

\bibitem{chen2024res}
Chi-Sheng Chen, Guan-Ying Chen, Dong Zhou, Di~Jiang, and Dai-Shi Chen.
\newblock Res-vmamba: Fine-grained food category visual classification using selective state space models with deep residual learning.
\newblock {\em arXiv preprint arXiv:2402.15761}, 2024.

\bibitem{chen2023shikra}
Keqin Chen, Zhao Zhang, Weili Zeng, Richong Zhang, Feng Zhu, and Rui Zhao.
\newblock Shikra: Unleashing multimodal llm's referential dialogue magic.
\newblock {\em arXiv preprint arXiv:2306.15195}, 2023.

\bibitem{chen2024rsmamba}
Keyan Chen, Bowen Chen, Chenyang Liu, Wenyuan Li, Zhengxia Zou, and Zhenwei Shi.
\newblock Rsmamba: Remote sensing image classification with state space model.
\newblock {\em arXiv preprint arXiv:2403.19654}, 2024.

\bibitem{chen2021decision}
Lili Chen, Kevin Lu, Aravind Rajeswaran, Kimin Lee, Aditya Grover, Misha Laskin, Pieter Abbeel, Aravind Srinivas, and Igor Mordatch.
\newblock Decision transformer: Reinforcement learning via sequence modeling.
\newblock {\em Advances in neural information processing systems}, 34:15084--15097, 2021.

\bibitem{choromanski2020rethinking}
Krzysztof~Marcin Choromanski, Valerii Likhosherstov, David Dohan, Xingyou Song, Andreea Gane, Tamas Sarlos, Peter Hawkins, Jared~Quincy Davis, Afroz Mohiuddin, Lukasz Kaiser, et~al.
\newblock Rethinking attention with performers.
\newblock In {\em International Conference on Learning Representations}, 2020.

\bibitem{chu2023mobilevlm}
Xiangxiang Chu, Limeng Qiao, Xinyang Lin, Shuang Xu, Yang Yang, Yiming Hu, Fei Wei, Xinyu Zhang, Bo~Zhang, Xiaolin Wei, et~al.
\newblock Mobilevlm: A fast, reproducible and strong vision language assistant for mobile devices.
\newblock {\em arXiv preprint arXiv:2312.16886}, 2023.

\bibitem{dai2024instructblip}
Wenliang Dai, Junnan Li, Dongxu Li, Anthony Meng~Huat Tiong, Junqi Zhao, Weisheng Wang, Boyang Li, Pascale~N Fung, and Steven Hoi.
\newblock Instructblip: Towards general-purpose vision-language models with instruction tuning.
\newblock {\em Advances in Neural Information Processing Systems}, 36, 2024.

\bibitem{dalla2023nucleotide}
Hugo Dalla-Torre, Liam Gonzalez, Javier Mendoza-Revilla, Nicolas~Lopez Carranza, Adam~Henryk Grzywaczewski, Francesco Oteri, Christian Dallago, Evan Trop, Bernardo~P de~Almeida, Hassan Sirelkhatim, et~al.
\newblock The nucleotide transformer: Building and evaluating robust foundation models for human genomics.
\newblock {\em bioRxiv}, pages 2023--01, 2023.

\bibitem{de2024griffin}
Soham De, Samuel~L Smith, Anushan Fernando, Aleksandar Botev, George Cristian-Muraru, Albert Gu, Ruba Haroun, Leonard Berrada, Yutian Chen, Srivatsan Srinivasan, et~al.
\newblock Griffin: Mixing gated linear recurrences with local attention for efficient language models.
\newblock {\em arXiv preprint arXiv:2402.19427}, 2024.

\bibitem{deng2009imagenet}
Jia Deng, Wei Dong, Richard Socher, Li-Jia Li, Kai Li, and Li~Fei-Fei.
\newblock Imagenet: A large-scale hierarchical image database.
\newblock In {\em 2009 IEEE conference on computer vision and pattern recognition}, pages 248--255. Ieee, 2009.

\bibitem{van2016wavenet}
Sander Dieleman, Heiga Zen, Karen Simonyan, Oriol Vinyals, Alex Graves, Nal Kalchbrenner, Andrew Senior, Koray Kavukcuoglu, et~al.
\newblock Wavenet: A generative model for raw audio.
\newblock {\em arXiv preprint arXiv:1609.03499}, 12, 2016.

\bibitem{feichtenhofer2020x3d}
Christoph Feichtenhofer.
\newblock X3d: Expanding architectures for efficient video recognition.
\newblock In {\em Proceedings of the IEEE/CVF conference on computer vision and pattern recognition}, pages 203--213, 2020.

\bibitem{feichtenhofer2019slowfast}
Christoph Feichtenhofer, Haoqi Fan, Jitendra Malik, and Kaiming He.
\newblock Slowfast networks for video recognition.
\newblock In {\em Proceedings of the IEEE/CVF international conference on computer vision}, pages 6202--6211, 2019.

\bibitem{feichtenhofer2022masked}
Christoph Feichtenhofer, Yanghao Li, Kaiming He, et~al.
\newblock Masked autoencoders as spatiotemporal learners.
\newblock {\em Advances in neural information processing systems}, 35:35946--35958, 2022.

\bibitem{fu2024monarch}
Dan Fu, Simran Arora, Jessica Grogan, Isys Johnson, Evan~Sabri Eyuboglu, Armin Thomas, Benjamin Spector, Michael Poli, Atri Rudra, and Christopher R{\'e}.
\newblock Monarch mixer: A simple sub-quadratic gemm-based architecture.
\newblock {\em Advances in Neural Information Processing Systems}, 36, 2024.

\bibitem{fu2022hungry}
Daniel~Y Fu, Tri Dao, Khaled~K Saab, Armin~W Thomas, Atri Rudra, and Christopher R{\'e}.
\newblock Hungry hungry hippos: Towards language modeling with state space models.
\newblock {\em arXiv preprint arXiv:2212.14052}, 2022.

\bibitem{fu2023simple}
Daniel~Y Fu, Elliot~L Epstein, Eric Nguyen, Armin~W Thomas, Michael Zhang, Tri Dao, Atri Rudra, and Christopher Re.
\newblock Simple hardware-efficient long conideoolutions for sequence modeling.
\newblock In {\em ICLR 2023 Workshop on Mathematical and Empirical Understanding of Foundation Models}, 2023.

\bibitem{fu2023flashfftconv}
Daniel~Y Fu, Hermann Kumbong, Eric Nguyen, and Christopher R{\'e}.
\newblock Flashfftconv: Efficient convolutions for long sequences with tensor cores.
\newblock {\em arXiv preprint arXiv:2311.05908}, 2023.

\bibitem{fu2020d4rl}
Justin Fu, Aviral Kumar, Ofir Nachum, George Tucker, and Sergey Levine.
\newblock D4rl: Datasets for deep data-driven reinforcement learning.
\newblock {\em arXiv preprint arXiv:2004.07219}, 2020.

\bibitem{garg2022can}
Shivam Garg, Dimitris Tsipras, Percy~S Liang, and Gregory Valiant.
\newblock What can transformers learn in-context? a case study of simple function classes.
\newblock {\em Advances in Neural Information Processing Systems}, 35:30583--30598, 2022.

\bibitem{goel2022s}
Karan Goel, Albert Gu, Chris Donahue, and Christopher R{\'e}.
\newblock It’s raw! audio generation with state-space models.
\newblock In {\em International Conference on Machine Learning}, pages 7616--7633. PMLR, 2022.

\bibitem{gong2024nnmamba}
Haifan Gong, Luoyao Kang, Yitao Wang, Xiang Wan, and Haofeng Li.
\newblock nnmamba: 3d biomedical image segmentation, classification and landmark detection with state space model.
\newblock {\em arXiv preprint arXiv:2402.03526}, 2024.

\bibitem{goyal2017something}
Raghav Goyal, Samira Ebrahimi~Kahou, Vincent Michalski, Joanna Materzynska, Susanne Westphal, Heuna Kim, Valentin Haenel, Ingo Fruend, Peter Yianilos, Moritz Mueller-Freitag, et~al.
\newblock The" something something" video database for learning and evaluating visual common sense.
\newblock In {\em Proceedings of the IEEE international conference on computer vision}, pages 5842--5850, 2017.

\bibitem{goyal2017making}
Yash Goyal, Tejas Khot, Douglas Summers-Stay, Dhruv Batra, and Devi Parikh.
\newblock Making the v in vqa matter: Elevating the role of image understanding in visual question answering.
\newblock In {\em Proceedings of the IEEE conference on computer vision and pattern recognition}, pages 6904--6913, 2017.

\bibitem{grazzi2024mamba}
Riccardo Grazzi, Julien Siems, Simon Schrodi, Thomas Brox, and Frank Hutter.
\newblock Is mamba capable of in-context learning?
\newblock {\em arXiv preprint arXiv:2402.03170}, 2024.

\bibitem{grisoni2023chemical}
Francesca Grisoni.
\newblock Chemical language models for de novo drug design: Challenges and opportunities.
\newblock {\em Current Opinion in Structural Biology}, 79:102527, 2023.

\bibitem{grisoni2021combining}
Francesca Grisoni, Berend~JH Huisman, Alexander~L Button, Michael Moret, Kenneth Atz, Daniel Merk, and Gisbert Schneider.
\newblock Combining generative artificial intelligence and on-chip synthesis for de novo drug design.
\newblock {\em Science Advances}, 7(24):eabg3338, 2021.

\bibitem{gu2023mamba}
Albert Gu and Tri Dao.
\newblock Mamba: Linear-time sequence modeling with selective state spaces.
\newblock {\em arXiv preprint arXiv:2312.00752}, 2023.

\bibitem{albert2020hippo}
Albert Gu, Tri Dao, Stefano Ermon, Atri Rudra, and Christopher R\'{e}.
\newblock Hippo: Recurrent memory with optimal polynomial projections.
\newblock In H.~Larochelle, M.~Ranzato, R.~Hadsell, M.F. Balcan, and H.~Lin, editors, {\em Advances in Neural Information Processing Systems}, volume~33, pages 1474--1487. Curran Associates, Inc., 2020.

\bibitem{gu2022parameterization}
Albert Gu, Karan Goel, Ankit Gupta, and Christopher R{\'e}.
\newblock On the parameterization and initialization of diagonal state space models.
\newblock {\em Advances in Neural Information Processing Systems}, 35:35971--35983, 2022.

\bibitem{gu2021efficiently}
Albert Gu, Karan Goel, and Christopher Re.
\newblock Efficiently modeling long sequences with structured state spaces.
\newblock In {\em International Conference on Learning Representations}, 2021.

\bibitem{49414}
Anmol Gulati, Chung-Cheng Chiu, James Qin, Jiahui Yu, Niki Parmar, Ruoming Pang, Shibo Wang, Wei Han, Yonghui Wu, Yu~Zhang, and Zhengdong Zhang, editors.
\newblock {\em Conformer: Convolution-augmented Transformer for Speech Recognition}, 2020.

\bibitem{guo2024mambair}
Hang Guo, Jinmin Li, Tao Dai, Zhihao Ouyang, Xudong Ren, and Shu-Tao Xia.
\newblock Mambair: A simple baseline for image restoration with state-space model.
\newblock {\em arXiv preprint arXiv:2402.15648}, 2024.

\bibitem{guo2022cmt}
Jianyuan Guo, Kai Han, Han Wu, Yehui Tang, Xinghao Chen, Yunhe Wang, and Chang Xu.
\newblock Cmt: Convolutional neural networks meet vision transformers.
\newblock In {\em Proceedings of the IEEE/CVF Conference on Computer Vision and Pattern Recognition}, pages 12175--12185, 2022.

\bibitem{guo2024mambamorph}
Tao Guo, Yinuo Wang, and Cai Meng.
\newblock Mambamorph: a mamba-based backbone with contrastive feature learning for deformable mr-ct registration.
\newblock {\em arXiv preprint arXiv:2401.13934}, 2024.

\bibitem{gupta2022diagonal}
Ankit Gupta, Albert Gu, and Jonathan Berant.
\newblock Diagonal state spaces are as effective as structured state spaces.
\newblock {\em Advances in Neural Information Processing Systems}, 35:22982--22994, 2022.

\bibitem{han2022turbo}
Tengda Han, Weidi Xie, and Andrew Zisserman.
\newblock Turbo training with token dropout.
\newblock {\em arXiv preprint arXiv:2210.04889}, 2022.

\bibitem{hao2024t}
Jing Hao, Lei He, and Kuo~Feng Hung.
\newblock T-mamba: Frequency-enhanced gated long-range dependency for tooth 3d cbct segmentation.
\newblock {\em arXiv preprint arXiv:2404.01065}, 2024.

\bibitem{hasani2021liquid}
Ramin Hasani, Mathias Lechner, Alexander Amini, Daniela Rus, and Radu Grosu.
\newblock Liquid time-constant networks.
\newblock In {\em Proceedings of the AAAI Conference on Artificial Intelligence}, volume~35, pages 7657--7666, 2021.

\bibitem{hasani2022liquid}
Ramin Hasani, Mathias Lechner, Tsun-Hsuan Wang, Makram Chahine, Alexander Amini, and Daniela Rus.
\newblock Liquid structural state-space models.
\newblock In {\em The Eleventh International Conference on Learning Representations}, 2022.

\bibitem{he2024densemamba}
Wei He, Kai Han, Yehui Tang, Chengcheng Wang, Yujie Yang, Tianyu Guo, and Yunhe Wang.
\newblock Densemamba: State space models with dense hidden connection for efficient large language models.
\newblock {\em arXiv preprint arXiv:2403.00818}, 2024.

\bibitem{hershey2016deep}
John~R Hershey, Zhuo Chen, Jonathan Le~Roux, and Shinji Watanabe.
\newblock Deep clustering: Discriminative embeddings for segmentation and separation.
\newblock In {\em 2016 IEEE international conference on acoustics, speech and signal processing (ICASSP)}, pages 31--35. IEEE, 2016.

\bibitem{hua2022transformer}
Weizhe Hua, Zihang Dai, Hanxiao Liu, and Quoc Le.
\newblock Transformer quality in linear time.
\newblock In {\em International conference on machine learning}, pages 9099--9117. PMLR, 2022.

\bibitem{huang2024localmamba}
Tao Huang, Xiaohuan Pei, Shan You, Fei Wang, Chen Qian, and Chang Xu.
\newblock Localmamba: Visual state space model with windowed selective scan.
\newblock {\em arXiv preprint arXiv:2403.09338}, 2024.

\bibitem{Hudson_2019_CVPR}
Drew~A. Hudson and Christopher~D. Manning.
\newblock Gqa: A new dataset for real-world visual reasoning and compositional question answering.
\newblock In {\em Proceedings of the IEEE/CVF Conference on Computer Vision and Pattern Recognition (CVPR)}, June 2019.

\bibitem{hussein2019timeception}
Noureldien Hussein, Efstratios Gavves, and Arnold~WM Smeulders.
\newblock Timeception for complex action recognition.
\newblock In {\em Proceedings of the IEEE/CVF Conference on Computer Vision and Pattern Recognition}, pages 254--263, 2019.

\bibitem{hussein2019videograph}
Noureldien Hussein, Efstratios Gavves, and Arnold~WM Smeulders.
\newblock Videograph: Recognizing minutes-long human activities in videos.
\newblock {\em arXiv preprint arXiv:1905.05143}, 2019.

\bibitem{hutchins2022block}
DeLesley Hutchins, Imanol Schlag, Yuhuai Wu, Ethan Dyer, and Behnam Neyshabur.
\newblock Block-recurrent transformers.
\newblock {\em Advances in neural information processing systems}, 35:33248--33261, 2022.

\bibitem{islam2022long}
Md~Mohaiminul Islam and Gedas Bertasius.
\newblock Long movie clip classification with state-space video models.
\newblock In {\em European Conference on Computer Vision}, pages 87--104. Springer, 2022.

\bibitem{jelassi2024repeat}
Samy Jelassi, David Brandfonbrener, Sham~M Kakade, and Eran Malach.
\newblock Repeat after me: Transformers are better than state space models at copying.
\newblock {\em arXiv preprint arXiv:2402.01032}, 2024.

\bibitem{jiang2024dual}
Xilin Jiang, Cong Han, and Nima Mesgarani.
\newblock Dual-path mamba: Short and long-term bidirectional selective structured state space models for speech separation.
\newblock {\em arXiv preprint arXiv:2403.18257}, 2024.

\bibitem{katharopoulos2020transformers}
Angelos Katharopoulos, Apoorv Vyas, Nikolaos Pappas, and Fran{\c{c}}ois Fleuret.
\newblock Transformers are rnns: Fast autoregressive transformers with linear attention.
\newblock In {\em International Conference on Machine Learning}, pages 5156--5165. PMLR, 2020.

\bibitem{kaul2020linear}
Shiva Kaul.
\newblock Linear dynamical systems as a core computational primitive.
\newblock {\em Advances in Neural Information Processing Systems}, 33:16808--16820, 2020.

\bibitem{kay2017kinetics}
Will Kay, Joao Carreira, Karen Simonyan, Brian Zhang, Chloe Hillier, Sudheendra Vijayanarasimhan, Fabio Viola, Tim Green, Trevor Back, Paul Natsev, et~al.
\newblock The kinetics human action video dataset.
\newblock {\em arXiv preprint arXiv:1705.06950}, 2017.

\bibitem{kenton2019bert}
Jacob Devlin Ming-Wei~Chang Kenton and Lee~Kristina Toutanova.
\newblock Bert: Pre-training of deep bidirectional transformers for language understanding.
\newblock In {\em Proceedings of NAACL-HLT}, pages 4171--4186, 2019.

\bibitem{kitaev2019reformer}
Nikita Kitaev, Lukasz Kaiser, and Anselm Levskaya.
\newblock Reformer: The efficient transformer.
\newblock In {\em International Conference on Learning Representations}, 2019.

\bibitem{kuehne2014language}
Hilde Kuehne, Ali Arslan, and Thomas Serre.
\newblock The language of actions: Recovering the syntax and semantics of goal-directed human activities.
\newblock In {\em Proceedings of the IEEE conference on computer vision and pattern recognition}, pages 780--787, 2014.

\bibitem{kuehne2011hmdb}
Hildegard Kuehne, Hueihan Jhuang, Est{\'\i}baliz Garrote, Tomaso Poggio, and Thomas Serre.
\newblock Hmdb: a large video database for human motion recognition.
\newblock In {\em 2011 International conference on computer vision}, pages 2556--2563. IEEE, 2011.

\bibitem{laurenccon2024obelics}
Hugo Lauren{\c{c}}on, Lucile Saulnier, L{\'e}o Tronchon, Stas Bekman, Amanpreet Singh, Anton Lozhkov, Thomas Wang, Siddharth Karamcheti, Alexander Rush, Douwe Kiela, et~al.
\newblock Obelics: An open web-scale filtered dataset of interleaved image-text documents.
\newblock {\em Advances in Neural Information Processing Systems}, 36, 2024.

\bibitem{lee2021fnet}
James Lee-Thorp, Joshua Ainslie, Ilya Eckstein, and Santiago Ontanon.
\newblock Fnet: Mixing tokens with fourier transforms.
\newblock {\em arXiv preprint arXiv:2105.03824}, 2021.

\bibitem{li2023mimic}
Bo~Li, Yuanhan Zhang, Liangyu Chen, Jinghao Wang, Fanyi Pu, Jingkang Yang, Chunyuan Li, and Ziwei Liu.
\newblock Mimic-it: Multi-modal in-context instruction tuning.
\newblock {\em arXiv preprint arXiv:2306.05425}, 2023.

\bibitem{li2023blip}
Junnan Li, Dongxu Li, Silvio Savarese, and Steven Hoi.
\newblock Blip-2: Bootstrapping language-image pre-training with frozen image encoders and large language models.
\newblock In {\em International conference on machine learning}, pages 19730--19742. PMLR, 2023.

\bibitem{li2024spmamba}
Kai Li and Guo Chen.
\newblock Spmamba: State-space model is all you need in speech separation.
\newblock {\em arXiv preprint arXiv:2404.02063}, 2024.

\bibitem{li2020ct}
Kunchang Li, Xianhang Li, Yali Wang, Jun Wang, and Yu~Qiao.
\newblock Ct-net: Channel tensorization network for video classification.
\newblock In {\em International Conference on Learning Representations}, 2020.

\bibitem{li2024videomamba}
Kunchang Li, Xinhao Li, Yi~Wang, Yinan He, Yali Wang, Limin Wang, and Yu~Qiao.
\newblock Videomamba: State space model for efficient video understanding.
\newblock {\em arXiv preprint arXiv:2403.06977}, 2024.

\bibitem{li2023unmasked}
Kunchang Li, Yali Wang, Yizhuo Li, Yi~Wang, Yinan He, Limin Wang, and Yu~Qiao.
\newblock Unmasked teacher: Towards training-efficient video foundation models.
\newblock In {\em Proceedings of the IEEE/CVF International Conference on Computer Vision}, pages 19948--19960, 2023.

\bibitem{li2022uniformer}
Kunchang Li, Yali Wang, Junhao Zhang, Peng Gao, Guanglu Song, Yu~Liu, Hongsheng Li, and Yu~Qiao.
\newblock Uniformer: Unifying convolution and self-attention for visual recognition.
\newblock {\em arXiv preprint arXiv:2201.09450}, 2022.

\bibitem{li2024stg}
Lincan Li, Hanchen Wang, Wenjie Zhang, and Adelle Coster.
\newblock Stg-mamba: Spatial-temporal graph learning via selective state space model.
\newblock {\em arXiv preprint arXiv:2403.12418}, 2024.

\bibitem{li2024mamba}
Shufan Li, Harkanwar Singh, and Aditya Grover.
\newblock Mamba-nd: Selective state space modeling for multi-dimensional data.
\newblock {\em arXiv preprint arXiv:2402.05892}, 2024.

\bibitem{li2024spikemba}
Wenrui Li, Xiaopeng Hong, and Xiaopeng Fan.
\newblock Spikemba: Multi-modal spiking saliency mamba for temporal video grounding.
\newblock {\em arXiv preprint arXiv:2404.01174}, 2024.

\bibitem{li2022mvitv2}
Yanghao Li, Chao-Yuan Wu, Haoqi Fan, Karttikeya Mangalam, Bo~Xiong, Jitendra Malik, and Christoph Feichtenhofer.
\newblock Mvitv2: Improved multiscale vision transformers for classification and detection.
\newblock In {\em Proceedings of the IEEE/CVF Conference on Computer Vision and Pattern Recognition}, pages 4804--4814, 2022.

\bibitem{li2022efficientformer}
Yanyu Li, Geng Yuan, Yang Wen, Eric Hu, Georgios Evangelidis, Sergey Tulyakov, Yanzhi Wang, and Jian Ren.
\newblock Efficientformer: Vision transformers at mobilenet speed.
\newblock {\em arXiv preprint arXiv:2206.01191}, 2022.

\bibitem{li2023evaluating}
Yifan Li, Yifan Du, Kun Zhou, Jinpeng Wang, Wayne~Xin Zhao, and Ji-Rong Wen.
\newblock Evaluating object hallucination in large vision-language models.
\newblock In {\em Proceedings of the 2023 Conference on Empirical Methods in Natural Language Processing}, pages 292--305, 2023.

\bibitem{Li2022WhatMC}
Yuhong Li, Tianle Cai, Yi~Zhang, De~huai Chen, and Debadeepta Dey.
\newblock What makes convolutional models great on long sequence modeling?
\newblock {\em ArXiv}, abs/2210.09298, 2022.

\bibitem{liang2023rethinking}
Chaoqi Liang, Weiqiang Bai, Lifeng Qiao, Yuchen Ren, Jianle Sun, Peng Ye, Hongliang Yan, Xinzhu Ma, Wangmeng Zuo, and Wanli Ouyang.
\newblock Rethinking the bert-like pretraining for dna sequences.
\newblock {\em arXiv preprint arXiv:2310.07644}, 2023.

\bibitem{liao2024lightm}
Weibin Liao, Yinghao Zhu, Xinyuan Wang, Cehngwei Pan, Yasha Wang, and Liantao Ma.
\newblock Lightm-unet: Mamba assists in lightweight unet for medical image segmentation.
\newblock {\em arXiv preprint arXiv:2403.05246}, 2024.

\bibitem{lieber2024jamba}
Opher Lieber, Barak Lenz, Hofit Bata, Gal Cohen, Jhonathan Osin, Itay Dalmedigos, Erez Safahi, Shaked Meirom, Yonatan Belinkov, Shai Shalev-Shwartz, et~al.
\newblock Jamba: A hybrid transformer-mamba language model.
\newblock {\em arXiv preprint arXiv:2403.19887}, 2024.

\bibitem{lim2021temporal}
Bryan Lim, Sercan~{\"O} Ar{\i}k, Nicolas Loeff, and Tomas Pfister.
\newblock Temporal fusion transformers for interpretable multi-horizon time series forecasting.
\newblock {\em International Journal of Forecasting}, 37(4):1748--1764, 2021.

\bibitem{lin2022learning}
Xudong Lin, Fabio Petroni, Gedas Bertasius, Marcus Rohrbach, Shih-Fu Chang, and Lorenzo Torresani.
\newblock Learning to recognize procedural activities with distant supervision.
\newblock In {\em Proceedings of the IEEE/CVF Conference on Computer Vision and Pattern Recognition}, pages 13853--13863, 2022.

\bibitem{liu2021pay}
Hanxiao Liu, Zihang Dai, David So, and Quoc~V Le.
\newblock Pay attention to mlps.
\newblock {\em Advances in Neural Information Processing Systems}, 34:9204--9215, 2021.

\bibitem{liu2024visual}
Haotian Liu, Chunyuan Li, Qingyang Wu, and Yong~Jae Lee.
\newblock Visual instruction tuning.
\newblock {\em Advances in neural information processing systems}, 36, 2024.

\bibitem{liu2024swin}
Jiarun Liu, Hao Yang, Hong-Yu Zhou, Yan Xi, Lequan Yu, Yizhou Yu, Yong Liang, Guangming Shi, Shaoting Zhang, Hairong Zheng, et~al.
\newblock Swin-umamba: Mamba-based unet with imagenet-based pretraining.
\newblock {\em arXiv preprint arXiv:2402.03302}, 2024.

\bibitem{liu2023mmbench}
Yuan Liu, Haodong Duan, Yuanhan Zhang, Bo~Li, Songyang Zhang, Wangbo Zhao, Yike Yuan, Jiaqi Wang, Conghui He, Ziwei Liu, et~al.
\newblock Mmbench: Is your multi-modal model an all-around player?
\newblock {\em arXiv preprint arXiv:2307.06281}, 2023.

\bibitem{liu2024vmamba}
Yue Liu, Yunjie Tian, Yuzhong Zhao, Hongtian Yu, Lingxi Xie, Yaowei Wang, Qixiang Ye, and Yunfan Liu.
\newblock Vmamba: Visual state space model.
\newblock {\em arXiv preprint arXiv:2401.10166}, 2024.

\bibitem{liu2022swin}
Ze~Liu, Han Hu, Yutong Lin, Zhuliang Yao, Zhenda Xie, Yixuan Wei, Jia Ning, Yue Cao, Zheng Zhang, Li~Dong, et~al.
\newblock Swin transformer v2: Scaling up capacity and resolution.
\newblock In {\em Proceedings of the IEEE/CVF Conference on Computer Vision and Pattern Recognition}, pages 12009--12019, 2022.

\bibitem{NEURIPS2022_11332b6b}
Pan Lu, Swaroop Mishra, Tanglin Xia, Liang Qiu, Kai-Wei Chang, Song-Chun Zhu, Oyvind Tafjord, Peter Clark, and Ashwin Kalyan.
\newblock Learn to explain: Multimodal reasoning via thought chains for science question answering.
\newblock In S.~Koyejo, S.~Mohamed, A.~Agarwal, D.~Belgrave, K.~Cho, and A.~Oh, editors, {\em Advances in Neural Information Processing Systems}, volume~35, pages 2507--2521. Curran Associates, Inc., 2022.

\bibitem{ma2024u}
Jun Ma, Feifei Li, and Bo~Wang.
\newblock U-mamba: Enhancing long-range dependency for biomedical image segmentation.
\newblock {\em arXiv preprint arXiv:2401.04722}, 2024.

\bibitem{ma2021luna}
Xuezhe Ma, Xiang Kong, Sinong Wang, Chunting Zhou, Jonathan May, Hao Ma, and Luke Zettlemoyer.
\newblock Luna: Linear unified nested attention.
\newblock {\em Advances in Neural Information Processing Systems}, 34:2441--2453, 2021.

\bibitem{ma2022mega}
Xuezhe Ma, Chunting Zhou, Xiang Kong, Junxian He, Liangke Gui, Graham Neubig, Jonathan May, and Luke Zettlemoyer.
\newblock Mega: Moving average equipped gated attention.
\newblock In {\em The Eleventh International Conference on Learning Representations}, 2022.

\bibitem{mehta2022long}
Harsh Mehta, Ankit Gupta, Ashok Cutkosky, and Behnam Neyshabur.
\newblock Long range language modeling via gated state spaces.
\newblock In {\em The Eleventh International Conference on Learning Representations}, 2022.

\bibitem{merity2016pointer}
Stephen Merity, Caiming Xiong, James Bradbury, and Richard Socher.
\newblock Pointer sentinel mixture models.
\newblock In {\em International Conference on Learning Representations}, 2016.

\bibitem{merk2018novo}
Daniel Merk, Lukas Friedrich, Francesca Grisoni, and Gisbert Schneider.
\newblock De novo design of bioactive small molecules by artificial intelligence.
\newblock {\em Molecular informatics}, 37(1-2):1700153, 2018.

\bibitem{nguyen2022s4nd}
Eric Nguyen, Karan Goel, Albert Gu, Gordon Downs, Preey Shah, Tri Dao, Stephen Baccus, and Christopher R{\'e}.
\newblock S4nd: Modeling images and videos as multidimensional signals with state spaces.
\newblock {\em Advances in neural information processing systems}, 35:2846--2861, 2022.

\bibitem{nguyen2023hyenadna}
Eric Nguyen, Michael Poli, Marjan Faizi, Armin~W Thomas, Michael Wornow, Callum Birch-Sykes, Stefano Massaroli, Aman Patel, Clayton~M Rabideau, Yoshua Bengio, et~al.
\newblock Hyenadna: Long-range genomic sequence modeling at single nucleotide resolution.
\newblock In {\em Thirty-seventh Conference on Neural Information Processing Systems}, 2023.

\bibitem{nie2022time}
Yuqi Nie, Nam~H Nguyen, Phanwadee Sinthong, and Jayant Kalagnanam.
\newblock A time series is worth 64 words: Long-term forecasting with transformers.
\newblock In {\em The Eleventh International Conference on Learning Representations}, 2022.

\bibitem{olsson2022context}
Catherine Olsson, Nelson Elhage, Neel Nanda, Nicholas Joseph, Nova DasSarma, Tom Henighan, Ben Mann, Amanda Askell, Yuntao Bai, Anna Chen, Tom Conerly, Dawn Drain, Deep Ganguli, Zac Hatfield-Dodds, Danny Hernandez, Scott Johnston, Andy Jones, Jackson Kernion, Liane Lovitt, Kamal Ndousse, Dario Amodei, Tom Brown, Jack Clark, Jared Kaplan, Sam McCandlish, and Chris Olah.
\newblock In-context learning and induction heads.
\newblock {\em Transformer Circuits Thread}, 2022.
\newblock https://transformer-circuits.pub/2022/in-context-learning-and-induction-heads/index.html.

\bibitem{orvieto2023resurrecting}
Antonio Orvieto, Samuel~L Smith, Albert Gu, Anushan Fernando, Caglar Gulcehre, Razvan Pascanu, and Soham De.
\newblock Resurrecting recurrent neural networks for long sequences.
\newblock In {\em International Conference on Machine Learning}, pages 26670--26698. PMLR, 2023.

\bibitem{ota2024decision}
Toshihiro Ota.
\newblock Decision mamba: Reinforcement learning via sequence modeling with selective state spaces.
\newblock {\em arXiv preprint arXiv:2403.19925}, 2024.

\bibitem{ozccelik2024chemical}
R{\i}za {\"O}z{\c{c}}elik, Sarah de~Ruiter, Emanuele Criscuolo, and Francesca Grisoni.
\newblock Chemical language modeling with structured state spaces.
\newblock 2024.

\bibitem{LibriSpeech}
Vassil Panayotov, Guoguo Chen, Daniel Povey, and Sanjeev Khudanpur.
\newblock Librispeech: An asr corpus based on public domain audio books.
\newblock In {\em 2015 IEEE International Conference on Acoustics, Speech and Signal Processing (ICASSP)}, pages 5206--5210, 2015.

\bibitem{park2024can}
Jongho Park, Jaeseung Park, Zheyang Xiong, Nayoung Lee, Jaewoong Cho, Samet Oymak, Kangwook Lee, and Dimitris Papailiopoulos.
\newblock Can mamba learn how to learn? a comparative study on in-context learning tasks.
\newblock {\em arXiv preprint arXiv:2402.04248}, 2024.

\bibitem{patrick2021keeping}
Mandela Patrick, Dylan Campbell, Yuki Asano, Ishan Misra, Florian Metze, Christoph Feichtenhofer, Andrea Vedaldi, and Joao~F Henriques.
\newblock Keeping your eye on the ball: Trajectory attention in video transformers.
\newblock {\em Advances in neural information processing systems}, 34:12493--12506, 2021.

\bibitem{patro2023efficiency}
Badri~N Patro and Vijay Agneeswaran.
\newblock Efficiency 360: Efficient vision transformers.
\newblock {\em arXiv preprint arXiv:2302.08374}, 2023.

\bibitem{patro2024simba}
Badri~N Patro and Vijay~S Agneeswaran.
\newblock Simba: Simplified mamba-based architecture for vision and multivariate time series.
\newblock {\em arXiv preprint arXiv:2403.15360}, 2024.

\bibitem{patro2024spectral}
Badri~N Patro, Vinay~P Namboodiri, and Vijay~S Agneeswaran.
\newblock Spectral convolutional transformer: Harmonizing real vs. complex multi-view spectral operators for vision transformer.
\newblock {\em arXiv preprint arXiv:2403.18063}, 2024.

\bibitem{patro2023spectformer}
Badri~N Patro, Vinay~P Namboodiri, and Vijay~Srinivas Agneeswaran.
\newblock Spectformer: Frequency and attention is what you need in a vision transformer.
\newblock {\em arXiv preprint arXiv:2304.06446}, 2023.

\bibitem{patro2023scattering}
Badri~Narayana Patro and Vijay~Srinivas Agneeswaran.
\newblock Scattering vision transformer: Spectral mixing matters.
\newblock In {\em Thirty-seventh Conference on Neural Information Processing Systems}, 2023.

\bibitem{peng2024limitations}
Binghui Peng, Srini Narayanan, and Christos Papadimitriou.
\newblock On limitations of the transformer architecture.
\newblock {\em arXiv preprint arXiv:2402.08164}, 2024.

\bibitem{peng2023rwkv}
Bo~Peng, Eric Alcaide, Quentin Anthony, Alon Albalak, Samuel Arcadinho, Stella Biderman, Huanqi Cao, Xin Cheng, Michael Chung, Leon Derczynski, et~al.
\newblock Rwkv: Reinventing rnns for the transformer era.
\newblock In {\em Findings of the Association for Computational Linguistics: EMNLP 2023}, pages 14048--14077, 2023.

\bibitem{BranchFormer}
Yifan Peng, Siddharth Dalmia, Ian Lane, and Shinji Watanabe.
\newblock Branchformer: Parallel mlp-attention architectures to capture local and global context for speech recognition and understanding.
\newblock {\em Proceedings of Machine Learning Research}, 162:17627--17643, 2022.
\newblock Publisher Copyright: Copyright {\textcopyright} 2022 by the author(s); 39th International Conference on Machine Learning, ICML 2022 ; Conference date: 17-07-2022 Through 23-07-2022.

\bibitem{pioro2024moe}
Maciej Pi{\'o}ro, Kamil Ciebiera, Krystian Kr{\'o}l, Jan Ludziejewski, and Sebastian Jaszczur.
\newblock Moe-mamba: Efficient selective state space models with mixture of experts.
\newblock {\em arXiv preprint arXiv:2401.04081}, 2024.

\bibitem{platonov2022critical}
Oleg Platonov, Denis Kuznedelev, Michael Diskin, Artem Babenko, and Liudmila Prokhorenkova.
\newblock A critical look at the evaluation of gnns under heterophily: Are we really making progress?
\newblock In {\em The Eleventh International Conference on Learning Representations}, 2022.

\bibitem{poli2023hyena}
Michael Poli, Stefano Massaroli, Eric Nguyen, Daniel~Y Fu, Tri Dao, Stephen Baccus, Yoshua Bengio, Stefano Ermon, and Christopher Re.
\newblock Hyena hierarchy: Towards larger convolutional language models.
\newblock 2023.

\bibitem{qiao2024vl}
Yanyuan Qiao, Zheng Yu, Longteng Guo, Sihan Chen, Zijia Zhao, Mingzhen Sun, Qi~Wu, and Jing Liu.
\newblock Vl-mamba: Exploring state space models for multimodal learning.
\newblock {\em arXiv preprint arXiv:2403.13600}, 2024.

\bibitem{qin2022toeplitz}
Zhen Qin, Xiaodong Han, Weixuan Sun, Bowen He, Dong Li, Dongxu Li, Yuchao Dai, Lingpeng Kong, and Yiran Zhong.
\newblock Toeplitz neural network for sequence modeling.
\newblock In {\em The Eleventh International Conference on Learning Representations}, 2022.

\bibitem{qin2021cosformer}
Zhen Qin, Weixuan Sun, Hui Deng, Dongxu Li, Yunshen Wei, Baohong Lv, Junjie Yan, Lingpeng Kong, and Yiran Zhong.
\newblock cosformer: Rethinking softmax in attention.
\newblock In {\em International Conference on Learning Representations}, 2021.

\bibitem{qin2023hierarchically}
Zhen Qin, Songlin Yang, and Yiran Zhong.
\newblock Hierarchically gated recurrent neural network for sequence modeling.
\newblock {\em arXiv preprint arXiv:2311.04823}, 2023.

\bibitem{qin2024hierarchically}
Zhen Qin, Songlin Yang, and Yiran Zhong.
\newblock Hierarchically gated recurrent neural network for sequence modeling.
\newblock {\em Advances in Neural Information Processing Systems}, 36, 2024.

\bibitem{quan2024multichannel}
Changsheng Quan and Xiaofei Li.
\newblock Multichannel long-term streaming neural speech enhancement for static and moving speakers.
\newblock {\em arXiv preprint arXiv:2403.07675}, 2024.

\bibitem{radosavovic2020designing}
Ilija Radosavovic, Raj~Prateek Kosaraju, Ross Girshick, Kaiming He, and Piotr Doll{\'a}r.
\newblock Designing network design spaces.
\newblock In {\em Proceedings of the IEEE/CVF conference on computer vision and pattern recognition}, pages 10428--10436, 2020.

\bibitem{rae2019compressive}
Jack~W Rae, Anna Potapenko, Siddhant~M Jayakumar, and Timothy~P Lillicrap.
\newblock Compressive transformers for long-range sequence modelling.
\newblock {\em arXiv preprint arXiv:1911.05507}, 2019.

\bibitem{rimon2024mamba}
Zohar Rimon, Tom Jurgenson, Orr Krupnik, Gilad Adler, and Aviv Tamar.
\newblock Mamba: an effective world model approach for meta-reinforcement learning.
\newblock {\em arXiv preprint arXiv:2403.09859}, 2024.

\bibitem{romero2022towards}
David~W Romero, David~M Knigge, Albert Gu, Erik~J Bekkers, Efstratios Gavves, Jakub~M Tomczak, and Mark Hoogendoorn.
\newblock Towards a general purpose cnn for long range dependencies in $ n $ d.
\newblock {\em arXiv preprint arXiv:2206.03398}, 2022.

\bibitem{ruan2024vm}
Jiacheng Ruan and Suncheng Xiang.
\newblock Vm-unet: Vision mamba unet for medical image segmentation.
\newblock {\em arXiv preprint arXiv:2402.02491}, 2024.

\bibitem{sanford2024representational}
Clayton Sanford, Daniel~J Hsu, and Matus Telgarsky.
\newblock Representational strengths and limitations of transformers.
\newblock {\em Advances in Neural Information Processing Systems}, 36, 2024.

\bibitem{sharir2021image}
Gilad Sharir, Asaf Noy, and Lihi Zelnik-Manor.
\newblock An image is worth 16x16 words, what is a video worth?
\newblock {\em arXiv preprint arXiv:2103.13915}, 2021.

\bibitem{shazeer2019fast}
Noam Shazeer.
\newblock Fast transformer decoding: One write-head is all you need.
\newblock {\em arXiv preprint arXiv:1911.02150}, 2019.

\bibitem{shen2024gamba}
Qiuhong Shen, Xuanyu Yi, Zike Wu, Pan Zhou, Hanwang Zhang, Shuicheng Yan, and Xinchao Wang.
\newblock Gamba: Marry gaussian splatting with mamba for single view 3d reconstruction.
\newblock {\em arXiv preprint arXiv:2403.18795}, 2024.

\bibitem{si2022inception}
Chenyang Si, Weihao Yu, Pan Zhou, Yichen Zhou, Xinchao Wang, and Shuicheng YAN.
\newblock Inception transformer.
\newblock In {\em Advances in Neural Information Processing Systems}, 2022.

\bibitem{singh2019towards}
Amanpreet Singh, Vivek Natarajan, Meet Shah, Yu~Jiang, Xinlei Chen, Dhruv Batra, Devi Parikh, and Marcus Rohrbach.
\newblock Towards vqa models that can read.
\newblock In {\em Proceedings of the IEEE/CVF conference on computer vision and pattern recognition}, pages 8317--8326, 2019.

\bibitem{smith2022simplified}
Jimmy~TH Smith, Andrew Warrington, and Scott Linderman.
\newblock Simplified state space layers for sequence modeling.
\newblock In {\em The Eleventh International Conference on Learning Representations}, 2022.

\bibitem{sun2019videobert}
Chen Sun, Austin Myers, Carl Vondrick, Kevin Murphy, and Cordelia Schmid.
\newblock Videobert: A joint model for video and language representation learning.
\newblock In {\em Proceedings of the IEEE/CVF International Conference on Computer Vision}, pages 7464--7473, 2019.

\bibitem{tang2019coin}
Yansong Tang, Dajun Ding, Yongming Rao, Yu~Zheng, Danyang Zhang, Lili Zhao, Jiwen Lu, and Jie Zhou.
\newblock Coin: A large-scale dataset for comprehensive instructional video analysis.
\newblock In {\em Proceedings of the IEEE/CVF Conference on Computer Vision and Pattern Recognition}, pages 1207--1216, 2019.

\bibitem{tay2021synthesizer}
Yi~Tay, Dara Bahri, Donald Metzler, Da-Cheng Juan, Zhe Zhao, and Che Zheng.
\newblock Synthesizer: Rethinking self-attention for transformer models.
\newblock In {\em International conference on machine learning}, pages 10183--10192. PMLR, 2021.

\bibitem{tay2020sparse}
Yi~Tay, Dara Bahri, Liu Yang, Donald Metzler, and Da-Cheng Juan.
\newblock Sparse sinkhorn attention.
\newblock In {\em International Conference on Machine Learning}, pages 9438--9447. PMLR, 2020.

\bibitem{tay2020long}
Yi~Tay, Mostafa Dehghani, Samira Abnar, Yikang Shen, Dara Bahri, Philip Pham, Jinfeng Rao, Liu Yang, Sebastian Ruder, and Donald Metzler.
\newblock Long range arena: A benchmark for efficient transformers.
\newblock In {\em International Conference on Learning Representations}, 2020.

\bibitem{thangavel2023limitations}
Thirupathi Thangavel.
\newblock Limitations of transformer architecture.
\newblock In {\em medium.com}, 2023.

\bibitem{tong2022videomae}
Zhan Tong, Yibing Song, Jue Wang, and Limin Wang.
\newblock Videomae: Masked autoencoders are data-efficient learners for self-supervised video pre-training.
\newblock {\em Advances in neural information processing systems}, 35:10078--10093, 2022.

\bibitem{touvron2021training}
Hugo Touvron, Matthieu Cord, Matthijs Douze, Francisco Massa, Alexandre Sablayrolles, and Herv{\'e} J{\'e}gou.
\newblock Training data-efficient image transformers \& distillation through attention.
\newblock In {\em International Conference on Machine Learning}, pages 10347--10357. PMLR, 2021.

\bibitem{tu2022maxvit}
Zhengzhong Tu, Hossein Talebi, Han Zhang, Feng Yang, Peyman Milanfar, Alan Bovik, and Yinxiao Li.
\newblock Maxvit: Multi-axis vision transformer.
\newblock In {\em Computer Vision--ECCV 2022: 17th European Conference, Tel Aviv, Israel, October 23--27, 2022, Proceedings, Part XXIV}, pages 459--479. Springer, 2022.

\bibitem{vaswani2017attention}
Ashish Vaswani, Noam Shazeer, Niki Parmar, Jakob Uszkoreit, Llion Jones, Aidan~N Gomez, {\L}ukasz Kaiser, and Illia Polosukhin.
\newblock Attention is all you need.
\newblock {\em Advances in neural information processing systems}, 30, 2017.

\bibitem{wang2024graph}
Chloe Wang, Oleksii Tsepa, Jun Ma, and Bo~Wang.
\newblock Graph-mamba: Towards long-range graph sequence modeling with selective state spaces.
\newblock {\em arXiv preprint arXiv:2402.00789}, 2024.

\bibitem{wang2024large}
Jinhong Wang, Jintai Chen, Danny Chen, and Jian Wu.
\newblock Large window-based mamba unet for medical image segmentation: Beyond convolution and self-attention.
\newblock {\em arXiv preprint arXiv:2403.07332}, 2024.

\bibitem{wang2023selective}
Jue Wang, Wentao Zhu, Pichao Wang, Xiang Yu, Linda Liu, Mohamed Omar, and Raffay Hamid.
\newblock Selective structured state-spaces for long-form video understanding.
\newblock In {\em Proceedings of the IEEE/CVF Conference on Computer Vision and Pattern Recognition}, pages 6387--6397, 2023.

\bibitem{wang2024mambabyte}
Junxiong Wang, Tushaar Gangavarapu, Jing~Nathan Yan, and Alexander~M Rush.
\newblock Mambabyte: Token-free selective state space model.
\newblock {\em arXiv preprint arXiv:2401.13660}, 2024.

\bibitem{wang2021tdn}
Limin Wang, Zhan Tong, Bin Ji, and Gangshan Wu.
\newblock Tdn: Temporal difference networks for efficient action recognition.
\newblock In {\em Proceedings of the IEEE/CVF conference on computer vision and pattern recognition}, pages 1895--1904, 2021.

\bibitem{wang2022scaled}
Pichao Wang, Xue Wang, Hao Luo, Jingkai Zhou, Zhipeng Zhou, Fan Wang, Hao Li, and Rong Jin.
\newblock Scaled relu matters for training vision transformers.
\newblock In {\em Proceedings of the AAAI Conference on Artificial Intelligence}, volume~36, pages 2495--2503, 2022.

\bibitem{wang2022bevt}
Rui Wang, Dongdong Chen, Zuxuan Wu, Yinpeng Chen, Xiyang Dai, Mengchen Liu, Yu-Gang Jiang, Luowei Zhou, and Lu~Yuan.
\newblock Bevt: Bert pretraining of video transformers.
\newblock In {\em Proceedings of the IEEE/CVF conference on computer vision and pattern recognition}, pages 14733--14743, 2022.

\bibitem{wang2020linformer}
Sinong Wang, Belinda~Z Li, Madian Khabsa, Han Fang, and Hao Ma.
\newblock Linformer: Self-attention with linear complexity.
\newblock {\em arXiv preprint arXiv:2006.04768}, 2020.

\bibitem{wang2024weak}
Ziyang Wang and Chao Ma.
\newblock Weak-mamba-unet: Visual mamba makes cnn and vit work better for scribble-based medical image segmentation.
\newblock {\em arXiv preprint arXiv:2402.10887}, 2024.

\bibitem{wichern2019wham}
Gordon Wichern, Joe Antognini, Michael Flynn, Licheng~Richard Zhu, Emmett McQuinn, Dwight Crow, Ethan Manilow, and Jonathan~Le Roux.
\newblock Wham!: Extending speech separation to noisy environments.
\newblock {\em arXiv preprint arXiv:1907.01160}, 2019.

\bibitem{wu2021towards}
Chao-Yuan Wu and Philipp Krahenbuhl.
\newblock Towards long-form video understanding.
\newblock In {\em Proceedings of the IEEE/CVF Conference on Computer Vision and Pattern Recognition}, pages 1884--1894, 2021.

\bibitem{wu2021autoformer}
Haixu Wu, Jiehui Xu, Jianmin Wang, and Mingsheng Long.
\newblock Autoformer: Decomposition transformers with auto-correlation for long-term series forecasting.
\newblock {\em Advances in Neural Information Processing Systems}, 34:22419--22430, 2021.

\bibitem{wu2024h}
Renkai Wu, Yinghao Liu, Pengchen Liang, and Qing Chang.
\newblock H-vmunet: High-order vision mamba unet for medical image segmentation.
\newblock {\em arXiv preprint arXiv:2403.13642}, 2024.

\bibitem{xie2024promamba}
Jianhao Xie, Ruofan Liao, Ziang Zhang, Sida Yi, Yuesheng Zhu, and Guibo Luo.
\newblock Promamba: Prompt-mamba for polyp segmentation.
\newblock {\em arXiv preprint arXiv:2403.13660}, 2024.

\bibitem{xing2024segmamba}
Zhaohu Xing, Tian Ye, Yijun Yang, Guang Liu, and Lei Zhu.
\newblock Segmamba: Long-range sequential modeling mamba for 3d medical image segmentation.
\newblock {\em arXiv preprint arXiv:2401.13560}, 2024.

\bibitem{xiong2021nystromformer}
Yunyang Xiong, Zhanpeng Zeng, Rudrasis Chakraborty, Mingxing Tan, Glenn Fung, Yin Li, and Vikas Singh.
\newblock Nystr{\"o}mformer: A nystr{\"o}m-based algorithm for approximating self-attention.
\newblock In {\em Proceedings of the AAAI Conference on Artificial Intelligence}, volume~35, pages 14138--14148, 2021.

\bibitem{yang2024plainmamba}
Chenhongyi Yang, Zehui Chen, Miguel Espinosa, Linus Ericsson, Zhenyu Wang, Jiaming Liu, and Elliot~J Crowley.
\newblock Plainmamba: Improving non-hierarchical mamba in visual recognition.
\newblock {\em arXiv preprint arXiv:2403.17695}, 2024.

\bibitem{yang2024cmvim}
Guangqian Yang, Kangrui Du, Zhihan Yang, Ye~Du, Yongping Zheng, and Shujun Wang.
\newblock Cmvim: Contrastive masked vim autoencoder for 3d multi-modal representation learning for ad classification.
\newblock {\em arXiv preprint arXiv:2403.16520}, 2024.

\bibitem{yang2024vivim}
Yijun Yang, Zhaohu Xing, and Lei Zhu.
\newblock Vivim: a video vision mamba for medical video object segmentation.
\newblock {\em arXiv preprint arXiv:2401.14168}, 2024.

\bibitem{yang2024remamber}
Yuhuan Yang, Chaofan Ma, Jiangchao Yao, Zhun Zhong, Ya~Zhang, and Yanfeng Wang.
\newblock Remamber: Referring image segmentation with mamba twister.
\newblock {\em arXiv preprint arXiv:2403.17839}, 2024.

\bibitem{yao2022wave}
Ting Yao, Yingwei Pan, Yehao Li, Chong-Wah Ngo, and Tao Mei.
\newblock Wave-vit: Unifying wavelet and transformers for visual representation learning.
\newblock In {\em Computer Vision--ECCV 2022: 17th European Conference, Tel Aviv, Israel, October 23--27, 2022, Proceedings, Part XXV}, pages 328--345. Springer, 2022.

\bibitem{ye2023mplug}
Qinghao Ye, Haiyang Xu, Guohai Xu, Jiabo Ye, Ming Yan, Yiyang Zhou, Junyang Wang, Anwen Hu, Pengcheng Shi, Yaya Shi, et~al.
\newblock mplug-owl: Modularization empowers large language models with multimodality.
\newblock {\em arXiv preprint arXiv:2304.14178}, 2023.

\bibitem{ye2024p}
Zi~Ye and Tianxiang Chen.
\newblock P-mamba: Marrying perona malik diffusion with mamba for efficient pediatric echocardiographic left ventricular segmentation.
\newblock {\em arXiv preprint arXiv:2402.08506}, 2024.

\bibitem{yin2023survey}
Shukang Yin, Chaoyou Fu, Sirui Zhao, Ke~Li, Xing Sun, Tong Xu, and Enhong Chen.
\newblock A survey on multimodal large language models.
\newblock {\em arXiv preprint arXiv:2306.13549}, 2023.

\bibitem{yu2024megabyte}
Lili Yu, D{\'a}niel Simig, Colin Flaherty, Armen Aghajanyan, Luke Zettlemoyer, and Mike Lewis.
\newblock Megabyte: Predicting million-byte sequences with multiscale transformers.
\newblock {\em Advances in Neural Information Processing Systems}, 36, 2024.

\bibitem{yu2023mm}
Weihao Yu, Zhengyuan Yang, Linjie Li, Jianfeng Wang, Kevin Lin, Zicheng Liu, Xinchao Wang, and Lijuan Wang.
\newblock Mm-vet: Evaluating large multimodal models for integrated capabilities.
\newblock {\em arXiv preprint arXiv:2308.02490}, 2023.

\bibitem{yuan2022volo}
Li~Yuan, Qibin Hou, Zihang Jiang, Jiashi Feng, and Shuicheng Yan.
\newblock Volo: Vision outlooker for visual recognition.
\newblock {\em IEEE Transactions on Pattern Analysis and Machine Intelligence}, 2022.

\bibitem{yuan2017chemical}
William Yuan, Dadi Jiang, Dhanya~K Nambiar, Lydia~P Liew, Michael~P Hay, Joshua Bloomstein, Peter Lu, Brandon Turner, Quynh-Thu Le, Robert Tibshirani, et~al.
\newblock Chemical space mimicry for drug discovery.
\newblock {\em Journal of chemical information and modeling}, 57(4):875--882, 2017.

\bibitem{yue2024medmamba}
Yubiao Yue and Zhenzhang Li.
\newblock Medmamba: Vision mamba for medical image classification.
\newblock {\em arXiv preprint arXiv:2403.03849}, 2024.

\bibitem{zaheer2020big}
Manzil Zaheer, Guru Guruganesh, Kumar~Avinava Dubey, Joshua Ainslie, Chris Alberti, Santiago Ontanon, Philip Pham, Anirudh Ravula, Qifan Wang, Li~Yang, et~al.
\newblock Big bird: Transformers for longer sequences.
\newblock {\em Advances in Neural Information Processing Systems}, 33:17283--17297, 2020.

\bibitem{zhao2024cobra}
Han Zhao, Min Zhang, Wei Zhao, Pengxiang Ding, Siteng Huang, and Donglin Wang.
\newblock Cobra: Extending mamba to multi-modal large language model for efficient inference.
\newblock {\em arXiv preprint arXiv:2403.14520}, 2024.

\bibitem{zheng2024fd}
Zhuoran Zheng and Jun Zhang.
\newblock Fd-vision mamba for endoscopic exposure correction.
\newblock {\em arXiv preprint arXiv:2402.06378}, 2024.

\bibitem{zhou2021informer}
Haoyi Zhou, Shanghang Zhang, Jieqi Peng, Shuai Zhang, Jianxin Li, Hui Xiong, and Wancai Zhang.
\newblock Informer: Beyond efficient transformer for long sequence time-series forecasting.
\newblock In {\em Proceedings of the AAAI Conference on Artificial Intelligence}, volume~35, pages 11106--11115, 2021.

\bibitem{zhou2021graph}
Jiaming Zhou, Kun-Yu Lin, Haoxin Li, and Wei-Shi Zheng.
\newblock Graph-based high-order relation modeling for long-term action recognition.
\newblock In {\em Proceedings of the IEEE/CVF Conference on Computer Vision and Pattern Recognition}, pages 8984--8993, 2021.

\bibitem{zhou2022fedformer}
Tian Zhou, Ziqing Ma, Qingsong Wen, Xue Wang, Liang Sun, and Rong Jin.
\newblock Fedformer: Frequency enhanced decomposed transformer for long-term series forecasting.
\newblock In {\em International Conference on Machine Learning}, pages 27268--27286. PMLR, 2022.

\bibitem{zhu2021long}
Chen Zhu, Wei Ping, Chaowei Xiao, Mohammad Shoeybi, Tom Goldstein, Anima Anandkumar, and Bryan Catanzaro.
\newblock Long-short transformer: Efficient transformers for language and vision.
\newblock {\em Advances in Neural Information Processing Systems}, 34:17723--17736, 2021.

\bibitem{zhu2023minigpt}
Deyao Zhu, Jun Chen, Xiaoqian Shen, Xiang Li, and Mohamed Elhoseiny.
\newblock Minigpt-4: Enhancing vision-language understanding with advanced large language models.
\newblock In {\em The Twelfth International Conference on Learning Representations}, 2023.

\bibitem{zhu2024vision}
Lianghui Zhu, Bencheng Liao, Qian Zhang, Xinlong Wang, Wenyu Liu, and Xinggang Wang.
\newblock Vision mamba: Efficient visual representation learning with bidirectional state space model.
\newblock {\em arXiv preprint arXiv:2401.09417}, 2024.

\bibitem{zhu2024llava}
Yichen Zhu, Minjie Zhu, Ning Liu, Zhicai Ou, Xiaofeng Mou, and Jian Tang.
\newblock Llava-phi: Efficient multi-modal assistant with small language model.
\newblock {\em arXiv preprint arXiv:2401.02330}, 2024.

\bibitem{zhu2021h}
Zhenhai Zhu and Radu Soricut.
\newblock H-transformer-1d: Fast one-dimensional hierarchical attention for sequences.
\newblock In {\em Proceedings of the 59th Annual Meeting of the Association for Computational Linguistics and the 11th International Joint Conference on Natural Language Processing (Volume 1: Long Papers)}, pages 3801--3815, 2021.

\bibitem{zuo2022efficient}
Simiao Zuo, Xiaodong Liu, Jian Jiao, Denis Charles, Eren Manavoglu, Tuo Zhao, and Jianfeng Gao.
\newblock Efficient long sequence modeling via state space augmented transformer.
\newblock {\em arXiv preprint arXiv:2212.08136}, 2022.

\end{thebibliography}
% }

%%%%%%%%%%%%%%%%%%%%%%%%%%%%%%%%%%%%%%%%%%%%%%%%%%%%%%%%%%%%
\appendix

\end{document}